%% file: ms.tex
\documentclass[runningheads]{llncs}
\usepackage{graphicx}
\usepackage{comment}
\usepackage{amsmath,amssymb} %
\usepackage{color}

\usepackage{xspace}

\makeatletter
\DeclareRobustCommand\onedot{\futurelet\@let@token\@onedot}
\def\@onedot{\ifx\@let@token.\else.\null\fi\xspace}

\def\eg{\emph{e.g}\onedot} 
 
\def\ie{\emph{i.e}\onedot} 
\def\cf{\emph{c.f}\onedot} \def\Cf{\emph{C.f}\onedot}
 \def\vs{\emph{vs}\onedot}
\def\wrt{w.r.t\onedot}

\makeatother
\usepackage{csquotes} %
\usepackage{amssymb} %
\usepackage{mathtools} %
\usepackage{placeins} %
\usepackage{bbold}
\usepackage{breqn}
\usepackage{graphbox} %
\usepackage{multirow}
\usepackage{booktabs} %
\usepackage{url}
\usepackage[utf8]{inputenc} %
\usepackage{hyperref}       %

\newcommand{\encoder}{E}
\newcommand{\decoder}{D}
\newcommand{\blackbox}{f}
\newcommand{\modelphi}{\Phi}
\newcommand{\modelpsi}{\Psi}
\newcommand{\modelrep}{z}
\newcommand{\modelrepdim}{N_\modelrep}
\newcommand{\modelinv}{v}
\newcommand{\aerep}{\bar{z}}
\newcommand{\aerepdim}{N_{\aerep}}
\newcommand{\x}{x}

\newcommand{\modelout}{\blackbox(\x)}
\newcommand{\xrec}{\bar{x}}
\newcommand{\normaldistr}{\mathcal{N}}
\newcommand{\id}{\mathbb{1}}
\newcommand{\condinn}{t}
\newcommand{\semanticinn}{e}
\newcommand{\compose}{\circ}

\newcommand{\semantic}{e}
\newcommand{\semanticdim}[1]{N_{\semantic_{#1}}}
\newcommand{\factorindices}{\mathcal{I}}
\newcommand{\loss}{\mathcal{L}}
\newcommand{\Nfactors}{K}
\newcommand{\diag}{\operatorname{diag}}

\newcommand{\correlationmatrix}{\Sigma^{ab}}

\newcommand{\RR}{\mathbb{R}}

\newcommand{\nimages}[1]{N^f}

\newcommand{\correlation}{\rho}
\providecommand{\impath}[1]{}
\providecommand{\imwidth}{}
\providecommand{\smallwidth}{}
\providecommand{\imscale}{}

\newcommand{\figmodel}{
\begin{figure*}[t]
\centering
\includegraphics[width=0.999\textwidth]{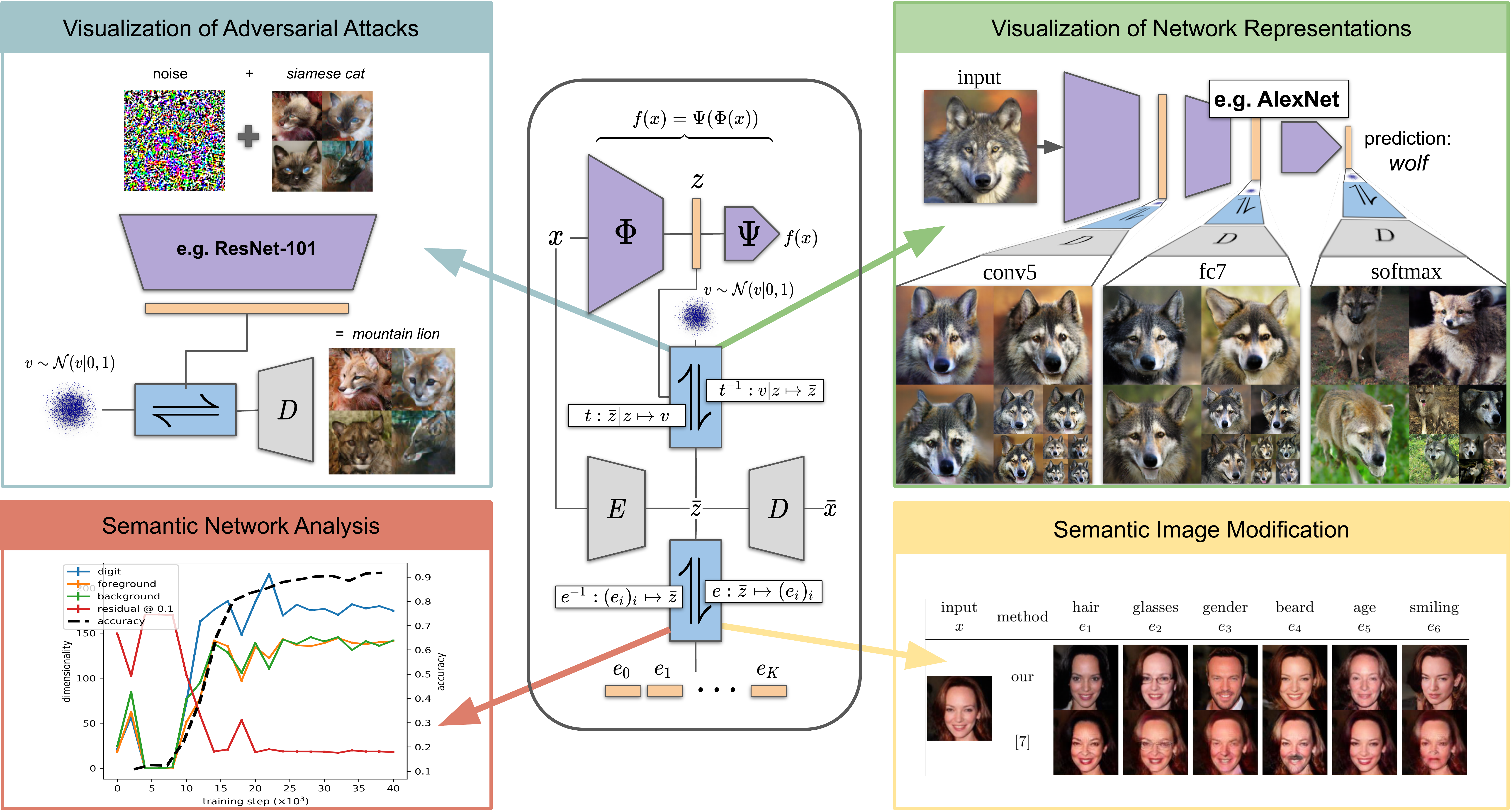}
\caption{
  Proposed architecture. We provide post-hoc interpretation for a given deep
  network $\blackbox = \modelpsi \compose \modelphi$. For a deep representation
  $\modelrep = \modelphi(\x)$ a conditional INN $\condinn$ recovers
  $\modelphi$'s invariances $\modelinv$ from a representation $\aerep$ which
  contains entangled information about \emph{both} $\modelrep$ and $\modelinv$.
  The INN $\semanticinn$ then translates the representation $\aerep$ into a
  factorized representation with accessible semantic concepts. This approach
  allows for various applications, including visualizations of network
  representations of natural and altered inputs, semantic network analysis
  and semantic image modifications.
}
\label{fig:overview} 
\end{figure*}
}

\newcommand{\comparefidsbrox}{
\begin{table}[t!]
  \setlength{\tabcolsep}{1em}
  \centering
  \caption{
    FID scores for layer visualizations of \emph{AlexNet}, obtained with our method and \cite{dosovitskiy2016generating} (D\&B). Scores are calculated on the \textsl{Animals} dataset.}
      \begin{small}
  \begin{tabular}{cccccc}
  \toprule
    \textsl{layer} &
    conv5 &
	fc6  &
    fc7 &
    fc8  &
    output \\
    \midrule
    \textbf{ours} &
    $\mathbf{23.6 \pm 0.5}$ &
    $\mathbf{24.3 \pm 0.7}$ &
    $\mathbf{24.9 \pm 0.4}$ &
    $\mathbf{26.4 \pm 0.4}$ &
    $\mathbf{27.4 \pm 0.3}$ \\
    
    D\&B &
    $25.2$ &
    $\mathbf{24.9}$ &
    $27.2$ &
    $36.1$ &
    $352.6$ \\
	\bottomrule

  \end{tabular}
      \end{small}
  \label{tab:comparefidsbrox}
 \end{table}
}

\newcommand{\factorevolution}{
\begin{figure}[t]
\centering
\includegraphics[width=0.95\textwidth]{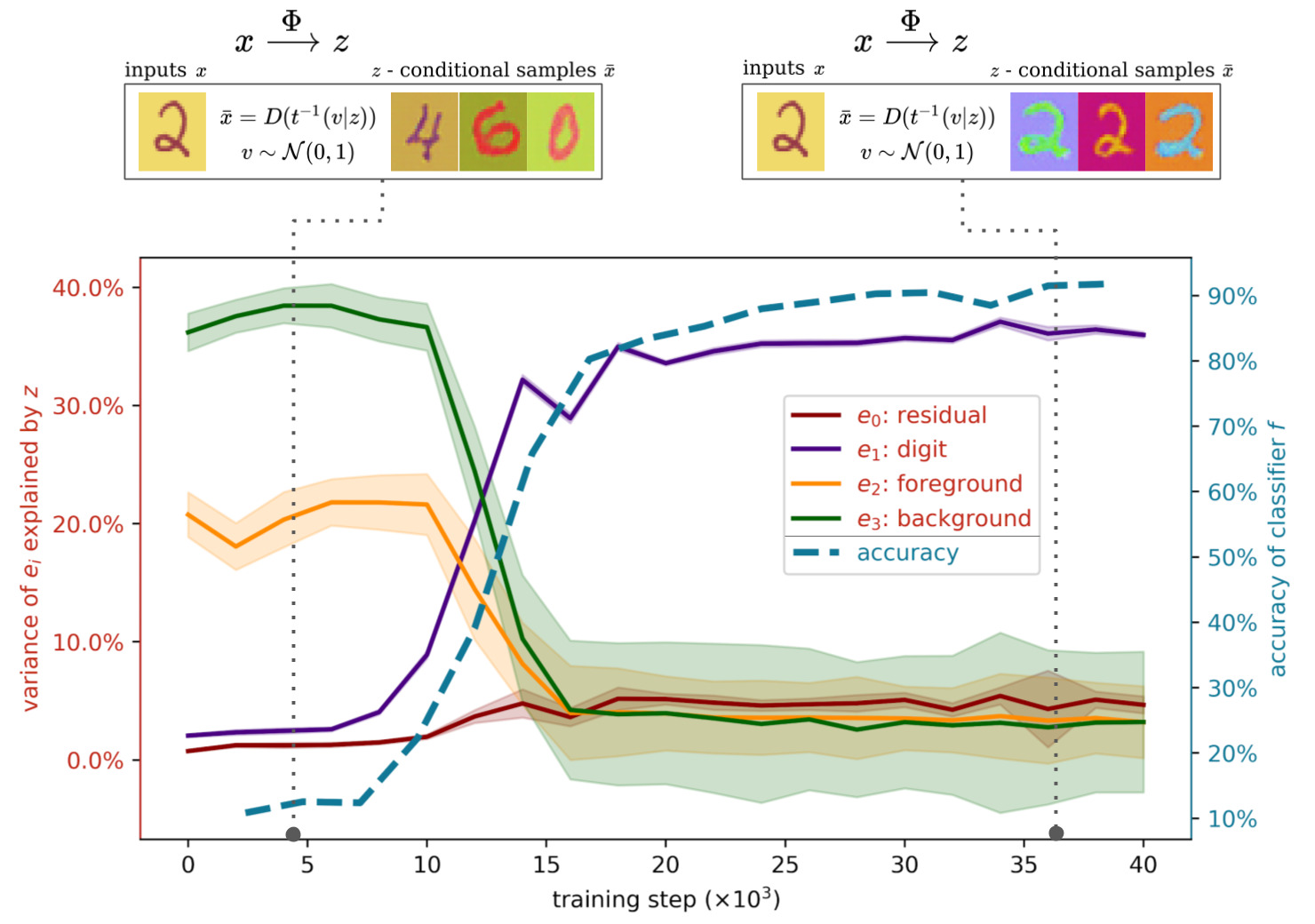}
\caption{
Analyzing the degree to which different semantic concepts are captured by a
  network representation changes as training progresses. For \emph{SqueezeNet} on
  \emph{ColorMNIST} we measure how much the data varies in different semantic concepts
  $e_i$ and how much of this variability is captured by $z$ at different
  training iterations.
  Early on $z$ is sensitive to foreground and
  background color, and later on it learns to focus on the digit attribute.
  The ability to encode this semantic concept is proportional to the
  classification accuracy achieved by $z$. At training iterations 4k and 36k we
  apply our method to visualize model representations and thereby illustrate
  how their content changes during training.
}
  \label{fig:factorevolution} \end{figure}
}

\newcommand{\advattacklayers}{
\renewcommand{\impath}[1]{img/attack/##1}
\renewcommand{\imwidth}{0.12\textwidth}
\renewcommand{\smallwidth}{0.095\textwidth}
\begin{figure}[t]
\centering
  \begin{tabular}{cccccccc}
    & &  & \multicolumn{4}{c}{visualizing perturbed representation at} & \\
    \cmidrule{4-7}
    \multicolumn{2}{c}{perturbation} & x & input & conv & fc & logits & prediction \\
    \midrule
  \parbox{\smallwidth}{\centering none} &
   &
  \includegraphics[width=\imwidth, align=c]{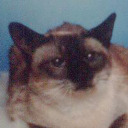} &
  \includegraphics[width=\imwidth, align=c]{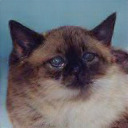} &
  \includegraphics[width=\imwidth, align=c]{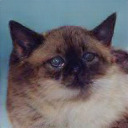} &
  \includegraphics[width=\imwidth, align=c]{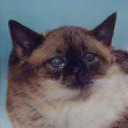} &
  \includegraphics[width=\imwidth, align=c]{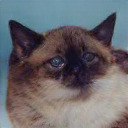} &
  \parbox{\imwidth}{\centering siamese cat} \\
  \parbox{\smallwidth}{\centering random} &
  \includegraphics[width=\imwidth, align=c]{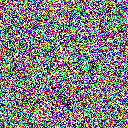} &
  \includegraphics[width=\imwidth, align=c]{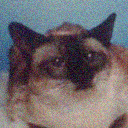} &
  \includegraphics[width=\imwidth, align=c]{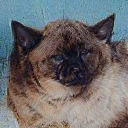} &
  \includegraphics[width=\imwidth, align=c]{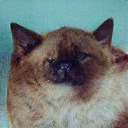} &
  \includegraphics[width=\imwidth, align=c]{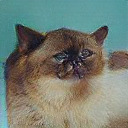} &
  \includegraphics[width=\imwidth, align=c]{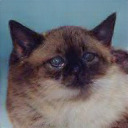} &
    \parbox{\imwidth}{\centering siamese cat} \\
  \parbox{\smallwidth}{\centering attack} &
  \includegraphics[width=\imwidth, align=c]{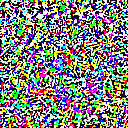} &
  \includegraphics[width=\imwidth, align=c]{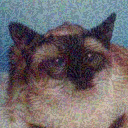} &
  \includegraphics[width=\imwidth, align=c]{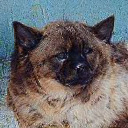} &
  \includegraphics[width=\imwidth, align=c]{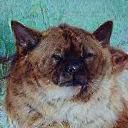} &
  \includegraphics[width=\imwidth, align=c]{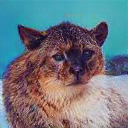} &
  \includegraphics[width=\imwidth, align=c]{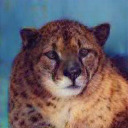} &
    \parbox{\imwidth}{\centering mountain lion} \\
  \midrule
    \multicolumn{3}{c}{\multirow{2}{*}{variance of $\aerep$ explained by $\modelinv$}} &
  ${11.82\%}$ &
  ${7.22\%}$ &
  ${49.59\%}$ &
  ${84.77\%}$ &
  \\
    \multicolumn{3}{c}{} &
  ${\scriptstyle (\pm 0.52)}$ &
  ${\scriptstyle (\pm 0.16)}$ &
  ${\scriptstyle (\pm 2.00)}$ &
  ${\scriptstyle (\pm 5.77)}$ &
  \\
\end{tabular}
  \caption{Visualizing FGSM adversarial attacks on \emph{ResNet-101}. To the
  human eye, the original image and its attacked version are almost
  indistinguishable. However, the input image is correctly classified as
  "siamese cat", while the attacked version is classified as "mountain
  lion". Our approach visualizes how the attack spreads throughout the network.
  Reconstructions of representations of attacked images demonstrate that the
  attack targets the semantic content of deep layers.
  The variance of $\aerep$ explained by $\modelinv$ combined with these visualizations
  show how increasing invariances cause vulnerability to adversarial attacks.
  }
\label{fig:attackanimalslayers}
\end{figure}
}

\newcommand{\facenetlayersdecoded}{
\begin{figure}[t]
\centering
\includegraphics[width=0.99\textwidth]{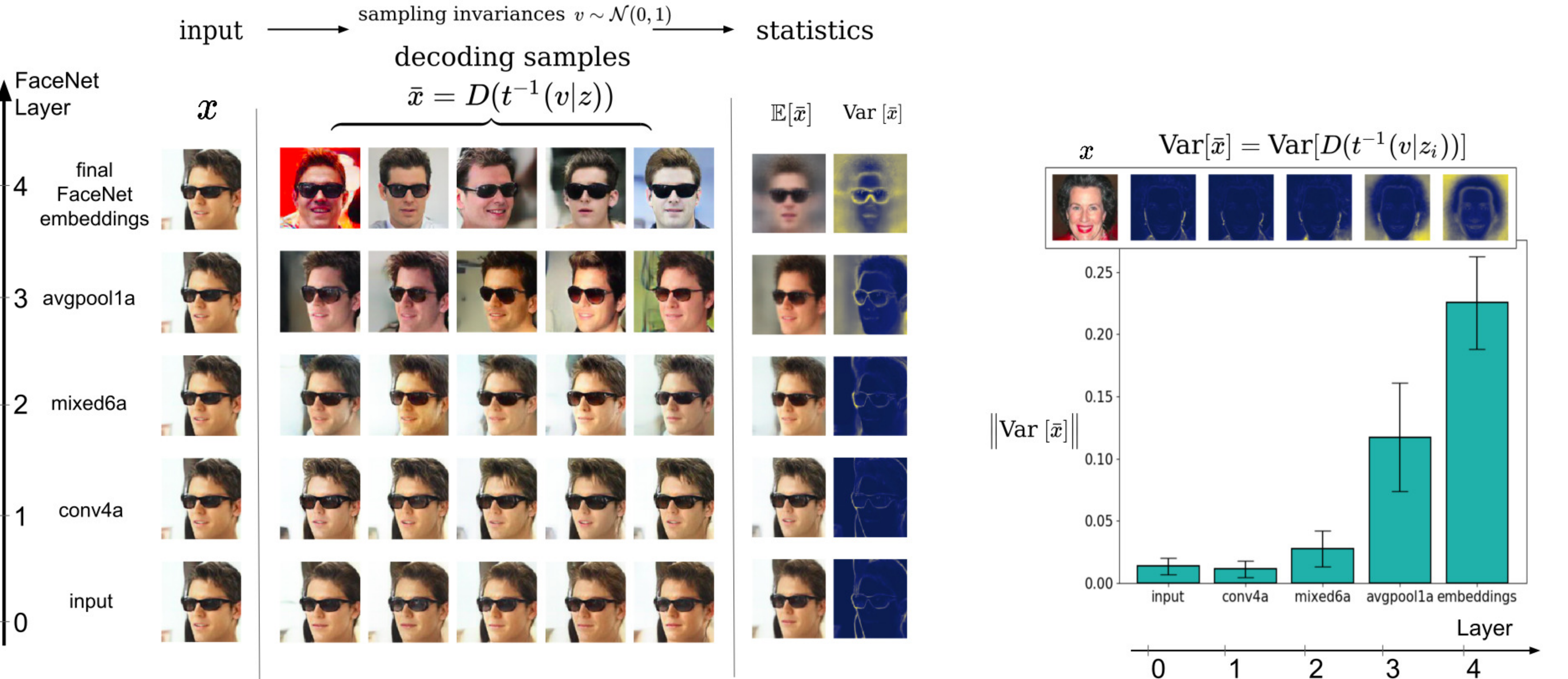}
\caption{
  \emph{left}: Visualizing \emph{FaceNet} representations and their invariances.
  Sampling multiple reconstructions $\xrec = \decoder(\condinn^{-1}(\modelinv
  \vert \modelrep))$ shows the degree of invariance learned by different layers.
  The invariance \wrt pose increases for deeper layers as expected for face
  identification. Surprisingly, FaceNet uses glasses as an identity feature
  throughout all its layers as evident from the spatial mean and variance plots,
  where the glasses are still visible. This reveals a bias and weakness
  of the model. \emph{right}: Spatially averaged variances over multiple $\x$
  for different layers.}
\label{fig:facenetlayersdecoded} \end{figure}
}

\newcommand{\texturebias}{
\renewcommand{\imscale}{0.16}
\begin{figure*}[!t]
\centering
  \begin{tabular}{c@{\hskip 0.8em}c@{\hskip 0.8em}c}
 & \multicolumn{2}{c}{samples $\xrec = \decoder(\condinn^{-1}(\modelinv \vert \modelrep))$ conditioned on \emph{ResNet} pre-logits $\modelrep = \modelphi(\x)$}\\ 
    \cmidrule{2-3}
 \multirow{2}{*}{inputs}  & $\modelphi_{vanilla}$: ResNet-50 trained on & $\modelphi_{stylized}$: ResNet-50 trained on  \\
 & standard ImageNet & stylized ImageNet \\
\midrule
\includegraphics[scale=\imscale, align=c]{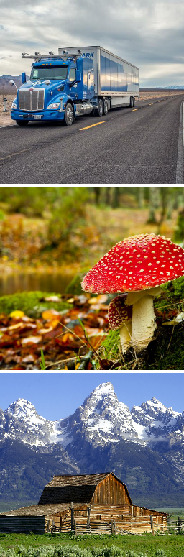} & \includegraphics[scale=\imscale, align=c]{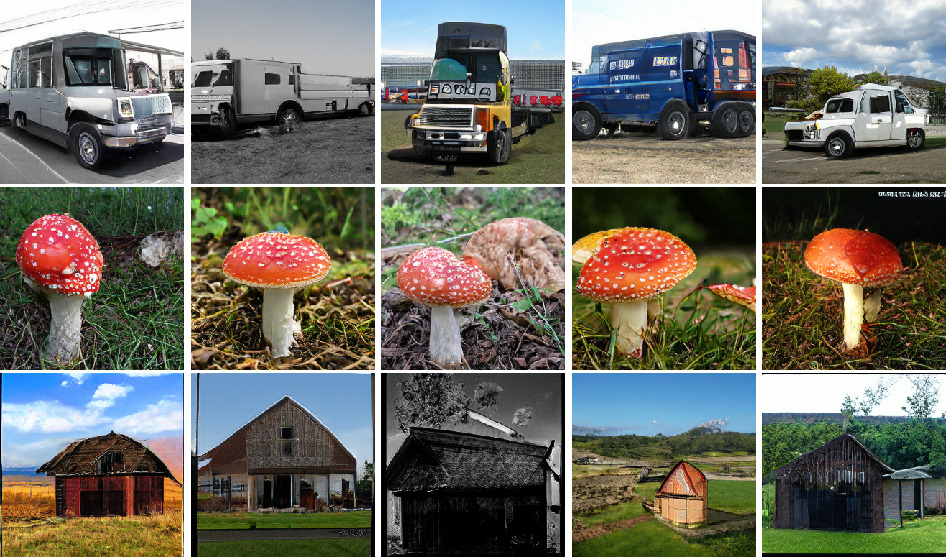} & \includegraphics[scale=\imscale, align=c]{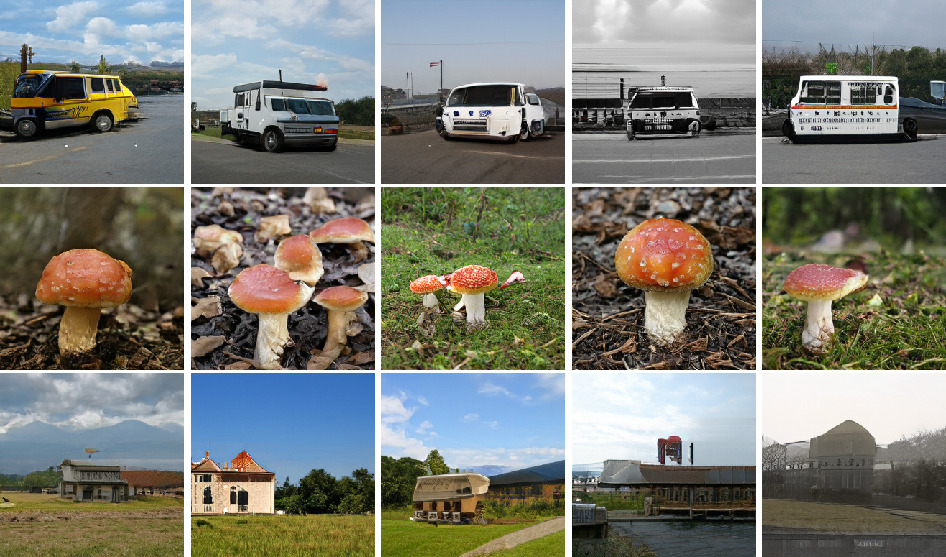} \\
\midrule
 \includegraphics[scale=\imscale, align=c]{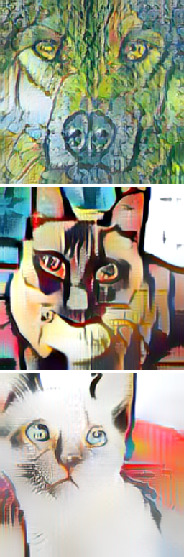} & \includegraphics[scale=\imscale, align=c]{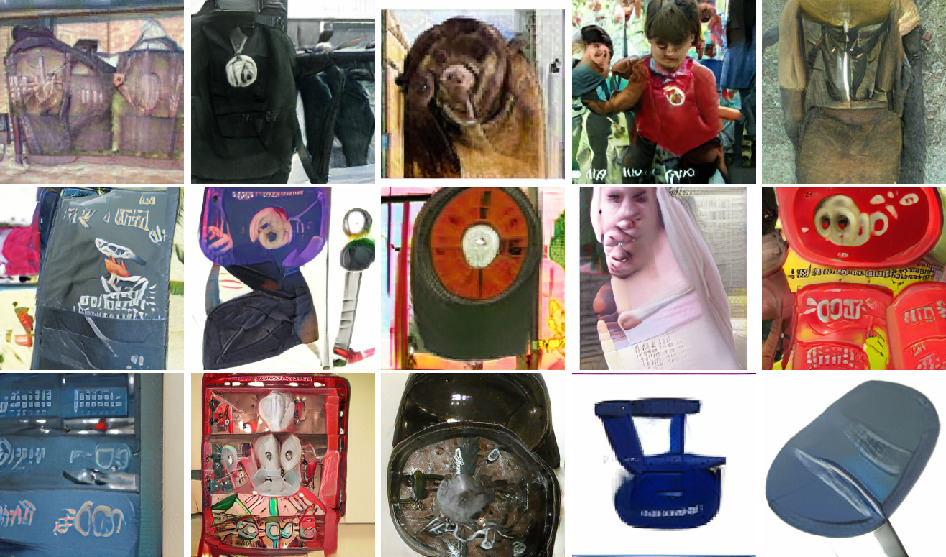} & \includegraphics[scale=\imscale, align=c]{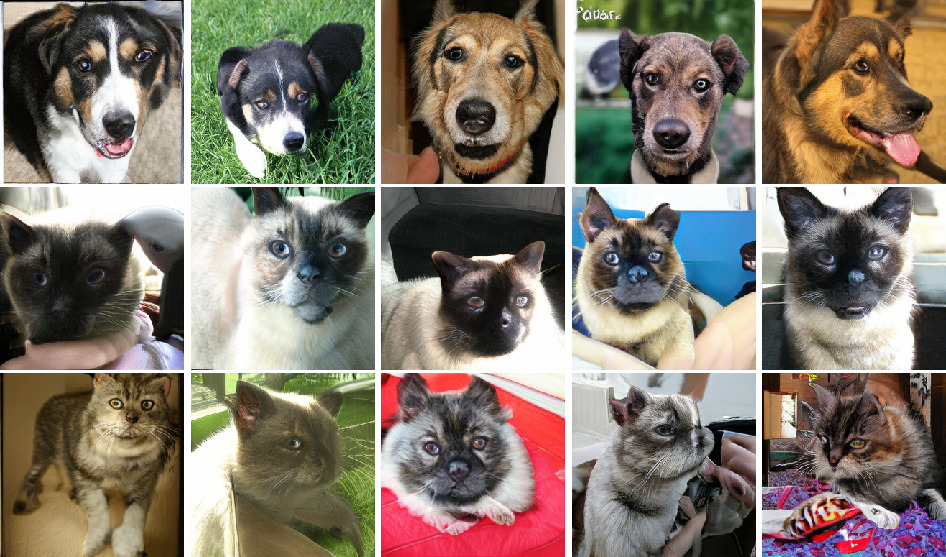} \\
\midrule
 \includegraphics[scale=\imscale, align=c]{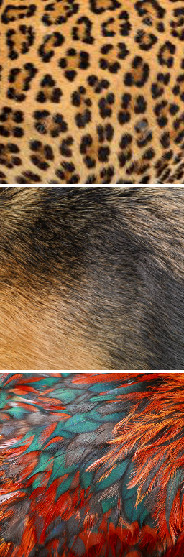} & \includegraphics[scale=\imscale, align=c]{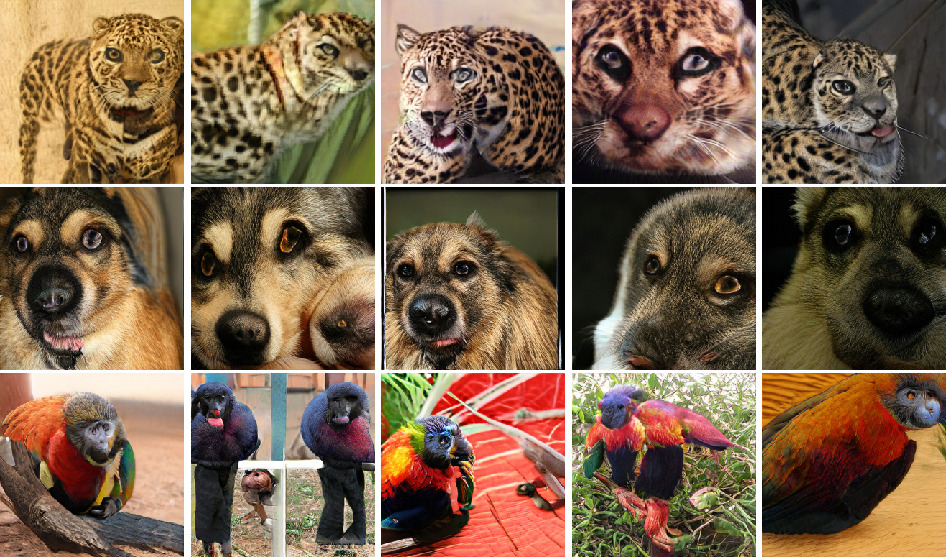} & \includegraphics[scale=\imscale, align=c]{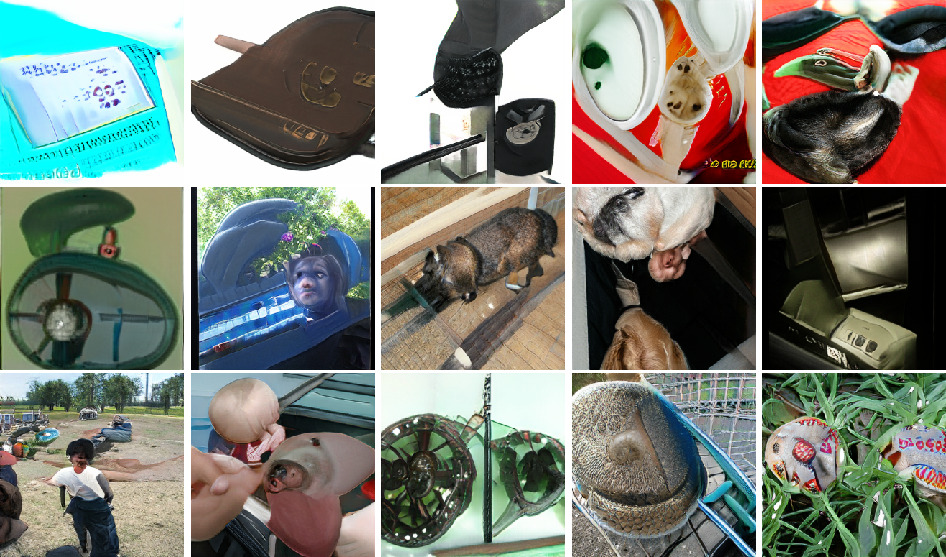} \\
\midrule
\includegraphics[scale=\imscale, align=c]{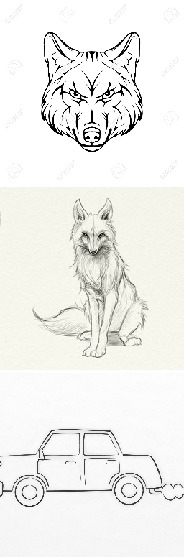} & \includegraphics[scale=\imscale, align=c]{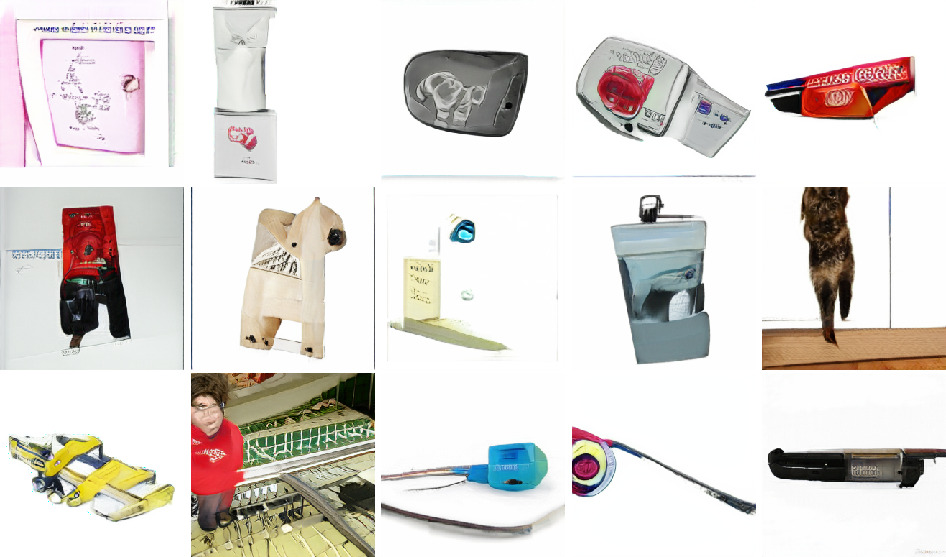} & \includegraphics[scale=\imscale, align=c]{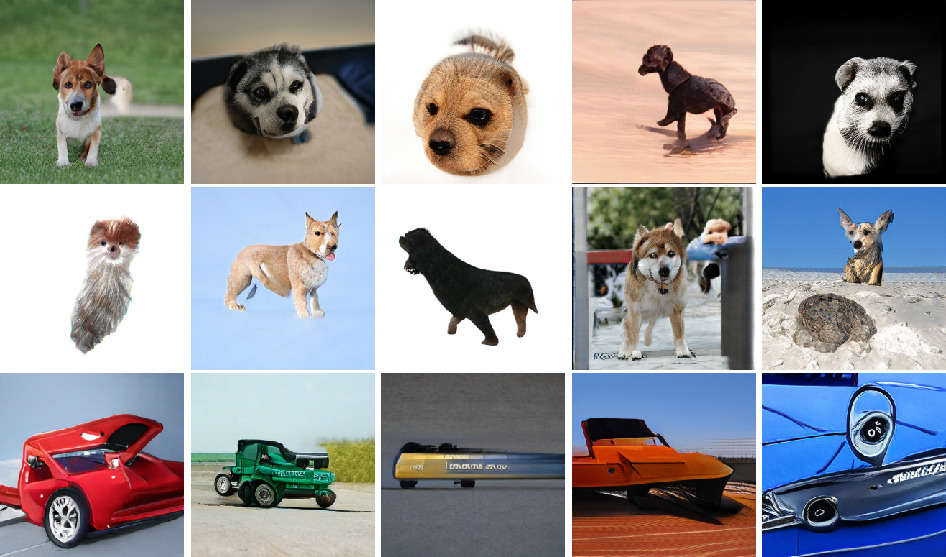}

\end{tabular}
\caption{
  Revealing texture bias in ImageNet classifiers. We compare visualizations of
  $z$ from the penultimate layer of \emph{ResNet-50}
  trained on standard ImageNet (left) and a stylized version of
  ImageNet (right).
  On natural images (rows 1-3) both models recognize the input,
  removing textures through stylization (rows 4-6) makes images unrecognizable
  to the standard model, however it recognizes objects from textured patches
  (rows 7-9). Rows 10-12 show that a model without texture bias can be used for
  sketch-to-image synthesis.
  \vspace{1em}
}
\label{fig:texturebias}
\end{figure*}
}

\newcommand{\comparison}{
\renewcommand{\imwidth}{0.875\textwidth}
\renewcommand{\smallwidth}{0.165\textwidth}
\begin{figure*}[!t]
\centering
\begin{tabular}{cc}
 & reconstructions $\xrec$ from representations $\modelrep = \modelphi(\x)$ of different layers \\	
  \cmidrule{2-2}
  method &
  \parbox{\smallwidth}{\centering input}
  \parbox{\smallwidth}{\centering conv5}
  \parbox{\smallwidth}{\centering fc6}
  \parbox{\smallwidth}{\centering fc7}
  \parbox{\smallwidth}{\centering fc8}
  \\
\midrule
\textbf{ours} & \includegraphics[align=c, width=\imwidth]{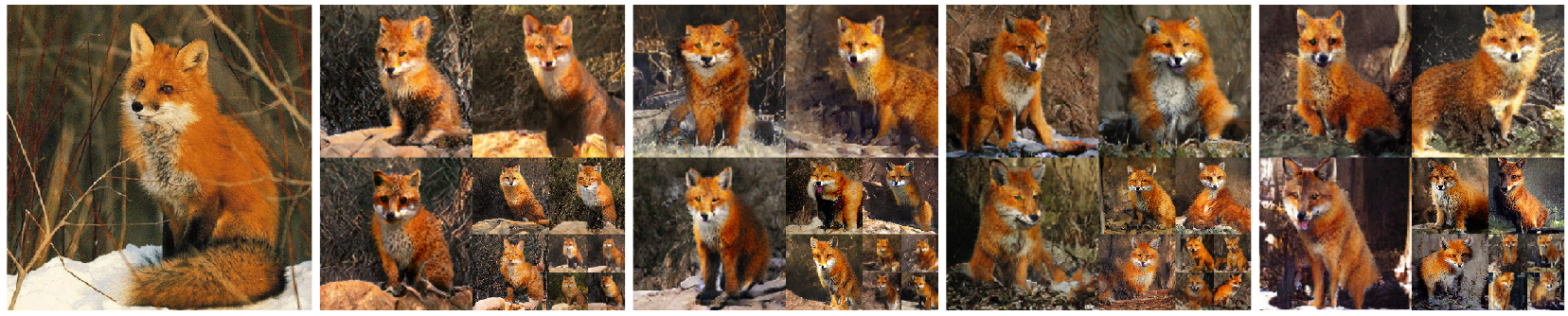} \\
D\&B \cite{dosovitskiy2016generating} & \includegraphics[align=c, width=\imwidth]{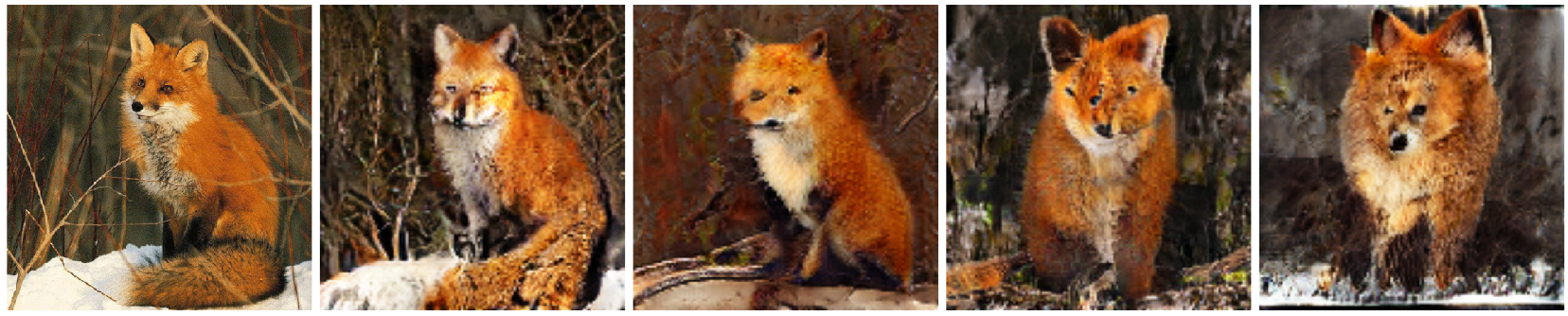} \\
M\&V \cite{mahendran2016visualizing} & \includegraphics[align=c, width=\imwidth]{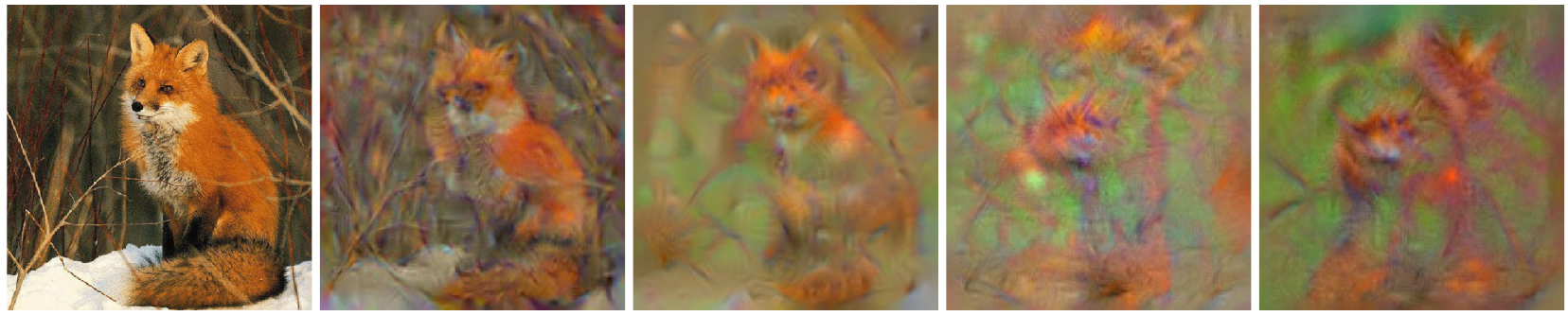} \\
\end{tabular}
\caption{Comparison to existing network inversion methods for \emph{AlexNet}
  \cite{krizhevsky2012imagenet}. In contrast to the methods of \cite{dosovitskiy2016generating} (D\&B) and \cite{mahendran2016visualizing} (M\&V), our invertible method explicitly samples the invariances of $\modelphi$ \wrt the data, which circumvents a common cause for artifacts and produces natural images independent of the depth of the layer which is reconstructed.
} 
\label{fig:comparison}
\end{figure*}
}

\newcommand{\facemod}{
\renewcommand{\imscale}{0.325}
\begin{figure*}[!t]
\centering
\begin{tabular}{ccccccc}
  input & hair & glasses & gender & beard & age & smiling\\ 
  $\x$ & $\semantic_1$ & $\semantic_2$ & $\semantic_3$ & $\semantic_4$ &
  $\semantic_5$ & $\semantic_6$ \\ 
\midrule
  \includegraphics[scale=\imscale, align=c]{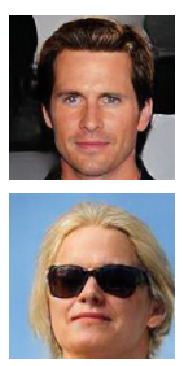} &
  \includegraphics[scale=\imscale, align=c]{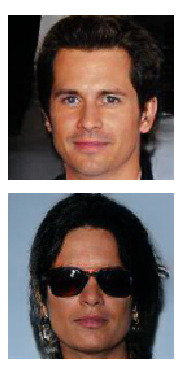} &
  \includegraphics[scale=\imscale, align=c]{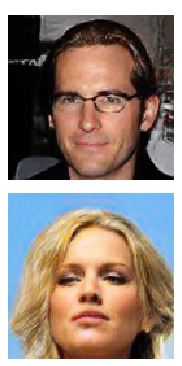} &
  \includegraphics[scale=\imscale, align=c]{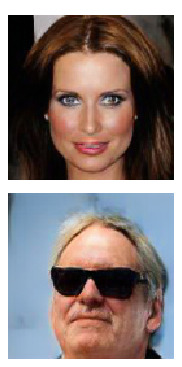} &
  \includegraphics[scale=\imscale, align=c]{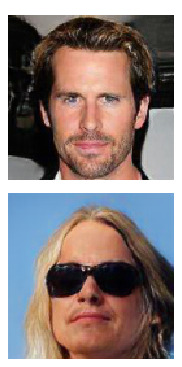} &
  \includegraphics[scale=\imscale, align=c]{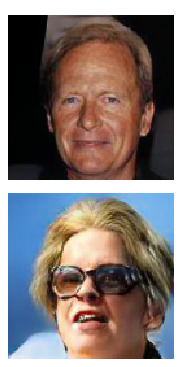} &
  \includegraphics[scale=\imscale, align=c]{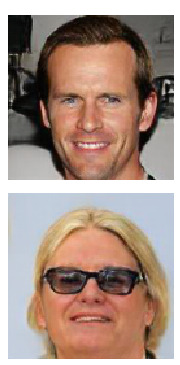} \\
\midrule
  mean embedding&
  ${0.872}$ &
  ${1.000}$ &
  ${1.061}$ &
  ${0.803}$ &
  ${0.874}$ &
  ${0.833}$ \\

  distance ($\pm$ std)&
  ${\scriptstyle (\pm 0.048)}$ &
  ${\scriptstyle (\pm 0.046)}$ &
  ${\scriptstyle (\pm 0.030)}$ &
  ${\scriptstyle (\pm 0.041)}$ &
  ${\scriptstyle (\pm 0.053)}$ &
  ${\scriptstyle (\pm 0.034)}$ \\

\end{tabular}
\caption{
  Semantic Modifications on CelebA.
  For each column, after inferring the semantic factors
  $(\semantic_i)_i=\semanticinn(\encoder(\x))$ of the input $x$,
  we replace one factor $\semantic_i$ by that from another randomly chosen
  image that differs in this concept. The inverse of $\semanticinn$ translates
  this semantic change back into a modified $\aerep$, which is decoded to a
  semantically modified image.
  Distances between \emph{FaceNet} embeddings before and after modification
  demonstrate its sensitivity to differences in gender and glasses (see also
  Fig.~\ref{fig:facenetlayersdecoded}).
}
\label{fig:facemod}
\end{figure*}
}

\begin{document}
\pagestyle{headings}
\mainmatter
\def\ECCVSubNumber{2861}  %

\title{Making Sense of CNNs: Interpreting Deep Representations \& Their Invariances with INNs}

\titlerunning{Making Sense of CNNs}
\author{Robin Rombach\thanks{Both authors contributed equally to this work.}
  \and Patrick Esser\inst{\star} \and Bj\"orn Ommer}

\authorrunning{R. Rombach et al.}
\institute{Interdisciplinary Center for Scientific Computing, HCI, Heidelberg University \\
}
\maketitle

\begin{abstract}
  To tackle increasingly complex tasks, it has become an essential ability of
  neural networks to learn abstract representations. These task-specific
  representations and, particularly, the invariances they capture turn neural
  networks into black box models that lack interpretability. To open such a
  black box, it is, therefore, crucial to uncover the different semantic
  concepts a model has learned as well as those that it has learned to be
  invariant to. We present an approach based on INNs that
  (i) recovers the task-specific, learned invariances by disentangling the
  remaining factor of variation in the data and that
  (ii) invertibly transforms these recovered invariances combined with the
  model representation into an equally expressive one with accessible semantic
  concepts.
  As a consequence, neural network representations
  become understandable by providing the means to (i) expose their semantic
  meaning, (ii) semantically modify a representation, and (iii) visualize
  individual learned semantic concepts and invariances. Our invertible approach
  significantly extends the abilities to understand black box models by
  enabling post-hoc interpretations of state-of-the-art networks without
  compromising their performance.
  Our implementation is available at \url{https://compvis.github.io/invariances/}.
\end{abstract}

\section{Introduction}

Key to the wide success of deep neural networks is end-to-end learning of
powerful hidden representations that aim to \emph{(i)} capture all
task-relevant characteristics while \emph{(ii)} being invariant to all other
variability in the data \cite{lecun2012learning,achille2018emergence}. Deep learning can yield abstract
representations that are perfectly adapted feature encodings for the task at
hand. However, their increasing abstraction capability and performance comes at
the expense of a lack in interpretability \cite{bach2015pixel}: Although the network may
solve a problem, it does not convey an understanding of its predictions or
their causes, oftentimes leaving the impression of a black box \cite{miller2019explanation}. In
particular, users are missing an explanation of semantic concepts that the
model has learned to \emph{represent} and of those it has learned to
\emph{ignore}, i.e. its invariances.

Providing such explanations and an understanding of network predictions and
their causes is thus crucial for transparent AI. Not only is this relevant to
discover limitations and promising directions for future improvements of the AI
system itself, but also for compliance with legislation
\cite{Goodman_2017,eu2020artificial}, knowledge distillation from such a system
\cite{lipton2016mythos}, and post-hoc verification of the model
\cite{samek2017explainable}. Consequently, research on interpretable deep
models has recently gained a lot of attention, particularly methods that
investigate latent representations to understand what the model has learned
\cite{samek2017explainable,szegedy2013intriguing,Bau_2017,Fong_2018,esser2020disentangling}.

\textbf{Challenges \& aims}
Assessing these latent representations is challenging due to two fundamental
issues: \emph{(i)} to achieve robustness and generalization despite noisy
inputs and data variability, hidden layers exhibit a distributed coding of
semantically meaningful concepts \cite{fong2018net2vec}. Attributing semantics
to a single neuron via backpropagation \cite{montavon2017explaining} or
synthesis \cite{yosinski2015understanding} is thus impossible without altering
the network \cite{montavon2018methods,Zhou_2016}, which typically degrades performance.
\emph{(ii)} end-to-end learning trains deep representations towards a goal
task, making them invariant to features irrelevant for this goal. Understanding
these characteristics that a representation has abstracted away is challenging,
since we essentially need to portray features that have been discarded.

\figmodel
These challenges call for a method that can interpret existing network
representations by recovering their invariances without modifying them. Given
these recovered invariances, we seek an invertible mapping that translates a
representation \emph{and} the invariances onto understandable semantic
concepts. The mapping disentangles the distributed encoding of the
high-dimensional representation and its invariances by projecting them onto
separate multi-dimensional factors that correspond to human understandable
semantic concepts. Both this translation and the recovering of invariances are implemented with invertible neural
networks (INNs)
\cite{redlich1993supervised,dinh2016density,kingma2018glow}.
For the translation, this guarantees that the resulting understandable
representation is equally expressive as the model representation combined with
the recovered invariances (no information is lost). Its invertibility also
warrants that feature modifications applied in the semantic domain correctly
adjust the recovered representation.

\textbf{Our contributions} to a comprehensive understanding of deep representations are
as follows: (i) We present an approach, which, by utilizing invertible neural
networks, improves the understanding of representations produced by existing
network architectures with no need for re-training or otherwise compromising
their performance.
(ii) Our generative approach is able to recover the invariances
that result from the non-injective projection (of input onto a latent
representation) which deep networks typically learn. This model then provides a
probabilistic visualization of the latent representation and its invariances.
(iii) We bijectively translate an arbitrarily abstract representation and its
invariances via a non-linear transformation into another representation of
equal expressiveness, but with accessible semantic concepts.
(iv) The invertibility also enables manipulation of the original latent
representations in a semantically understandable manner, thus facilitating
further diagnostics of a network.

\section{Background}
Two main approaches to interpretable AI can be identified, those which aim to
incorporate interpretability directly into the design of models, and those
which aim to provide interpretability to existing models \cite{montavon2018methods}.
Approaches from the first category range from modifications of network
architectures \cite{Zhou_2016}, over regularization of models encouraging
interpretability \cite{Lorenz2019UnsupervisedPD,plumb2019regularizing}, towards combinations of both
\cite{zhang2018interpretable}. However, these approaches always involve a
trade-off between model performance and model interpretability. Being of the
latter category, our approach allows to interpret representations of existing
models without compromising their performance.

To better understand what an existing model has learned, its representations
must be studied \cite{samek2017explainable}. \cite{szegedy2013intriguing} shows
that both random directions and coordinate axes in the feature space of
networks can represent semantic properties and concludes that they are not
necessarily represented by individual neurons. Different works attempt to
select groups of neurons which have a certain semantic meaning, such as based
on scenes \cite{zhou2014object}, objects \cite{Simon_2015} and object parts
\cite{Simon2014PartDD}. \cite{Bau_2017} studied the interpretability of neurons, and found that a rotation of the representation space
spanned by the neurons decreases its interpretability. While this suggests
that the neurons provide a more interpretable basis compared to a random basis,
\cite{Fong_2018} shows that the choice of basis is not the only challenge for
interpretability of representations. Their findings demonstrate that learned
representations are distributed, \ie a single semantic concept is encoded by an
activation pattern involving multiple neurons, and a single neuron is involved
in the encoding of multiple different semantic concepts.
Instead of selecting a set of neurons directly, \cite{esser2020disentangling}
learns an INN that transforms the original representation space to an
interpretable space, where a single semantic concept is represented by a known
group of neurons and a single neuron is involved in the encoding of just a
single semantic concept.
However, to interpret not only the representation itself but also its
invariances, it is insufficient to transform only the representation itself.
Our approach therefore transforms the latent representation space of an
autoencoder, which has the capacity to represent
its inputs faithfully,
and subsequently translates a model representation and its invariances into
this space for semantic interpretation and visualization.

A large body of works approach interpretability of existing networks based on
visualizations. \cite{Selvaraju_2019} uses gradients of network outputs with
respect to a convolutional layer to obtain coarse localization maps.
\cite{bach2015pixel} proposes an approach to obtain pixel-wise relevance scores
for a specific class of models which is generalized in
\cite{montavon2017explaining}. To obtain richer visual interpretations,
\cite{Zeiler_2014,simonyan2013deep,yosinski2015understanding,mahendran2016visualizing}
reconstruct images which maximally activate certain neurons.
\cite{nguyen2016synthesizing} uses a generator network for this task, which was
introduced in \cite{dosovitskiy2016generating} for reconstructing images from
their feature representation.
Our key insight is that these existing approaches do not explicitly account for
the invariances learned by a model. Invariances imply that feature
inversion is a one-to-many mapping and thus they must be recovered to solve the task.
Recently, \cite{Shocher_2020_CVPR} introduced a GAN-based approach that
utilizes features of a pre-trained classifier as a semantic pyramid for image
generation.
\cite{nash2019inverting} used samples from an autoregressive model of images
conditioned on a feature representation to gain insights into the
representation's invariances.
In contrast, our approach recovers an explicit representation of the
invariances, which can be recombined with modified feature representations, and
thus makes the effect of modifications to representations, \eg through
adversarial attacks, visible.

Other works consider visual interpretations for specialized models.
\cite{santurkar2019image} showed that the quality of images which maximally
activate certain neurons is significantly improved when activating neurons of
an adversarially robust classifer.
\cite{bau2018gan} explores the relationship between neurons and the images
produced by a Generative Adversarial Network.
For the same class of models, \cite{goetschalckx2019ganalyze} finds directions
in their input space which represent semantic concepts corresponding to certain
cognitive properties.
Such semantic directions have previously also been found in classifier networks
\cite{upchurch2017deep} but requires aligned data.
All of these approaches require either special training of models, are
limited to a very special class of models which already provide visualizations
or depend on special assumptions on model and data. In contrast, our approach
can be applied to arbitrary models without re-training or modifying them, and
provides both visualizations and semantic explanations, for both the model's
representation and its learned invariances.

\section{Approach}
Common tasks of computer vision can be phrased as a mapping from an
input image $\x$ to some output
$\blackbox(\x)$ such as a
classification of the image, a regression (e.g. of object locations), a
(semantic) segmentation map, or a re-synthesis that yields another image. Deep
learning utilizes a hierarchy of intermediate network layers that gradually
transform the input into increasingly more abstract representations. Let
$\modelrep=\modelphi(\x) \in \mathbb{R}^{\modelrepdim}$ be the representation
extracted by one such layer (without loss of generality we consider $\modelrep$
to be a $\modelrepdim$-dim vector, flattening it if necessary) and
$\modelout=\modelpsi(\modelrep)=\modelpsi(\modelphi(\x))$ the mapping onto the
output.

An essential characteristic of a deep feature encoding $\modelrep$ is the
increasing abstractness of higher feature encoding layers and the resulting
reduction of information.
To explain a latent representation, we need to recover its invariances
$\modelinv$ and
make
$\modelrep$ \emph{and} $\modelinv$ interpretable by learning a bijective
mapping onto understandable semantic concepts, see Fig.~\ref{fig:overview}.
Sec.~\ref{sec:recover} describes our INN $\condinn$ to recover an encoding
$\modelinv$ of the invariances. Due to the generative nature of $\condinn$, our
approach can correctly sample visualizations of the model representation and
its invariances without
leaving the underlying data distribution and introducing artifacts. With
$\modelinv$ then available, Sec.~\ref{sec:semantic} presents an INN
$\semanticinn$ that translates $\condinn$'s encoding of $\modelrep$ and
$\modelinv$ without losing information onto disentangled semantic concepts.
Moreover, the invertibility allows modifications in the semantic domain to
correctly project back onto the original representation or into image
space.

\subsection{Recovering the Invariances of Deep Models}
\label{sec:recover}
\textbf{Learning an Encoding to Help Recover Invariances} Key to a deep representation is not only the information
$\modelrep$ captures, but also what is has learned to abstract away. To learn
what $\modelrep$ misses with respect to $\x$, we need an encoding $\aerep$,
which, in contrast to $\modelrep$, includes these invariances. Without making
prior assumptions about the deep model $\blackbox$, autoencoders provide a
generic way to obtain such an encoding $\aerep$, since they ensure that their
input $\x$ can be recovered from their learned representation $\aerep$, which
hence also comprises the invariances.

Therefore, we learn an autoencoder with an encoder $\encoder$ that provides the
data representation $\aerep = \encoder(\x)$ and a decoder $\decoder$ producing
the data reconstruction $\xrec = \decoder(\aerep)$. Sec.~\ref{sec:explore} will
utilize the decoding from $\aerep$ to $\xrec$ to visualize both $\modelrep$ and
$\modelinv$. The autoencoder is trained to reconstruct its inputs by minimizing
a perceptual metric between input and reconstruction, $\Vert \x - \xrec
\Vert$, as in \cite{dosovitskiy2016generating}. The details of the architecture and training
procedure can be found in Sec.~\ref{subsec:autoencoder}. It is crucial that the
autoencoder only needs to be trained once on the training data. Consequently,
the same $\encoder$ can be used to interpret different representations
$\modelrep$, \eg different models or layers within a model, thus
ensuring fair comparisons between them. Moreover, the complexity of the
autoencoder can be adjusted based on the computational needs, allowing us to
work with much lower dimensional encodings $\aerep$ compared to reconstructing
the invariances directly from the images $\x$. This reduces the computational
demands of our approach significantly.

\subsubsection*{Learning a Conditional INN that Recovers Invariances}
Due to the reconstruction task of the autoencoder, $\aerep$ not only contains
the invariances $\modelinv$, but also the representation $\modelrep$. Thus, we
must disentangle
\cite{Esser_2019,li2019mixnmatch,Kotovenko2019ContentAS}
$\modelinv$ and $\modelrep$ using a mapping $\condinn(\cdot \vert \modelrep):
\aerep \mapsto \modelinv = \condinn(\aerep \vert \modelrep)$ which, depending
on $\modelrep$, extracts $\modelinv$ from $\aerep$.

Besides extracting the invariances from a given $\aerep$, $\condinn$ must also
enable an inverse mapping from given model representations $\modelrep$ to
$\aerep$ to support a further mapping onto semantic concepts
(Sec.~\ref{sec:semantic}) and visualization based on $\decoder(\aerep)$. There
are many different $\x$ with $\modelphi(\x)  = \modelrep$, namely all those
$\x$ which differ only in properties that $\modelphi$ is invariant to. Thus,
there are also many different $\aerep$ that this mapping must recover.
Consequently, the mapping from $\modelrep$ to $\aerep$ is set-valued. However,
to understand $\blackbox$ we do not want to recover all possible $\aerep$, but
only those which are likely under the training distribution of the autoencoder.
In particular, this excludes unnatural images such as those obtained by
DeepDream \cite{mordvintsev2015inceptionism}, or adversarial attacks
\cite{szegedy2013intriguing}. In conclusion, we need to sample $\aerep \sim
p(\aerep \vert \modelrep)$.

To avoid a costly inversion process of $\modelphi$, $\condinn$ must be
invertible (implemented as an INN) so that a change of variables
\begin{equation}
  p(\aerep \vert \modelrep)
  = \frac{p(\modelinv \vert \modelrep)}
  {\vert \det \nabla (t^{-1})(\modelinv \vert \modelrep) \vert}
  \quad \text{where } \modelinv = \condinn(\aerep \vert \modelrep)
\end{equation}
yields $p(\aerep \vert \modelrep)$ by means of the distribution $p(\modelinv
\vert \modelrep)$ of invariances, given a model representation $\modelrep$.
Here, the denominator denotes the absolute  value of the determinant of
Jacobian $\nabla (t^{-1})$ of $\modelinv \mapsto t^{-1}(\modelinv \vert
\modelrep)=\aerep$, which is efficient to compute for common invertible
network architectures.
Consequently, we obtain $\aerep$ for given $\modelrep$ by sampling from the
invariant space $\modelinv$ given $\modelrep$ and then applying $\condinn^{-1}$,
    \begin{equation}
      \aerep \sim p(\aerep \vert \modelrep)
      \quad \iff \quad
      \modelinv \sim p(\modelinv \vert \modelrep)
      \label{eq:sample}
      ,
      \aerep = \condinn^{-1}(\modelinv \vert \modelrep) .
    \end{equation}
Since $\modelinv$ is the invariant space for $\modelrep$, both are
complementary thus implying independence $p(\modelinv \vert \modelrep) =
p(\modelinv)$.  Because a powerful transformation $\condinn^{-1}$ can transform
between two arbitrary densities, we can assume without loss of generality a
Gaussian prior $p(\modelinv) = \normaldistr(\modelinv \vert 0, \id)$. Given
this prior, our task is then to learn the transformation $\condinn$ that maps
$\normaldistr(\modelinv \vert 0, \id)$ onto $p(\aerep \vert \modelrep)$. To
this end, we maximize the log-likelihood of $\aerep$ given $\modelrep$,
which results in a per-example loss of
\begin{equation}
  \ell(\aerep, \modelrep) = -\log p(\aerep \vert \modelrep) = -\log \normaldistr(\condinn(\aerep\vert \modelrep)) - \log
  \vert \det \nabla t(\aerep \vert \modelrep) \vert .
\end{equation}
Minimizing this loss over the training data distribution $p(x)$ gives
$\condinn$, a bijective mapping between $\aerep$ and ($\modelrep, \modelinv $),
\begin{align}
  \loss(\condinn)
  &= \mathbb{E}_{x\sim p(x)} \left[\ell(\encoder(\x), \modelphi(\x)) \right]\\
  & = \mathbb{E}_{x\sim p(x)} \left[\frac{1}{2}\Vert \condinn(\encoder(\x) \vert \modelphi(\x))\Vert^2
  + \aerepdim \log 2\pi
  - \log \vert \det \nabla t(\encoder(\x) \vert \modelphi(\x)) \vert\right]
\label{eq:cinnloss}
\end{align}
Note that both $\encoder$ and $\modelphi$ remain fixed during minimization of $\loss$.

\subsection{Interpreting Representations and Their Invariances}
\label{sec:revealingsemantics}
\textbf{Visualizing Representations and Invariances}
For an image representation $\modelrep = \modelphi(\x)$, Eq.~\eqref{eq:sample}
presents an efficient approach (a single forward pass through the INN
$\condinn$) to sample an encoding $\aerep$, which is a combination of
$\modelrep$ with a particular realization of its invariances $\modelinv$.
Sampling multiple realizations of $\aerep$ for a given $\modelrep$ highlights
what remains constant and what changes due to different $\modelinv$:
information preserved in the representation $\modelrep$ remains constant over
different samples and information discarded by the model ends up in the
invariances $\modelinv$ and shows changes over different samples. Visualizing
the samples $\aerep \sim p(\aerep \vert \modelrep)$ with $\xrec =
\decoder(\aerep)$ portrays this constancy and changes due to different
$\modelinv$. To complement this visualization, in the following, we learn a
transformation of $\aerep$ into a semantically meaningful representation which
allows to uncover the semantics captured by $\modelrep$ and $\modelinv$.

\subsubsection*{Learning an INN to Produce Semantic
Interpretations}
\label{sec:semantic}
The autoencoder representation $\aerep$ is an equivalent representation of
$(\modelrep, \modelinv)$ but its feature dimensions do not necessarily
correspond to semantic concepts \cite{fong2018net2vec}. More generally, without
supervision, we cannot reliably discover semantically meaningful, explanatory
factors of $\aerep$ \cite{locatello2018challenging}.
In order to explain $\aerep$ in terms of given semantic concepts, we apply the approach of \cite{esser2020disentangling} and learn a bijective transformation of $\aerep$ to an interpretable representation $\semanticinn(\aerep)$ where different groups of components, called factors, correspond to semantic concepts. 

To learn the transformation $\semanticinn$, we parameterize $\semanticinn$ by an INN and assume
that semantic concepts are defined implicitly by pairs of images, \ie for each
semantic concept we have access to training pairs $\x^a, \x^b$ that have the
respective concept in common. For example, the semantic concept `smiling' is
defined by pairs of images, where either both images show smiling persons or
both images show non-smiling persons. Applying this formulation, input pairs which are similar in a certain semantic concept are similar in the corresponding factor of the interpretable representation $\semanticinn(\aerep)$.

Following \cite{esser2020disentangling}, the loss for training the invertible network $\semanticinn$ is then given by 
 \begin{align}
    \loss(\semanticinn)
    = \mathbb{E}_{x^a, x^b} 
      &\left[-\log p(\semanticinn(\encoder(x^a)), \semanticinn(\encoder(x^b))) \right. \nonumber \\
      & \left. -\log \vert \det \nabla \semanticinn(\encoder(x^a)) \vert
      -\log \vert \det \nabla \semanticinn(\encoder(x^b)) \vert \right].
  \end{align}
Further details regarding the application of this approach within our setting can be found in the supplementary, Sec.~\ref{subsec:suppsemanticinn}.

\subsubsection*{Interpretation by Applying the Learned INNs}
After training, the combination of $\semanticinn$ with $\condinn$ from
Sec.~\ref{sec:recover} provides semantic interpretations given a model
representation $\modelrep$: Eq.~\eqref{eq:sample} gives realizations of the invariances
$\modelinv$ which are combined with $\modelrep$ to produce $\aerep =
\condinn^{-1}(\modelinv \vert \modelrep)$. Then $\semanticinn$ transforms
$\aerep$ without loss of information into a semantically accessible
representation $(\semantic_i)_i = \semanticinn(\aerep) = \semanticinn
(\condinn^{-1}(\modelinv \vert \modelrep))$ consisting of different semantic
factors $\semantic_i$. Comparing the $\semantic_i$ for different model
representations $\modelrep$ and invariances $\modelinv$ allows us to observe
which semantic concepts the model representation $\modelrep=\modelphi(\cdot)$
is sensitive to, and which it is invariant to.

\subsubsection{Semantic Modifications of Latent Representations}
\label{sec:explore}
$\condinn^{-1}$ and $\semanticinn$ not only interpret a representation
$\modelrep$ in terms of accessible semantic concepts $(\semantic_i)_i$. Given
$\modelinv \sim p(\modelinv)$, they also allow to modify
$\aerep=\condinn^{-1}(\modelinv \vert \modelrep)$ in a semantically meaningful
manner by altering its corresponding $(\semantic_i)_i$ and then applying the
inverse translation $\semanticinn^{-1}$,
\begin{equation}
  \aerep
  \xrightarrow{\semanticinn}
  (\semantic_i)
  \xrightarrow{\text{modification}}
  (\semantic_i^{*})
  \xrightarrow{\semanticinn^{-1}}
  \aerep^{*}
\end{equation}
The modified representation $\aerep^{*}$  is then readily transformed back into
image space $\xrec^{*} = \decoder(\aerep^{*})$. Besides visual interpretation
of the modification, $\xrec^{*}$ can be fed into the model
$\modelpsi(\modelphi(\xrec^{*}))$ to probe for sensitivity to certain semantic
concepts.

\section{Experiments}
To explore the applicability of our approach, we conduct experiments on several
models which we aim to understand: \emph{SqueezeNet}
\cite{iandola2016squeezenet}, which provides lightweight classification,
\emph{FaceNet} \cite{schroff2015facenet}, a baseline for face recognition and
clustering, trained on the \emph{VGGFace2 dataset} \cite{cao2018vggface2}, and
variants \emph{ResNet} \cite{he2016deep}, a popular architecture, often used when finetuning a classifier on a specific task and dataset.

Experiments are conducted on the following datasets: \textsl{CelebA}
\cite{liu2015faceattributes}, \textsl{AnimalFaces} \cite{liu2019few},
\textsl{Animals} (containing carnivorous animals, see Sec.~\ref{supp:comparison}), \textsl{ImageNet} \cite{deng2009imagenet} and \textsl{ColorMNIST}, which is an augmented version of the \textsl{MNIST}
dataset \cite{lecun1998mnist}, where both background and foreground
have random, independent colors.

\subsection{Comparison to Existing Methods}
\label{sec:compare}
A key insight of our work is that reconstructions from a given model's
representation $\modelrep = \modelphi(\x)$ are impossible if the invariances
the model has learned are not considered. In Fig.~\ref{fig:comparison} we
compare to existing methods that either try to reconstruct the image via
gradient-based optimization \cite{mahendran2016visualizing} or by training a
reconstruction network directly on the representations $\modelrep$
\cite{dosovitskiy2016generating}.
\comparison
By conditionally sampling images $\xrec = \decoder(\aerep)$, where we obtain
$\aerep$ via the INN $\condinn$ as described in Eq.~\eqref{eq:sample} based on
the invariances $\modelinv \sim p(\modelinv) =
\mathcal{N}(0, \id)$, we bypass this shortcoming and obtain natural images
without artifacts for any layer depth. The increased image quality is further
confirmed by the FID scores reported in Tab.~\ref{tab:comparefidsbrox}.
\comparefidsbrox
\subsection{Understanding Models}
\label{sec:understandingmodels}
\subsubsection{Interpreting a Face Recognition Model}
\label{subsubsec:noninvertiblenet}
\emph{FaceNet} \cite{schroff2015facenet} is a widely accepted baseline in the field of
face recognition. This model embeds input images of human faces into a latent
space where similar images have a small $L_2$-distance. We aim to understand
the process of face recognition within this model by analyzing and visualizing
learned invariances for several layers explicitly; see
Tab.~\ref{tab:interpretfacenet} for a
detailed breakdown of the various layers of \emph{FaceNet}.  For the experiment, we
use a pretrained \emph{FaceNet} and train the generative model presented in
Eq.~\eqref{eq:sample} by conditioning on various layers.  \facenetlayersdecoded
Fig.~\ref{fig:facenetlayersdecoded} depicts the amount of variance present in
each selected layer when generating $n=250$ samples for each of 100 different
input images. This variance serves as a proxy for the amount of
abstraction capability \emph{FaceNet} has learned in its respective layers: More
abstract representations allow for a rich variety of corresponding synthesized
images, which results in a large variance in image space when being decoded.
We observe an approximate exponential growth of learned
invariances with increasing layer depth, suggesting that abstraction mainly
happens in the deepest layers of the network. Furthermore, we are able to
synthesize images that correspond to the given model representation for each
selected layer. 
\subsubsection{How Does Relevance of Different Concepts Emerge During Training?}
\label{subsec:factorevo}
\factorevolution
Humans tend to provide explanations of entities by describing them in terms of
their semantics, e.g. size or color. In a similar fashion, we want to
semantically understand how a network (here: \emph{SqueezeNet}) learns to solve a
given problem. \\ Intuitively, a network should for example be able to solve a
given classification problem by focusing on the relevant information while
discarding task-irrelevant information.  To build on this intuition, we
construct a toy problem: Digit classification on ColorMNIST. We expect the
model to ignore both the random back- and foreground color of the input data,
as it does not help making a classification decision.  Thus, we
apply the invertible approach presented in Sec.~\ref{sec:revealingsemantics}
and recover three distinct factors: \textsl{digit class}, \textsl{background
color} and \textsl{foreground color}. To capture the semantic
changes occuring over the course of training of this classifier, we couple 20
instances of the invertible interpretation model on the last convolutional
layer, each representing a checkpoint between iteration 0 and iteration 40000
(equally distributed).
The result is shown in Fig.~\ref{fig:factorevolution}: We see that the
\textsl{digit} factor becomes increasingly more relevant, with its relevance
being strongly correlated to the accuracy of the model.
\subsection{Effects of Data Shifts on Models}
\label{sec:understandinputs}
This section investigates the effects that altering the input data has on the
model we want to understand. We examine these effects by manipulating the
input data explicitly through adversarial attacks or image stylization.
\subsubsection{How Do Adversarial Attacks Affect Network Representations?}
\advattacklayers
Here, we experiment with
\textsl{Fast Gradient Sign} (FGSM) attacks \cite{goodfellow2014explaining},
which manipulate the input image by maximizing the objective of a given
classification model. To understand how such an attack modifies representations
of a given model, we first compute the image's invariances with respect to the
model as $\modelinv = \condinn(\encoder(\x) \vert \modelphi(\x))$. For an
attacked image $\x^{*}$, we then compute the attacked representation as
$\modelrep^{*}=\modelphi(\x^{*})$. Decoding this representation with the
original invariance $\modelinv$, allows us to precisely visualize what the
adversarial attack changed. This decoding, $\xrec^{*} =
\decoder(\condinn(\modelinv \vert \modelrep^{*}))$, is shown in
Fig.~\ref{fig:attackanimalslayers}. We observe that, over layers of the
network, the adversarial attack gradually changes the representation towards
its target. Its ability to do so is strongly correlated with the amount of
invariances, quantified as the total variance explained by $\modelinv$ (see
Sec.~\ref{supp:explainedvar}), for a given layer as also observed in
\cite{jacobsen2018excessive}. For additional examples, see Fig.~\ref{fig:extraattacks}.

\texturebias
\subsubsection{How Does Training on Different Data Affect the Model?}
\cite{geirhos2018imagenet} proposed the hypothesis that classification networks
based on convolutional blocks mainly focus on texture patterns to obtain class
probabilities. We further validate this hypothesis by training our invertible
network $\condinn$ conditioned on pre-logits $\modelrep = \modelphi(\x)$ (\ie
the penultimate layer) of two ResNet-50 realizations.  As shown in
Fig.~\ref{fig:texturebias}, a ResNet architecture trained on standard ImageNet
is susceptible to the so-called "texture-bias", as samples generated
conditioned on representation of pure texture images consistently show valid
images of corresponding input classes. We furthermore visualize that this
behavior can indeed be removed by training the same architecture on a stylized
version of ImageNet \footnote{we used weights available at
\url{https://github.com/rgeirhos/texture-vs-shape}}; the classifier does focus
on shape. Rows 10-12 of Fig.~\ref{fig:texturebias} show that the
proposed approach can be used to generate sketch-based content with the
texture-agnostic network.
\subsection{Modifying Representations}
\label{sec:modrep}
Invertible access to semantic concepts enables targeted modifications of
representations $\aerep$. In combination with a decoder for $\aerep$, we obtain
semantic image editing capabilities.
We provide an example in
Fig.~\ref{fig:facemod}, where we modify the factors hair color, glasses,
gender, beard, age and smile.
We infer $\aerep=\encoder(\x)$ from an input image. Our semantic
INN $\semanticinn$ then translates this representation into semantic factors 
$(\semantic_i)_i=\semanticinn(\aerep)$, where individual semantic concepts can
be modified independently via the corresponding factor $\semantic_i$.
In particular, we can replace each factor with that from another image,
effectively transferring semantics from one representation onto another. Due to
the invertibility of $\semanticinn$, the modified representation can
be translated back into the space of the autoencoder and is readily decoded to
a modified image $x^{*}$. Additional examples can be found in
Sec.~\ref{supp:modrep}.

To observe which semantic concepts \emph{FaceNet} is sensitive to, we compute the average
distance $\Vert f(\x) - f(\x^{*})\Vert$ between its embeddings of $x$ and
semantically modified $x^{*}$ over the test set (last row in Fig.~\ref{fig:facemod}).
Evidently,
FaceNet is particularly sensitive to differences in gender and glasses. The
latter suggests a failure of FaceNet to identify persons correctly after they
put on glasses.
\facemod
\section{Conclusion}
Understanding a representation in terms of both its semantics and learned
invariances is crucial for interpretation of deep networks. We presented an
approach to \emph{(i)} recover the invariances a model has learned and
\emph{(ii)} translate the representation and its invariances onto an equally
expressive yet semantically accessible encoding. Our diagnostic method is
applicable in a plug-and-play fashion on top of existing deep models with no
need to alter or retrain them.
Since our translation onto semantic factors is bijective, it loses no
information and also allows for semantic modifications.
Moreover,
recovering invariances probabilistically
guarantees that we can correctly visualize representations and
sample them without leaving the underlying distribution,
which is a common cause for artifacts. Altogether, our approach constitutes a
powerful, widely applicable diagnostic pipeline for explaining deep
representations.

\section*{Acknowledgments}
This work has been supported in part by the German Research
Foundation (DFG) projects 371923335, 421703927, and EXC 2181/1 - 390900948 and
the German federal ministry BMWi within the project ``KI Absicherung''.

\FloatBarrier
\clearpage

\input{supplementary}

\FloatBarrier
\clearpage

\bibliographystyle{splncs04}
\bibliography{ms}
\end{document}

%% file: supplementary.tex
\appendix
\renewcommand{\thefigure}{S\arabic{figure}}
\setcounter{figure}{0}
\renewcommand{\thetable}{S\arabic{table}}
\setcounter{table}{0}
\providecommand{\impath}[1]{}
\providecommand{\starpath}[1]{}
\providecommand{\imwidth}{}

\newcommand{\domainshiftfacenet}{
\begin{figure}[htb]
\centering
\includegraphics[width=0.9\textwidth]{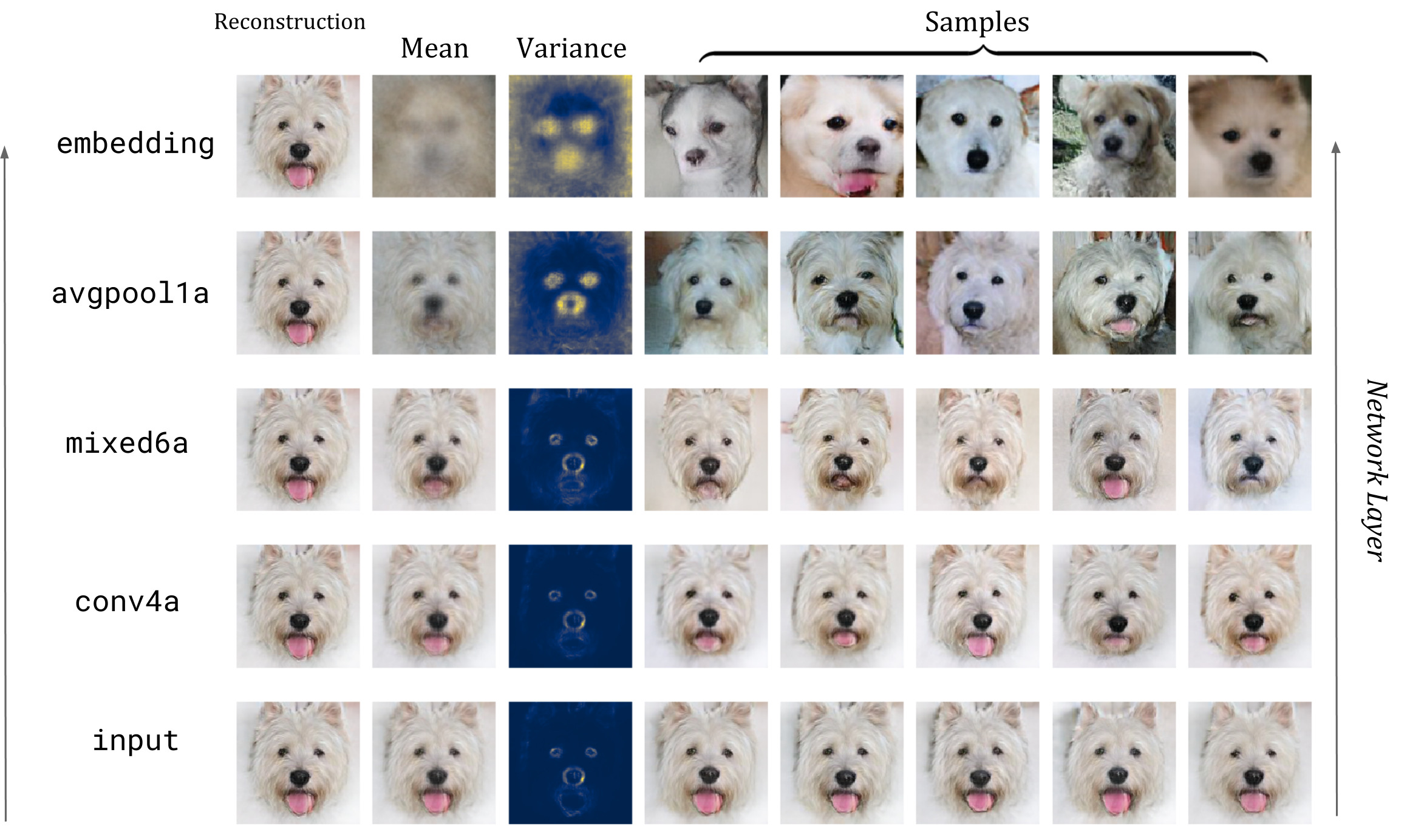}
\caption{
Shifting domains: Human faces to animal faces evaluated with a fixed \emph{FaceNet}.
  The evaluation procedure is similar to the method
  described in Fig.~\ref{fig:facenetlayersdecoded}. Although never trained on
  data consisting of something else than human faces, \emph{FaceNet} is able to
  capture the "identity" of the input to a certain degree. Information about
  appearance is approximately preserved until the last layer, \ie the final identity embedding.}
  \label{fig:facenetdomainshift} \end{figure}
}

\newcommand{\extraattacksfig}{
\begin{figure}[!t]
\centering
\includegraphics[width=0.999\textwidth]{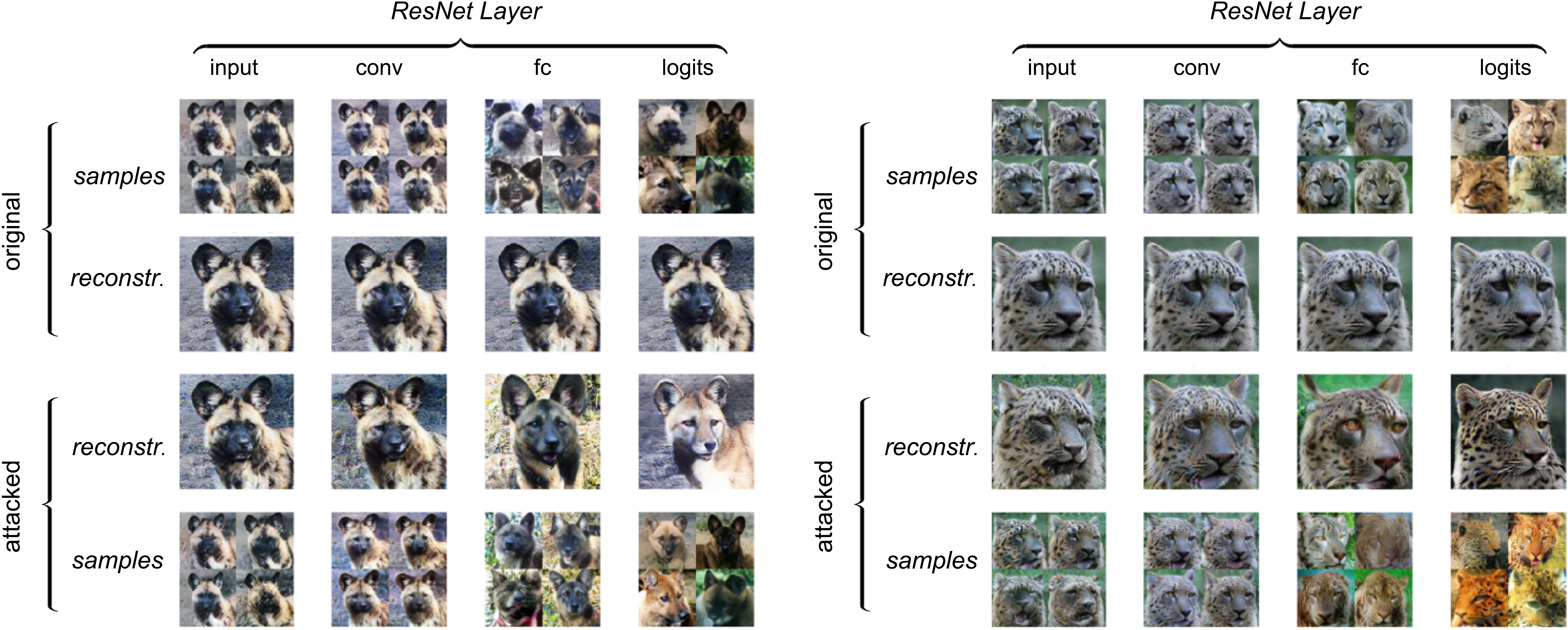}
\phantom{aaa}
\includegraphics[width=0.999\textwidth]{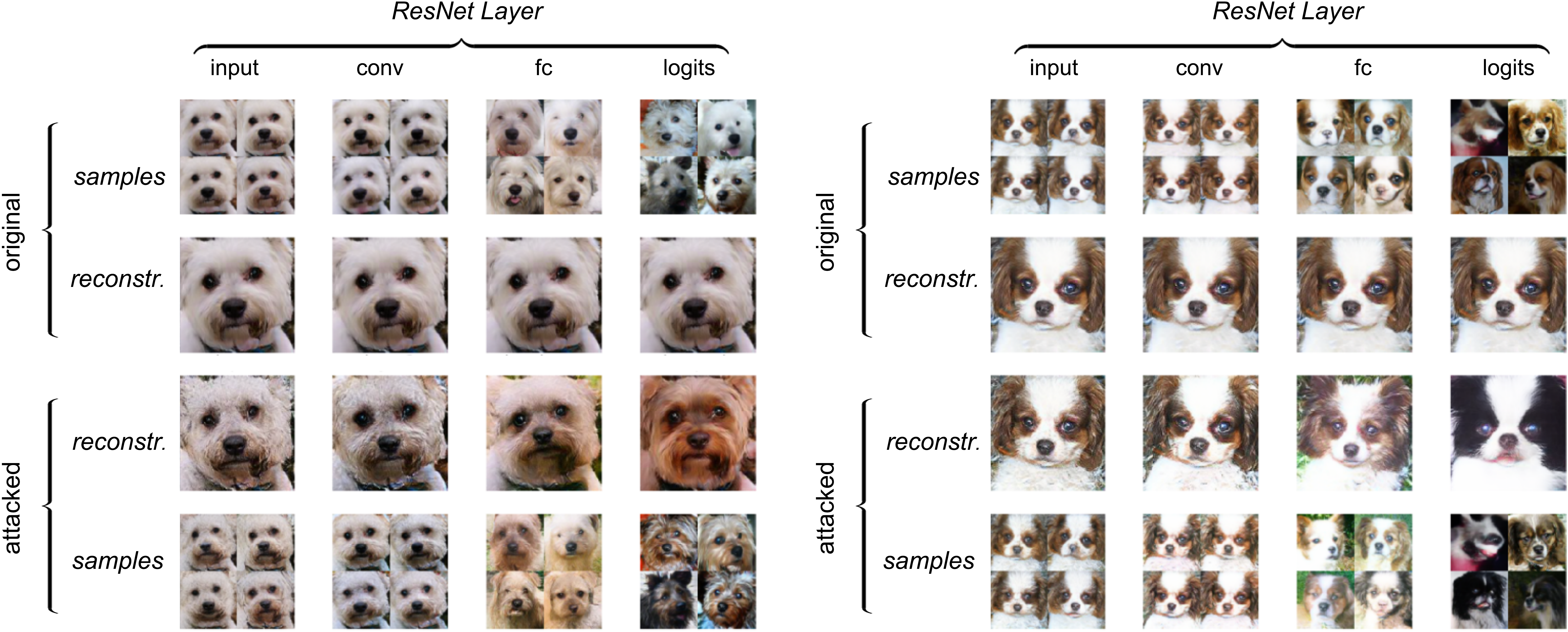}
\caption{More visualizations of adversarial attacks as in Fig.~\ref{fig:attackanimalslayers}. Predictions of original \vs attacked version of the input image for all depicted examples: \emph{top left:} `Lycaon pictus' \vs `Cuon alpinus'; \emph{top right:} `Snow Leopard' \vs `Leopard'; \emph{bottom left:} `West Highland white Terrier' \vs `Yorkshire Terrier'; \emph{bottom right:} `Blenheim Spaniel' \vs `Japanese Spaniel'.}
\label{fig:extraattacks}
\end{figure}
}

\newcommand{\interpretedarchitectures}{
\begin{table*}[ht]
  \caption{\label{tab:interpretmodels} High-level architectures of FaceNet and
  ResNet, depicted as \texttt{pytorch}-modules. Layers investigated in our
  experiments are marked in bold. Spatial sizes are provided as a visual aid
  and vary from model to model in our experiments. If not stated otherwise, we
  always extract from the \emph{last} layer in a series of blocks (\eg in
  Tab.~\ref{tab:interpretresnet}: $23\times$ \textbf{BottleNeck down} $\to
  \mathbb{R}^{8 \times 8 \times 1024}$ refers to the last module in the series
  of 23 blocks.)}
  \centering
  \parbox[b]{0.45\textwidth}{
    \centering
    \caption{\label{tab:interpretfacenet} \textsl{FaceNet:} Implementations
    of layers \texttt{Mixed, Block35, Block17, Block8} can be found at
    \protect\url{https://github.com/timesler/facenet-pytorch}. In l.4, the
    representation from the \textbf{2nd} convolutional layer is extraced.
    Furthermore, BN refers to batch normalization.}
    \begin{tabular}[t]{c}
      \toprule
      RGB image $x\in \mathbb{R}^{128 \times 128 \times 3}$ \\
      \midrule
      $3 \times $ Conv, BN, ReLU $\to \mathbb{R}^{61\times 61 \times 64}$\\
      \midrule
      MaxPool $\to \mathbb{R}^{30\times 30 \times 64}$\\
      \midrule
      $3 \times $ \textbf{Conv}, BN, ReLU $\to \mathbb{R}^{13\times 13 \times 256}$\\
      \midrule
      $5 \times$ Block35 $\to \mathbb{R}^{13\times 13 \times 256}$\\
      \midrule
      \textbf{Mixed down} $\to \mathbb{R}^{6\times 6 \times 896}$\\
      \midrule
      $10 \times$ Block17 $\to \mathbb{R}^{6\times 6 \times 896}$\\
      \midrule
      Mixed down $\to \mathbb{R}^{2\times 2 \times 1792}$\\
      \midrule
      $5 \times$ Block8 $\to \mathbb{R}^{2\times 2 \times 1792}$\\
      \midrule
      \textbf{AdaAvgPool}  $\to \mathbb{R}^{1\times 1 \times 1792}$\\
      \midrule
      Dropout, Linear, BN $\to \mathbb{R}^{512}$ \\
      \midrule
      \textbf{identity embedding} $\to \mathbb{R}^{512}$ \\
      \bottomrule
    \end{tabular}
  }
  \hspace{1em}
  \parbox[b]{0.45\textwidth}{
    \centering
    \caption{\label{tab:interpretresnet} \textsl{ResNet-101:} See
    \protect\url{https://pytorch.org/docs/stable/torchvision/models.html}		for
    details on other variants of ResNet.}
    \begin{tabular}[t]{c}
      \toprule
      RGB image $x\in \mathbb{R}^{128 \times 128 \times 3}$ \\
      \midrule
      Conv down $\to \mathbb{R}^{64\times 64 \times 64}$\\
      \midrule
      Norm, ReLU, \textbf{MaxPool} $\to \mathbb{R}^{32 \times 32 \times 64}$ \\
      \midrule
      $3\times$ BottleNeck $\to \mathbb{R}^{32 \times 32 \times 256}$ \\
      \midrule
      $4\times$ BottleNeck down $\to \mathbb{R}^{16 \times 16 \times 512}$ \\
      \midrule
      $23\times$ \textbf{BottleNeck down} $\to \mathbb{R}^{8 \times 8 \times 1024}$ \\
      \midrule
      $3\times$ BottleNeck down $\to \mathbb{R}^{4 \times 4 \times 2048}$ \\
      \midrule
      \textbf{AvgPool}, FC \\
      \midrule
      \textbf{output} $\to \mathbb{R}^{1000}$ \\
      \bottomrule
    \end{tabular}
  }
\end{table*}
}

\newcommand{\interpretedarchitecturestwo}{
\begin{table}[ht]
  \caption{\label{tab:interpretmodelstwo} High-level architectures of
  SquuezeNet and AlexNet, depicted as \texttt{pytorch}-modules. \Cf
  Tab.\ref{tab:interpretmodels} for further details.}
  \centering
  \parbox[b]{.45\textwidth}{
    \centering
    \caption{\label{tab:interpretsqueezenet}
    \textsl{SqueezeNet}. We extract the penultimate \textsl{Fire} block for
    interpretation in Sec.~\ref{sec:understandingmodels}.}
    \begin{tabular}[t]{c}
      \toprule
      RGB image $x\in \mathbb{R}^{128 \times 128 \times 3}$ \\
      \midrule
      Conv, ReLU, MaxPool $\to \mathbb{R}^{31\times 31 \times 64}$\\
      \midrule
      $2\times$ Fire $\to \mathbb{R}^{31 \times 31 \times 128}$ \\
      \midrule
      MaxPool $\to \mathbb{R}^{15 \times 15 \times 128}$ \\
      \midrule
      $2\times$ Fire $\to \mathbb{R}^{15 \times 15 \times 256}$ \\
      \midrule
      MaxPool $\to \mathbb{R}^{7 \times 7 \times 256}$ \\
      \midrule
      \textbf{$4\times$ Fire} $\to \mathbb{R}^{7 \times 7 \times 512}$ \\
      \midrule
      Dropout, Conv, ReLU $\to \mathbb{R}^{7 \times 7 \times 1000}$ \\
      \midrule
      AdaAvgPool $\to \mathbb{R}^{7 \times 7 \times 1000}$ \\
      \midrule
      output $\to \mathbb{R}^{1000}$ \\
      \bottomrule
    \end{tabular}
  }
  \hspace{1em}
  \parbox[b]{.45\textwidth}{
    \centering
    \caption{\label{tab:interpretalexnet} \textsl{AlexNet:} The first
    convolution uses kernel size 11.}
    \begin{tabular}[t]{c}
      \toprule
      RGB image $x\in \mathbb{R}^{128 \times 128 \times 3}$ \\
      \midrule
      Conv, ReLU, MaxPool $\to \mathbb{R}^{15\times 15 \times 64}$\\
      \midrule
      Conv, ReLU, MaxPool $\to \mathbb{R}^{7\times 7 \times 192}$\\
      \midrule
      Conv, ReLU $\to \mathbb{R}^{7 \times 7 \times 384}$ \\
      \midrule
      $2 \times$ \textbf{Conv}, ReLU $\to \mathbb{R}^{7 \times 7 \times 256}$ \\
      \midrule
      MaxPool $\to \mathbb{R}^{3 \times 3 \times 256}$ \\
      \midrule
      AdaAvgPool, Flatten $\to \mathbb{R}^{9216}$ \\
      \midrule
      Dropout, \textbf{Linear}, ReLU $\to \mathbb{R}^{4096}$ \\
      \midrule
      Dropout, \textbf{Linear}, ReLU $\to \mathbb{R}^{4096}$ \\
      \midrule
      \textbf{Linear} $\to \mathbb{R}^{1000}$ \\
      \bottomrule
    \end{tabular}
  }
\end{table}
}

\newcommand{\suppcelebamod}{
\begin{figure}[htb]
  \renewcommand{\impath}[1]{supp/celebamod/##1}
  \renewcommand{\starpath}[1]{supp/starganmod/##1}
  \renewcommand{\imwidth}{0.11\textwidth}
  \centering
  \begin{tabular}{cccccccc}
    input &
    \multirow{2}{*}{method} &
    hair &
    glasses &
    gender &
    beard &
    age &
    smiling\\ 
    $\x$ &
    &
    $\semantic_1$ &
    $\semantic_2$ &
    $\semantic_3$ &
    $\semantic_4$ &
    $\semantic_5$ &
    $\semantic_6$ \\ 
    \midrule

    \multirow{2}{*}[-0.5em]{\includegraphics[width=\imwidth, align=c]{\impath{celebamod_001_000.jpg}}} & 
    our &
    \includegraphics[width=\imwidth, align=c]{\impath{celebamod_001_001.jpg}} & 
    \includegraphics[width=\imwidth, align=c]{\impath{celebamod_001_002.jpg}} & 
    \includegraphics[width=\imwidth, align=c]{\impath{celebamod_001_003.jpg}} & 
    \includegraphics[width=\imwidth, align=c]{\impath{celebamod_001_004.jpg}} & 
    \includegraphics[width=\imwidth, align=c]{\impath{celebamod_001_005.jpg}} & 
    \includegraphics[width=\imwidth, align=c]{\impath{celebamod_001_006.jpg}} \\
    & 
    \cite{choi2018stargan} &
    \includegraphics[width=\imwidth, align=c]{\starpath{stargan-10.jpg}} & 
    \includegraphics[width=\imwidth, align=c]{\starpath{stargan-13.jpg}} & 
    \includegraphics[width=\imwidth, align=c]{\starpath{stargan-14.jpg}} & 
    \includegraphics[width=\imwidth, align=c]{\starpath{stargan-15.jpg}} & 
    \includegraphics[width=\imwidth, align=c]{\starpath{stargan-16.jpg}} & 
    \includegraphics[width=\imwidth, align=c]{\starpath{stargan-17.jpg}} \\
    \midrule

    \multirow{2}{*}[-0.5em]{\includegraphics[width=\imwidth, align=c]{\impath{celebamod_002_000.jpg}}} & 
    our &
    \includegraphics[width=\imwidth, align=c]{\impath{celebamod_002_001.jpg}} & 
    \includegraphics[width=\imwidth, align=c]{\impath{celebamod_002_002.jpg}} & 
    \includegraphics[width=\imwidth, align=c]{\impath{celebamod_002_003.jpg}} & 
    \includegraphics[width=\imwidth, align=c]{\impath{celebamod_002_004.jpg}} & 
    \includegraphics[width=\imwidth, align=c]{\impath{celebamod_002_005.jpg}} & 
    \includegraphics[width=\imwidth, align=c]{\impath{celebamod_002_006.jpg}} \\
    & 
    \cite{choi2018stargan} &
    \includegraphics[width=\imwidth, align=c]{\starpath{stargan-73.jpg}} & 
    \includegraphics[width=\imwidth, align=c]{\starpath{stargan-76.jpg}} & 
    \includegraphics[width=\imwidth, align=c]{\starpath{stargan-77.jpg}} & 
    \includegraphics[width=\imwidth, align=c]{\starpath{stargan-78.jpg}} & 
    \includegraphics[width=\imwidth, align=c]{\starpath{stargan-79.jpg}} & 
    \includegraphics[width=\imwidth, align=c]{\starpath{stargan-80.jpg}} \\
    \midrule

    \multirow{2}{*}[-0.5em]{\includegraphics[width=\imwidth, align=c]{\impath{celebamod_003_000.jpg}}} & 
    our &
    \includegraphics[width=\imwidth, align=c]{\impath{celebamod_003_001.jpg}} & 
    \includegraphics[width=\imwidth, align=c]{\impath{celebamod_003_002.jpg}} & 
    \includegraphics[width=\imwidth, align=c]{\impath{celebamod_003_003.jpg}} & 
    \includegraphics[width=\imwidth, align=c]{\impath{celebamod_003_004.jpg}} & 
    \includegraphics[width=\imwidth, align=c]{\impath{celebamod_003_005.jpg}} & 
    \includegraphics[width=\imwidth, align=c]{\impath{celebamod_003_006.jpg}} \\
    & 
    \cite{choi2018stargan} &
    \includegraphics[width=\imwidth, align=c]{\starpath{stargan-91.jpg}} & 
    \includegraphics[width=\imwidth, align=c]{\starpath{stargan-94.jpg}} & 
    \includegraphics[width=\imwidth, align=c]{\starpath{stargan-95.jpg}} & 
    \includegraphics[width=\imwidth, align=c]{\starpath{stargan-96.jpg}} & 
    \includegraphics[width=\imwidth, align=c]{\starpath{stargan-97.jpg}} & 
    \includegraphics[width=\imwidth, align=c]{\starpath{stargan-98.jpg}} \\
    \midrule

    \multirow{2}{*}[-0.5em]{\includegraphics[width=\imwidth, align=c]{\impath{celebamod_004_000.jpg}}} & 
    our &
    \includegraphics[width=\imwidth, align=c]{\impath{celebamod_004_001.jpg}} & 
    \includegraphics[width=\imwidth, align=c]{\impath{celebamod_004_002.jpg}} & 
    \includegraphics[width=\imwidth, align=c]{\impath{celebamod_004_003.jpg}} & 
    \includegraphics[width=\imwidth, align=c]{\impath{celebamod_004_004.jpg}} & 
    \includegraphics[width=\imwidth, align=c]{\impath{celebamod_004_005.jpg}} & 
    \includegraphics[width=\imwidth, align=c]{\impath{celebamod_004_006.jpg}} \\
    & 
    \cite{choi2018stargan} &
    \includegraphics[width=\imwidth, align=c]{\starpath{stargan-01.jpg}} & 
    \includegraphics[width=\imwidth, align=c]{\starpath{stargan-04.jpg}} & 
    \includegraphics[width=\imwidth, align=c]{\starpath{stargan-05.jpg}} & 
    \includegraphics[width=\imwidth, align=c]{\starpath{stargan-06.jpg}} & 
    \includegraphics[width=\imwidth, align=c]{\starpath{stargan-07.jpg}} & 
    \includegraphics[width=\imwidth, align=c]{\starpath{stargan-08.jpg}} \\
    \midrule

    \multirow{2}{*}[-0.5em]{\includegraphics[width=\imwidth, align=c]{\impath{celebamod_005_000.jpg}}} & 
    our &
    \includegraphics[width=\imwidth, align=c]{\impath{celebamod_005_001.jpg}} & 
    \includegraphics[width=\imwidth, align=c]{\impath{celebamod_005_002.jpg}} & 
    \includegraphics[width=\imwidth, align=c]{\impath{celebamod_005_003.jpg}} & 
    \includegraphics[width=\imwidth, align=c]{\impath{celebamod_005_004.jpg}} & 
    \includegraphics[width=\imwidth, align=c]{\impath{celebamod_005_005.jpg}} & 
    \includegraphics[width=\imwidth, align=c]{\impath{celebamod_005_006.jpg}} \\
    & 
    \cite{choi2018stargan} &
    \includegraphics[width=\imwidth, align=c]{\starpath{stargan-109.jpg}} & 
    \includegraphics[width=\imwidth, align=c]{\starpath{stargan-112.jpg}} & 
    \includegraphics[width=\imwidth, align=c]{\starpath{stargan-113.jpg}} & 
    \includegraphics[width=\imwidth, align=c]{\starpath{stargan-114.jpg}} & 
    \includegraphics[width=\imwidth, align=c]{\starpath{stargan-115.jpg}} & 
    \includegraphics[width=\imwidth, align=c]{\starpath{stargan-116.jpg}} \\
    \midrule

    \multirow{2}{*}[-0.5em]{\includegraphics[width=\imwidth, align=c]{\impath{celebamod_006_000.jpg}}} & 
    our &
    \includegraphics[width=\imwidth, align=c]{\impath{celebamod_006_001.jpg}} & 
    \includegraphics[width=\imwidth, align=c]{\impath{celebamod_006_002.jpg}} & 
    \includegraphics[width=\imwidth, align=c]{\impath{celebamod_006_003.jpg}} & 
    \includegraphics[width=\imwidth, align=c]{\impath{celebamod_006_004.jpg}} & 
    \includegraphics[width=\imwidth, align=c]{\impath{celebamod_006_005.jpg}} & 
    \includegraphics[width=\imwidth, align=c]{\impath{celebamod_006_006.jpg}} \\
    & 
    \cite{choi2018stargan} &
    \includegraphics[width=\imwidth, align=c]{\starpath{stargan-46.jpg}} & 
    \includegraphics[width=\imwidth, align=c]{\starpath{stargan-49.jpg}} & 
    \includegraphics[width=\imwidth, align=c]{\starpath{stargan-50.jpg}} & 
    \includegraphics[width=\imwidth, align=c]{\starpath{stargan-51.jpg}} & 
    \includegraphics[width=\imwidth, align=c]{\starpath{stargan-52.jpg}} & 
    \includegraphics[width=\imwidth, align=c]{\starpath{stargan-53.jpg}} \\
    \midrule

  \end{tabular}
  \caption{
    Additional examples corresponding to Fig.~\ref{fig:facemod}. In each
    column, we replace a semantic factor $\semantic_i(\encoder(\x))$ by
    $\semantic_i^*$, which is obtained from another, randomly
    chosen, image that differs in the corresponding attribute (see
    Sec.~\ref{supp:modrep}). Subsequently we decode a semantically modified
    image using the invertibility of $\semanticinn$ to obtain
    $\xrec^*=\decoder(\semanticinn^{-1}((\semantic_i^*)))$.
    The results of \textsl{StarGAN} \cite{choi2018stargan} are obtained by negating the binary
    value for the column's attribute. FID scores in
    Fig.~\ref{supp:celebamodtwo}.
  }
  \label{supp:celebamod}
\end{figure}
}

\newcommand{\suppcelebamodtwo}{
\begin{figure}[htb]
  \renewcommand{\impath}[1]{supp/celebamod/##1}
  \renewcommand{\starpath}[1]{supp/starganmod/##1}
  \renewcommand{\imwidth}{0.11\textwidth}
  \centering
  \begin{tabular}{cccccccc}
    input &
    \multirow{2}{*}{method} &
    hair &
    glasses &
    gender &
    beard &
    age &
    smiling\\ 
    $\x$ &
    &
    $\semantic_1$ &
    $\semantic_2$ &
    $\semantic_3$ &
    $\semantic_4$ &
    $\semantic_5$ &
    $\semantic_6$ \\ 
    \midrule

    \multirow{2}{*}[-0.5em]{\includegraphics[width=\imwidth, align=c]{\impath{celebamod_007_000.jpg}}} & 
    our &
    \includegraphics[width=\imwidth, align=c]{\impath{celebamod_007_001.jpg}} & 
    \includegraphics[width=\imwidth, align=c]{\impath{celebamod_007_002.jpg}} & 
    \includegraphics[width=\imwidth, align=c]{\impath{celebamod_007_003.jpg}} & 
    \includegraphics[width=\imwidth, align=c]{\impath{celebamod_007_004.jpg}} & 
    \includegraphics[width=\imwidth, align=c]{\impath{celebamod_007_005.jpg}} & 
    \includegraphics[width=\imwidth, align=c]{\impath{celebamod_007_006.jpg}} \\
    & 
    \cite{choi2018stargan} &
    \includegraphics[width=\imwidth, align=c]{\starpath{stargan-37.jpg}} & 
    \includegraphics[width=\imwidth, align=c]{\starpath{stargan-40.jpg}} & 
    \includegraphics[width=\imwidth, align=c]{\starpath{stargan-41.jpg}} & 
    \includegraphics[width=\imwidth, align=c]{\starpath{stargan-42.jpg}} & 
    \includegraphics[width=\imwidth, align=c]{\starpath{stargan-43.jpg}} & 
    \includegraphics[width=\imwidth, align=c]{\starpath{stargan-44.jpg}} \\
    \midrule

    \multirow{2}{*}[-0.5em]{\includegraphics[width=\imwidth, align=c]{\impath{celebamod_010_000.jpg}}} & 
    our &
    \includegraphics[width=\imwidth, align=c]{\impath{celebamod_010_001.jpg}} & 
    \includegraphics[width=\imwidth, align=c]{\impath{celebamod_010_002.jpg}} & 
    \includegraphics[width=\imwidth, align=c]{\impath{celebamod_010_003.jpg}} & 
    \includegraphics[width=\imwidth, align=c]{\impath{celebamod_010_004.jpg}} & 
    \includegraphics[width=\imwidth, align=c]{\impath{celebamod_010_005.jpg}} & 
    \includegraphics[width=\imwidth, align=c]{\impath{celebamod_010_006.jpg}} \\
    & 
    \cite{choi2018stargan} &
    \includegraphics[width=\imwidth, align=c]{\starpath{stargan-19.jpg}} & 
    \includegraphics[width=\imwidth, align=c]{\starpath{stargan-22.jpg}} & 
    \includegraphics[width=\imwidth, align=c]{\starpath{stargan-23.jpg}} & 
    \includegraphics[width=\imwidth, align=c]{\starpath{stargan-24.jpg}} & 
    \includegraphics[width=\imwidth, align=c]{\starpath{stargan-25.jpg}} & 
    \includegraphics[width=\imwidth, align=c]{\starpath{stargan-26.jpg}} \\
    \midrule

    \multirow{2}{*}[-0.5em]{\includegraphics[width=\imwidth, align=c]{\impath{celebamod_011_000.jpg}}} & 
    our &
    \includegraphics[width=\imwidth, align=c]{\impath{celebamod_011_001.jpg}} & 
    \includegraphics[width=\imwidth, align=c]{\impath{celebamod_011_002.jpg}} & 
    \includegraphics[width=\imwidth, align=c]{\impath{celebamod_011_003.jpg}} & 
    \includegraphics[width=\imwidth, align=c]{\impath{celebamod_011_004.jpg}} & 
    \includegraphics[width=\imwidth, align=c]{\impath{celebamod_011_005.jpg}} & 
    \includegraphics[width=\imwidth, align=c]{\impath{celebamod_011_006.jpg}} \\
    & 
    \cite{choi2018stargan} &
    \includegraphics[width=\imwidth, align=c]{\starpath{stargan-29.jpg}} & 
    \includegraphics[width=\imwidth, align=c]{\starpath{stargan-31.jpg}} & 
    \includegraphics[width=\imwidth, align=c]{\starpath{stargan-32.jpg}} & 
    \includegraphics[width=\imwidth, align=c]{\starpath{stargan-33.jpg}} & 
    \includegraphics[width=\imwidth, align=c]{\starpath{stargan-34.jpg}} & 
    \includegraphics[width=\imwidth, align=c]{\starpath{stargan-35.jpg}} \\
    \midrule

    \multirow{2}{*}[-0.5em]{\includegraphics[width=\imwidth, align=c]{\impath{celebamod_012_000.jpg}}} & 
    our &
    \includegraphics[width=\imwidth, align=c]{\impath{celebamod_012_001.jpg}} & 
    \includegraphics[width=\imwidth, align=c]{\impath{celebamod_012_002.jpg}} & 
    \includegraphics[width=\imwidth, align=c]{\impath{celebamod_012_003.jpg}} & 
    \includegraphics[width=\imwidth, align=c]{\impath{celebamod_012_004.jpg}} & 
    \includegraphics[width=\imwidth, align=c]{\impath{celebamod_012_005.jpg}} & 
    \includegraphics[width=\imwidth, align=c]{\impath{celebamod_012_006.jpg}} \\
    & 
    \cite{choi2018stargan} &
    \includegraphics[width=\imwidth, align=c]{\starpath{stargan-82.jpg}} & 
    \includegraphics[width=\imwidth, align=c]{\starpath{stargan-85.jpg}} & 
    \includegraphics[width=\imwidth, align=c]{\starpath{stargan-86.jpg}} & 
    \includegraphics[width=\imwidth, align=c]{\starpath{stargan-87.jpg}} & 
    \includegraphics[width=\imwidth, align=c]{\starpath{stargan-88.jpg}} & 
    \includegraphics[width=\imwidth, align=c]{\starpath{stargan-89.jpg}} \\
    \midrule

    \multirow{2}{*}[-0.5em]{\includegraphics[width=\imwidth, align=c]{\impath{celebamod_013_000.jpg}}} & 
    our &
    \includegraphics[width=\imwidth, align=c]{\impath{celebamod_013_001.jpg}} & 
    \includegraphics[width=\imwidth, align=c]{\impath{celebamod_013_002.jpg}} & 
    \includegraphics[width=\imwidth, align=c]{\impath{celebamod_013_003.jpg}} & 
    \includegraphics[width=\imwidth, align=c]{\impath{celebamod_013_004.jpg}} & 
    \includegraphics[width=\imwidth, align=c]{\impath{celebamod_013_005.jpg}} & 
    \includegraphics[width=\imwidth, align=c]{\impath{celebamod_013_006.jpg}} \\
    & 
    \cite{choi2018stargan} &
    \includegraphics[width=\imwidth, align=c]{\starpath{stargan-65.jpg}} & 
    \includegraphics[width=\imwidth, align=c]{\starpath{stargan-67.jpg}} & 
    \includegraphics[width=\imwidth, align=c]{\starpath{stargan-68.jpg}} & 
    \includegraphics[width=\imwidth, align=c]{\starpath{stargan-69.jpg}} & 
    \includegraphics[width=\imwidth, align=c]{\starpath{stargan-70.jpg}} & 
    \includegraphics[width=\imwidth, align=c]{\starpath{stargan-71.jpg}} \\
    \midrule

    \multirow{2}{*}{FID} & 
    our &
    \textbf{16.24} & 
    \textbf{24.97} & 
    \textbf{15.17} & 
    \textbf{12.84} & 
    \textbf{13.21} & 
    \textbf{12.96} \\
    & 
    \cite{choi2018stargan} &
    20.94 & 
    41.27 & 
    20.04 & 
    19.88 & 
    21.77 & 
    14.47 \\

  \end{tabular}
  \caption{
    Additional examples as in Fig.~\ref{supp:celebamod}. Moreover, the last row
    contains FID scores \cite{heusel2017gans} of semantically modified images obtained by our approach
    and \cite{choi2018stargan}, which shows that our approach consistently
    outperforms \cite{choi2018stargan}.
  }
  \label{supp:celebamodtwo}
\end{figure}
}

\newcommand{\suppcmnistvis}{
\begin{figure}[htb]
  \renewcommand{\impath}[1]{supp/cmnistvis/##1}
  \renewcommand{\imwidth}{0.065\textwidth}
  \centering
  \begin{tabular}{c@{\hskip 1em}ccccc@{\hskip 1em}ccccc}
    input & \multicolumn{10}{c}{Decoded samples $\xrec = \decoder(\condinn^{-1}(\modelinv \vert \modelrep))$} \\
    \midrule
 $\x$ & \multicolumn{5}{c}{4000 training iterations} & \multicolumn{5}{c}{36000 training iterations} \\
    \midrule
    
    \includegraphics[width=\imwidth, align=c]{\impath{4k/xinput_000011.jpg}} & 

    \includegraphics[width=\imwidth, align=c]{\impath{4k/xsample_00_000011.jpg}} & 
    \includegraphics[width=\imwidth, align=c]{\impath{4k/xsample_01_000011.jpg}} & 
    \includegraphics[width=\imwidth, align=c]{\impath{4k/xsample_05_000011.jpg}} & 
    \includegraphics[width=\imwidth, align=c]{\impath{4k/xsample_07_000011.jpg}} & 
    \includegraphics[width=\imwidth, align=c]{\impath{4k/xsample_04_000011.jpg}} &
    
    \includegraphics[width=\imwidth, align=c]{\impath{36k/xsample_00_000011.jpg}} & 
    \includegraphics[width=\imwidth, align=c]{\impath{36k/xsample_08_000011.jpg}} & 
    \includegraphics[width=\imwidth, align=c]{\impath{36k/xsample_02_000011.jpg}} & 
    \includegraphics[width=\imwidth, align=c]{\impath{36k/xsample_07_000011.jpg}} & 
    \includegraphics[width=\imwidth, align=c]{\impath{36k/xsample_04_000011.jpg}} \\
    
     \includegraphics[width=\imwidth, align=c]{\impath{4k/xinput_000024.jpg}} & 

    \includegraphics[width=\imwidth, align=c]{\impath{4k/xsample_00_000024.jpg}} & 
    \includegraphics[width=\imwidth, align=c]{\impath{4k/xsample_01_000024.jpg}} & 
    \includegraphics[width=\imwidth, align=c]{\impath{4k/xsample_05_000024.jpg}} & 
    \includegraphics[width=\imwidth, align=c]{\impath{4k/xsample_07_000024.jpg}} & 
    \includegraphics[width=\imwidth, align=c]{\impath{4k/xsample_04_000024.jpg}} &
    
    \includegraphics[width=\imwidth, align=c]{\impath{36k/xsample_00_000024.jpg}} & 
    \includegraphics[width=\imwidth, align=c]{\impath{36k/xsample_08_000024.jpg}} & 
    \includegraphics[width=\imwidth, align=c]{\impath{36k/xsample_02_000024.jpg}} & 
    \includegraphics[width=\imwidth, align=c]{\impath{36k/xsample_07_000024.jpg}} & 
    \includegraphics[width=\imwidth, align=c]{\impath{36k/xsample_04_000024.jpg}} \\

	 \includegraphics[width=\imwidth, align=c]{\impath{4k/xinput_000066.jpg}} & 

    \includegraphics[width=\imwidth, align=c]{\impath{4k/xsample_00_000066.jpg}} & 
    \includegraphics[width=\imwidth, align=c]{\impath{4k/xsample_01_000066.jpg}} & 
    \includegraphics[width=\imwidth, align=c]{\impath{4k/xsample_05_000066.jpg}} & 
    \includegraphics[width=\imwidth, align=c]{\impath{4k/xsample_07_000066.jpg}} & 
    \includegraphics[width=\imwidth, align=c]{\impath{4k/xsample_04_000066.jpg}} &
    
    \includegraphics[width=\imwidth, align=c]{\impath{36k/xsample_00_000066.jpg}} & 
    \includegraphics[width=\imwidth, align=c]{\impath{36k/xsample_08_000066.jpg}} & 
    \includegraphics[width=\imwidth, align=c]{\impath{36k/xsample_02_000066.jpg}} & 
    \includegraphics[width=\imwidth, align=c]{\impath{36k/xsample_07_000066.jpg}} & 
    \includegraphics[width=\imwidth, align=c]{\impath{36k/xsample_04_000066.jpg}} \\
    
  	 \includegraphics[width=\imwidth, align=c]{\impath{4k/xinput_000099.jpg}} & 

    \includegraphics[width=\imwidth, align=c]{\impath{4k/xsample_00_000099.jpg}} & 
    \includegraphics[width=\imwidth, align=c]{\impath{4k/xsample_01_000099.jpg}} & 
    \includegraphics[width=\imwidth, align=c]{\impath{4k/xsample_05_000099.jpg}} & 
    \includegraphics[width=\imwidth, align=c]{\impath{4k/xsample_07_000099.jpg}} & 
    \includegraphics[width=\imwidth, align=c]{\impath{4k/xsample_04_000099.jpg}} &
    
    \includegraphics[width=\imwidth, align=c]{\impath{36k/xsample_00_000099.jpg}} & 
    \includegraphics[width=\imwidth, align=c]{\impath{36k/xsample_08_000099.jpg}} & 
    \includegraphics[width=\imwidth, align=c]{\impath{36k/xsample_02_000099.jpg}} & 
    \includegraphics[width=\imwidth, align=c]{\impath{36k/xsample_07_000099.jpg}} & 
    \includegraphics[width=\imwidth, align=c]{\impath{36k/xsample_04_000099.jpg}} \\

    	 \includegraphics[width=\imwidth, align=c]{\impath{4k/xinput_000043.jpg}} & 

    \includegraphics[width=\imwidth, align=c]{\impath{4k/xsample_00_000043.jpg}} & 
    \includegraphics[width=\imwidth, align=c]{\impath{4k/xsample_01_000043.jpg}} & 
    \includegraphics[width=\imwidth, align=c]{\impath{4k/xsample_05_000043.jpg}} & 
    \includegraphics[width=\imwidth, align=c]{\impath{4k/xsample_07_000043.jpg}} & 
    \includegraphics[width=\imwidth, align=c]{\impath{4k/xsample_04_000043.jpg}} &
    
    \includegraphics[width=\imwidth, align=c]{\impath{36k/xsample_00_000043.jpg}} & 
    \includegraphics[width=\imwidth, align=c]{\impath{36k/xsample_08_000043.jpg}} & 
    \includegraphics[width=\imwidth, align=c]{\impath{36k/xsample_02_000043.jpg}} & 
    \includegraphics[width=\imwidth, align=c]{\impath{36k/xsample_07_000043.jpg}} & 
    \includegraphics[width=\imwidth, align=c]{\impath{36k/xsample_04_000043.jpg}} \\    
    
    	 \includegraphics[width=\imwidth, align=c]{\impath{4k/xinput_000070.jpg}} & 

    \includegraphics[width=\imwidth, align=c]{\impath{4k/xsample_00_000070.jpg}} & 
    \includegraphics[width=\imwidth, align=c]{\impath{4k/xsample_01_000070.jpg}} & 
    \includegraphics[width=\imwidth, align=c]{\impath{4k/xsample_05_000070.jpg}} & 
    \includegraphics[width=\imwidth, align=c]{\impath{4k/xsample_07_000070.jpg}} & 
    \includegraphics[width=\imwidth, align=c]{\impath{4k/xsample_04_000070.jpg}} &
    
    \includegraphics[width=\imwidth, align=c]{\impath{36k/xsample_00_000070.jpg}} & 
    \includegraphics[width=\imwidth, align=c]{\impath{36k/xsample_08_000070.jpg}} & 
    \includegraphics[width=\imwidth, align=c]{\impath{36k/xsample_02_000070.jpg}} & 
    \includegraphics[width=\imwidth, align=c]{\impath{36k/xsample_07_000070.jpg}} & 
    \includegraphics[width=\imwidth, align=c]{\impath{36k/xsample_04_000070.jpg}} \\

    	 \includegraphics[width=\imwidth, align=c]{\impath{4k/xinput_000085.jpg}} & 

    \includegraphics[width=\imwidth, align=c]{\impath{4k/xsample_00_000085.jpg}} & 
    \includegraphics[width=\imwidth, align=c]{\impath{4k/xsample_01_000085.jpg}} & 
    \includegraphics[width=\imwidth, align=c]{\impath{4k/xsample_05_000085.jpg}} & 
    \includegraphics[width=\imwidth, align=c]{\impath{4k/xsample_07_000085.jpg}} & 
    \includegraphics[width=\imwidth, align=c]{\impath{4k/xsample_04_000085.jpg}} &
    
    \includegraphics[width=\imwidth, align=c]{\impath{36k/xsample_00_000085.jpg}} & 
    \includegraphics[width=\imwidth, align=c]{\impath{36k/xsample_08_000085.jpg}} & 
    \includegraphics[width=\imwidth, align=c]{\impath{36k/xsample_02_000085.jpg}} & 
    \includegraphics[width=\imwidth, align=c]{\impath{36k/xsample_07_000085.jpg}} & 
    \includegraphics[width=\imwidth, align=c]{\impath{36k/xsample_04_000085.jpg}} \\

    	 \includegraphics[width=\imwidth, align=c]{\impath{4k/xinput_000080.jpg}} & 

    \includegraphics[width=\imwidth, align=c]{\impath{4k/xsample_00_000080.jpg}} & 
    \includegraphics[width=\imwidth, align=c]{\impath{4k/xsample_01_000080.jpg}} & 
    \includegraphics[width=\imwidth, align=c]{\impath{4k/xsample_05_000080.jpg}} & 
    \includegraphics[width=\imwidth, align=c]{\impath{4k/xsample_07_000080.jpg}} & 
    \includegraphics[width=\imwidth, align=c]{\impath{4k/xsample_04_000080.jpg}} &
    
    \includegraphics[width=\imwidth, align=c]{\impath{36k/xsample_00_000080.jpg}} & 
    \includegraphics[width=\imwidth, align=c]{\impath{36k/xsample_08_000080.jpg}} & 
    \includegraphics[width=\imwidth, align=c]{\impath{36k/xsample_02_000080.jpg}} & 
    \includegraphics[width=\imwidth, align=c]{\impath{36k/xsample_07_000080.jpg}} & 
    \includegraphics[width=\imwidth, align=c]{\impath{36k/xsample_04_000080.jpg}} \\

    	 \includegraphics[width=\imwidth, align=c]{\impath{4k/xinput_000087.jpg}} & 

    \includegraphics[width=\imwidth, align=c]{\impath{4k/xsample_00_000087.jpg}} & 
    \includegraphics[width=\imwidth, align=c]{\impath{4k/xsample_01_000087.jpg}} & 
    \includegraphics[width=\imwidth, align=c]{\impath{4k/xsample_05_000087.jpg}} & 
    \includegraphics[width=\imwidth, align=c]{\impath{4k/xsample_07_000087.jpg}} & 
    \includegraphics[width=\imwidth, align=c]{\impath{4k/xsample_04_000087.jpg}} &
    
    \includegraphics[width=\imwidth, align=c]{\impath{36k/xsample_00_000087.jpg}} & 
    \includegraphics[width=\imwidth, align=c]{\impath{36k/xsample_08_000087.jpg}} & 
    \includegraphics[width=\imwidth, align=c]{\impath{36k/xsample_02_000087.jpg}} & 
    \includegraphics[width=\imwidth, align=c]{\impath{36k/xsample_07_000087.jpg}} & 
    \includegraphics[width=\imwidth, align=c]{\impath{36k/xsample_04_000087.jpg}} \\
 
  \end{tabular}
  \caption{
    Additional $\modelrep$ conditional samples after 4k and 36k training steps, as in
    Fig.~\ref{fig:factorevolution}. Each row is conditioned on $\modelrep =
    \modelphi(\x)$ and each column is conditioned on a $\modelinv \sim
    \normaldistr(\modelinv \vert 0, \id)$. At 4k (resp. 36k) iterations, $\modelrep$
    explains $2.57\%$ (resp. $36.08\%$) of the variance in the digit factor.
    Thus, the digit class of samples obtained at 4k iterations change with the
    sampled invariances across columns, while it stays the same at 36k
    iterations. Conversely, at 4k (resp. 36k) iterations, $\modelrep$ explains
    $38.44\%$ (resp. $2.76\%$) of the variance in the background color factor.
    Thus, the background color of samples obtained at 4k iterations change with
    the sampled representation $\modelrep=\modelphi(\x)$ across rows, while it
    stays the same at 38k iterations.
  }
  \label{supp:cmnistvis}
\end{figure}
}

\newcommand{\flowblock}{
\begin{figure}[htb]
\centering
\includegraphics[width=0.6\textwidth]{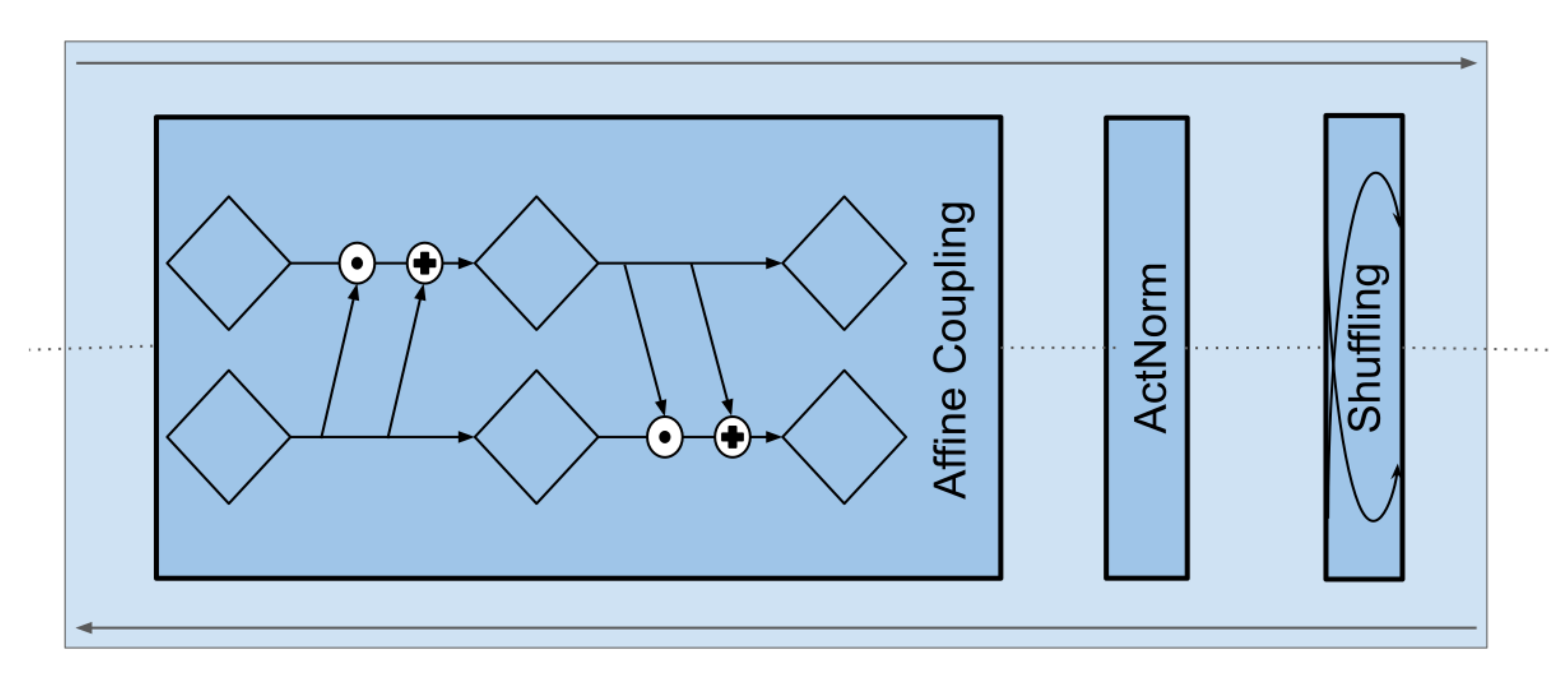}
\caption{A single invertible block used to build our invertible neural networks.}
  \label{fig:flowblock} \end{figure}
}

\newcommand{\flowarchs}{
\begin{figure}[htb]
\centering
\includegraphics[width=0.95\textwidth]{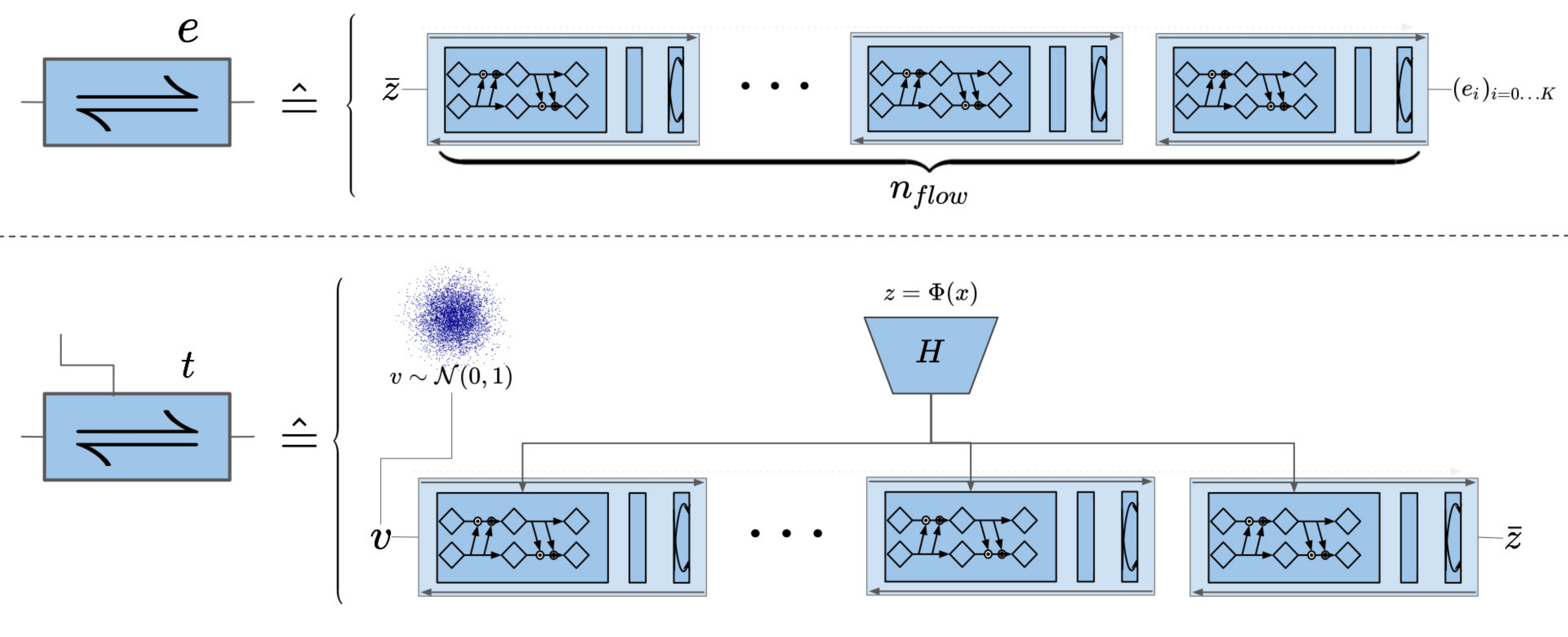}
\caption{Architectures of our INN models. \emph{top:} The semantic INN $\semanticinn$ consists of stacked invertible blocks. \emph{bottom:} The conditional INN $\condinn$ is composed of a embedding module $H$ that downsamples (upsamples if necessary) a given model representation $h = H(\modelrep) = H(\modelphi(\x))$. Subsequently, $h$ is concatenated to the inputs of each block of the invertible model.}
  \label{fig:flowarchs} \end{figure}
}

\newcommand{\bigganconcat}{
\begin{figure}[htb]
\centering
\includegraphics[width=0.95\textwidth]{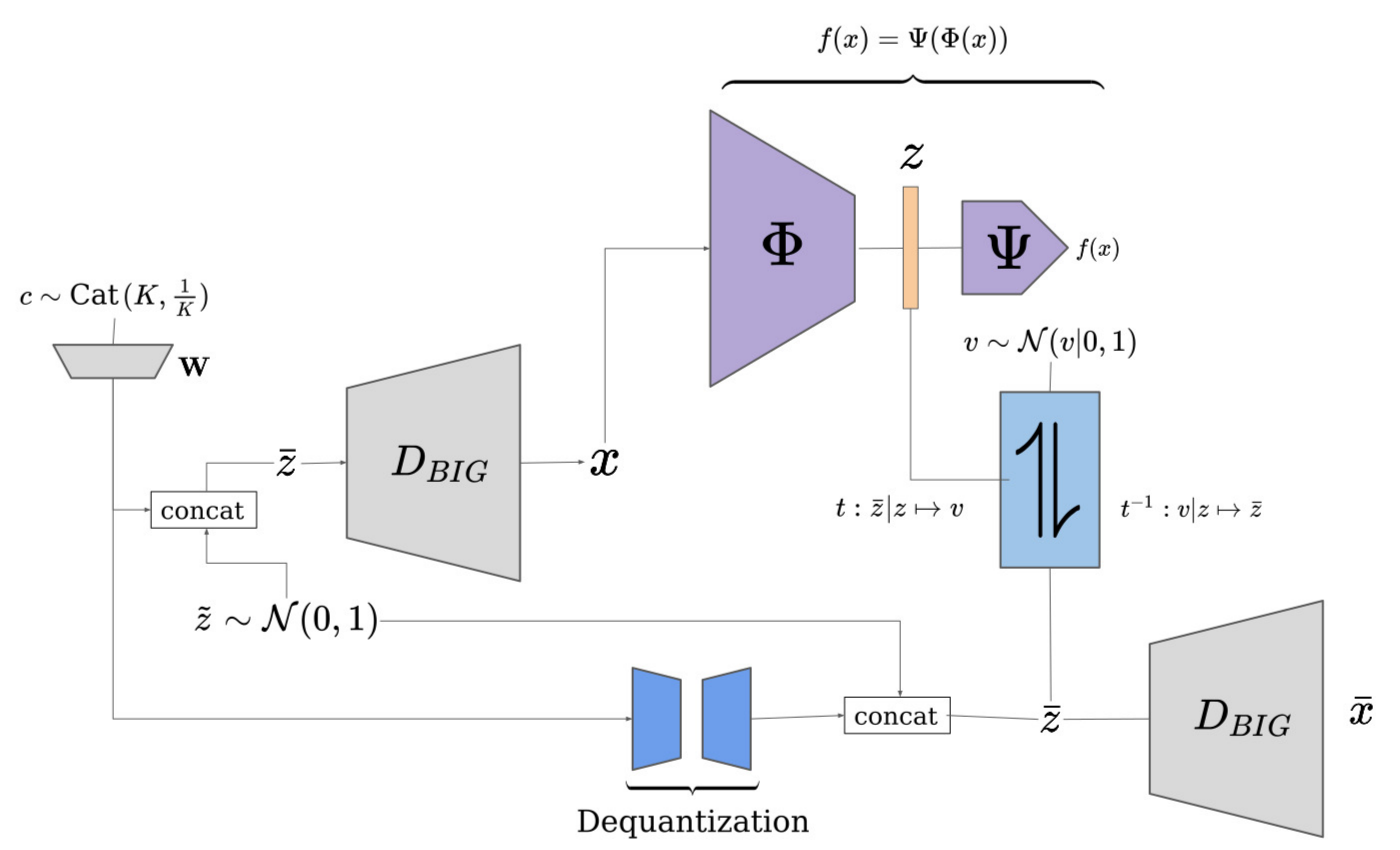}
\caption{Applying our approach to \emph{BigGAN} \cite{brock2018large}. We directly train $\condinn$ on latent codes of the generator model, utilizing a simple variational autoencoder model for dequantization of discrete classes $c$. See Sec.\ref{suppsec:domainshift} for technical details.}
\label{fig:concatbig} \end{figure}
}

\newcommand{\trainvstest}{
\begin{figure}[htb]
\centering
\includegraphics[width=0.55\textwidth]{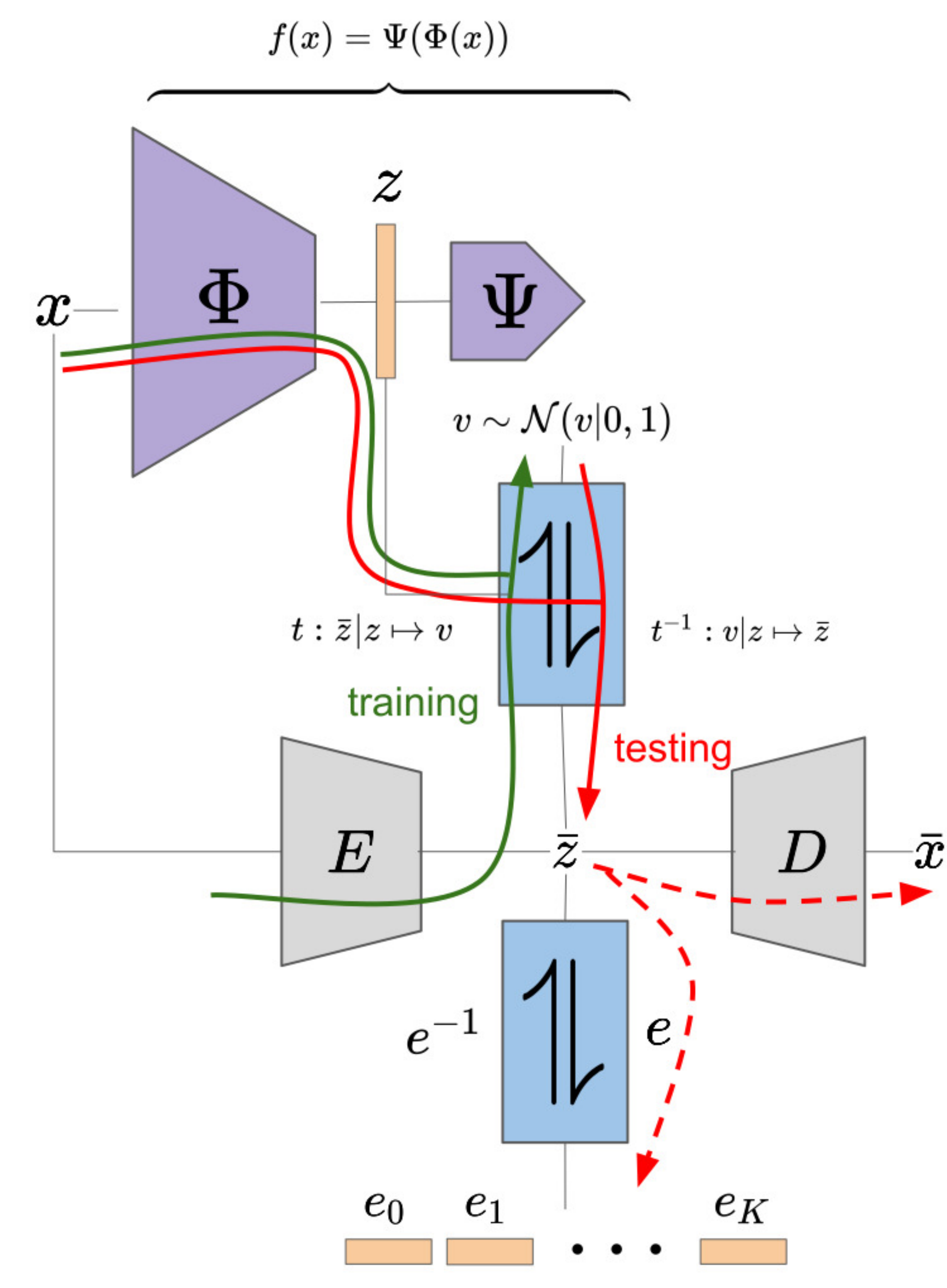}
\caption{Graphical distinction of information flow during training and inference. During training of $\condinn$, the encoder $\encoder$ provides an (approximately complete) data representation, which is used to learn the invariances of a given model's representations $\modelrep$. At inference, the encoder is not neccessarily needed anymore: Given a representation $\modelrep = \modelphi(\x)$, invariances can be sampled from the prior distribution and decoded into data space trough $\condinn^{-1}$.}
  \label{fig:traintest} \end{figure}
}

\newcommand{\aearchitecture}{
\begin{table}[htb]
  \caption{\label{tab:aearchitecture} Autoencoder architecture for
  \textsl{CelebA}, \textsl{AnimalFaces} and \textsl{Animals} at
  resolution $128 \times 128$.}
  \centering
  \parbox[b]{.45\textwidth}{
    \centering
    \caption{\label{tab:aeencoder} \textsl{Resnet-101} based Encoder.}
    \begin{tabular}[t]{c}
      \toprule
      RGB image $x\in \mathbb{R}^{128 \times 128 \times 3}$ \\
      \midrule
      Conv down $\to \mathbb{R}^{64\times 64 \times 64}$\\
      \midrule
      Norm, ReLU, MaxPool $\to \mathbb{R}^{32 \times 32 \times 64}$ \\
      \midrule
      $3\times$ BottleNeck $\to \mathbb{R}^{32 \times 32 \times 256}$ \\
      \midrule
      $4\times$ BottleNeck down $\to \mathbb{R}^{16 \times 16 \times 512}$ \\
      \midrule
      $23\times$ BottleNeck down $\to \mathbb{R}^{8 \times 8 \times 1024}$ \\
      \midrule
      $3\times$ BottleNeck down $\to \mathbb{R}^{4 \times 4 \times 2048}$ \\
      \midrule
      AvgPool, FC $\mapsto (\mu, \sigma^2) \in
      \mathbb{R}^{128}\times \mathbb{R}^{128}$ \\
      \bottomrule
    \end{tabular}
  }
  \hspace{1em}
  \parbox[b]{.45\textwidth}{
    \centering
    \caption{\label{tab:aedecoder} Decoder based on \textsl{BigGAN}.}
    \begin{tabular}[t]{c}
      \toprule
      $\aerep\in \mathbb{R}^{128} \sim \mathcal{N}(\mu, \diag(\sigma^2))$ \\
      $3\times$ (FC, LReLU) $\to \mathbb{R}^{256}$ \\
      FC, Softmax $\to \mathbb{R}^{1000}$ \\
      Embed $\mapsto h \in \mathbb{R}^{128}$ \\
      \midrule
      FC($\aerep$) $\to \mathbb{R}^{4\times 4\times 16\cdot96}$ \\
      \midrule
      ResBlock($\aerep$,$h$) up $\to \mathbb{R}^{8\times 8\times 16\cdot96}$ \\
      \midrule
      ResBlock($\aerep$,$h$) up $\to \mathbb{R}^{16\times 16\times 8\cdot96}$ \\
      \midrule
      ResBlock($\aerep$,$h$) up $\to \mathbb{R}^{32\times 32\times 4\cdot96}$ \\
      \midrule
      ResBlock($\aerep$,$h$) up $\to \mathbb{R}^{64\times 64\times 2\cdot96}$ \\
      \midrule
      Non-Local Block $\to \mathbb{R}^{64\times 64\times 2\cdot96}$\\
      \midrule
      ResBlock($\aerep$,$h$) up $\to \mathbb{R}^{64\times 64\times 96}$ \\
      \midrule
      Norm, ReLU, Conv up $\to \mathbb{R}^{128\times 128\times 3}$ \\
      \midrule
      Tanh $\mapsto \xrec\in\mathbb{R}^{128\times 128\times 3}$ \\
      \bottomrule
    \end{tabular}
  }
\end{table}
}
\newcommand{\aearchitecturecmnist}{
\begin{table}[b]
  \caption{\label{supp:aearchitecture:cmnist} Autoencoder architecture for
  \textsl{ColorMNIST} at resolution $28 \times 28$.}
  \centering
  \parbox[b]{.45\textwidth}{
    \centering
    \caption{\label{tab:cmnist:aeencoder} Encoder.}
    \begin{tabular}[t]{c}
      \toprule
      RGB image $x\in \mathbb{R}^{28 \times 28 \times 3}$ \\
      \midrule
      Conv, Norm, LReLU $\to \mathbb{R}^{14\times 14 \times 64}$\\
      \midrule
      Conv, Norm, LReLU $\to \mathbb{R}^{7\times 7 \times 128}$\\
      \midrule
      FC $\mapsto (\mu, \sigma^2) \in
      \mathbb{R}^{64}\times \mathbb{R}^{64}$ \\
      \bottomrule
    \end{tabular}
  }
  \parbox[b]{.45\textwidth}{
    \centering
    \caption{\label{tab:cmnist:aedecoder} Decoder.}
    \begin{tabular}[t]{c}
      \toprule
      $z\in \mathbb{R}^{64} \sim \mathcal{N}(\mu, \diag(\sigma^2))$ \\
      \midrule
      FC $\to \mathbb{R}^{7\times 7\times 128}$ \\
      \midrule
      Conv transpose, Norm, LReLU $\to \mathbb{R}^{14\times 14 \times 64}$\\
      \midrule
      Conv transpose, Tanh $\to \mathbb{R}^{28\times 28 \times 3}$\\
      \bottomrule
    \end{tabular}
  }
\end{table}
}
\newcommand{\aearchitecturedisc}{
\begin{table}[htb]
  \caption{\label{tab:aemetricarchitecture} Architectures used to compute
  image metrics for \textsl{CelebA}, \textsl{AnimalFaces} and
  \textsl{Animals} at resolution $128 \times 128$.}
  \centering
  \parbox[b]{.45\textwidth}{
    \centering
    \caption{\label{tab:aevgg} VGG-16 pretrained on ImageNet for
    feature extraction. Output of bold layers are used to compute
    feature distances.}
    \begin{tabular}[t]{c}
      \toprule
      RGB image $x\in \mathbb{R}^{128 \times 128 \times 3}$ \\
      \midrule
      \textbf{$2\times$ Conv, ReLU} $\to \mathbb{R}^{128\times 128 \times 64}$\\
      \midrule
      MaxPool $\to \mathbb{R}^{64\times 64 \times 64}$\\
      \midrule
      \textbf{$2\times$ Conv, ReLU} $\to \mathbb{R}^{64\times 64 \times 128}$\\
      \midrule
      MaxPool $\to \mathbb{R}^{32\times 32 \times 128}$\\
      \midrule
      \textbf{$3\times$ Conv, ReLU} $\to \mathbb{R}^{32\times 32 \times 256}$\\
      \midrule
      MaxPool $\to \mathbb{R}^{16\times 16 \times 256}$\\
      \midrule
      \textbf{$3\times$ Conv, ReLU} $\to \mathbb{R}^{16\times 16 \times 512}$\\
      \midrule
      MaxPool $\to \mathbb{R}^{8\times 8 \times 512}$\\
      \midrule
      \textbf{$3\times$ Conv, ReLU} $\to \mathbb{R}^{8\times 8 \times 512}$\\
      \bottomrule
    \end{tabular}
  }
  \hspace{1em}
  \parbox[b]{.45\textwidth}{
    \centering
    \caption{\label{tab:aedisc} Discriminator. All convolutions
    use kernel size 4. Norm refers to Batch Normalization, Leaky ReLU
    uses slope parameter 0.2.}
    \begin{tabular}[t]{c}
      \toprule
      RGB image $x\in \mathbb{R}^{128 \times 128 \times 3}$ \\
      \midrule
      Conv down, LReLU $\to \mathbb{R}^{64\times 64 \times 64}$\\
      \midrule
      Conv down, Norm, LReLU $\to \mathbb{R}^{32 \times 32 \times 128}$ \\
      \midrule
      Conv down, Norm, LReLU $\to \mathbb{R}^{16 \times 16 \times 256}$ \\
      \midrule
      Conv down, Norm, LReLU $\to \mathbb{R}^{8 \times 8 \times 512}$ \\
      \midrule
      Conv, Norm, LReLU $\to \mathbb{R}^{8 \times 8 \times 512}$ \\
      \midrule
      Conv $\to \mathbb{R}^{8 \times 8 \times 1}$ \\
      \bottomrule
    \end{tabular}
  }
\end{table}
}

\newcommand{\supptexturebias}{
\begin{figure}[htb]
\centering
\begin{tabular}{ccc}
 & \multicolumn{2}{c}{samples $\xrec = \decoder(\condinn^{-1}(\modelinv \vert \modelrep))$ conditioned on \emph{ResNet} pre-logits $\modelrep = \modelphi(\x)$}\\ 
  \cmidrule{2-3}
 \multirow{2}{*}{inputs}  & $\modelphi_{vanilla}$: ResNet-50 trained on & $\modelphi_{stylized}$: ResNet-50 trained on  \\
 & standard ImageNet & stylized ImageNet \\
\midrule
\includegraphics[scale=0.15, align=c]{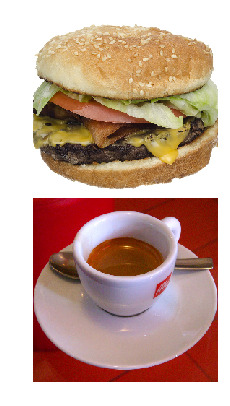} & 
\includegraphics[scale=0.15, align=c]{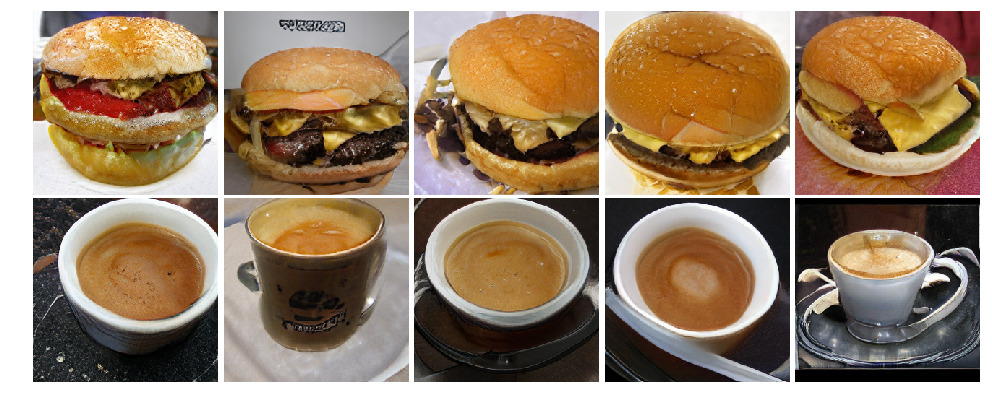} & 
\includegraphics[scale=0.15, align=c]{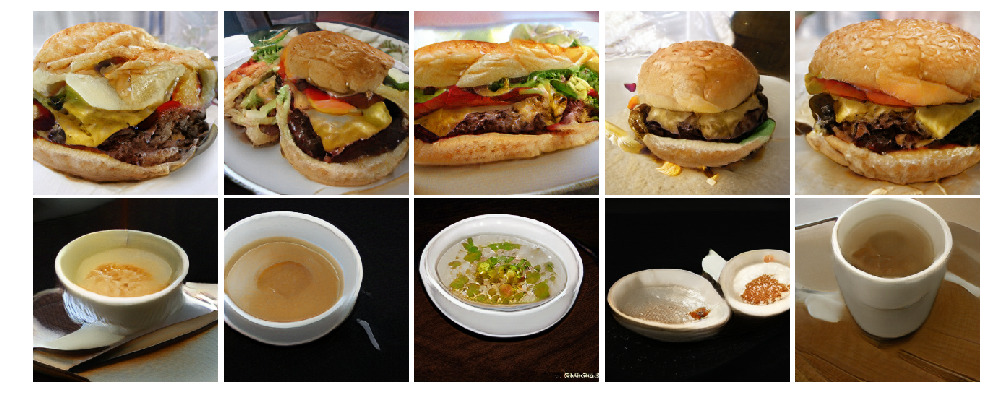} \\
\midrule
\includegraphics[scale=0.15, align=c]{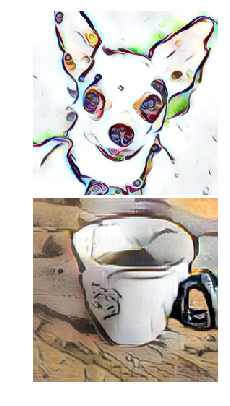} & 
\includegraphics[scale=0.15, align=c]{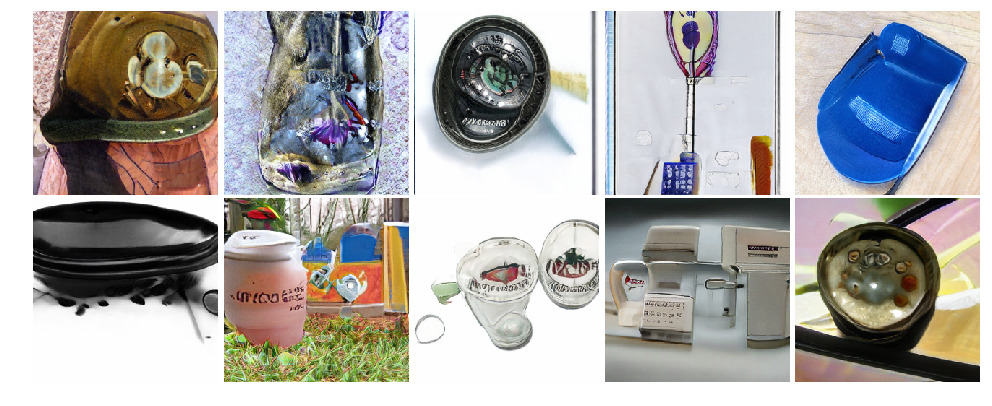} & 
\includegraphics[scale=0.15, align=c]{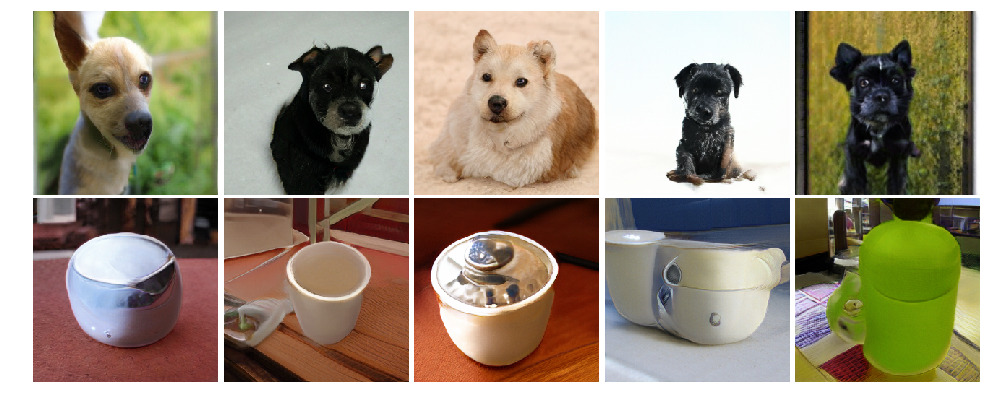} \\
\midrule
\includegraphics[scale=0.15, align=c]{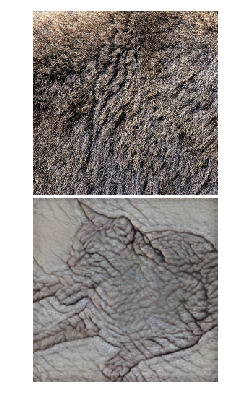} & 
\includegraphics[scale=0.15, align=c]{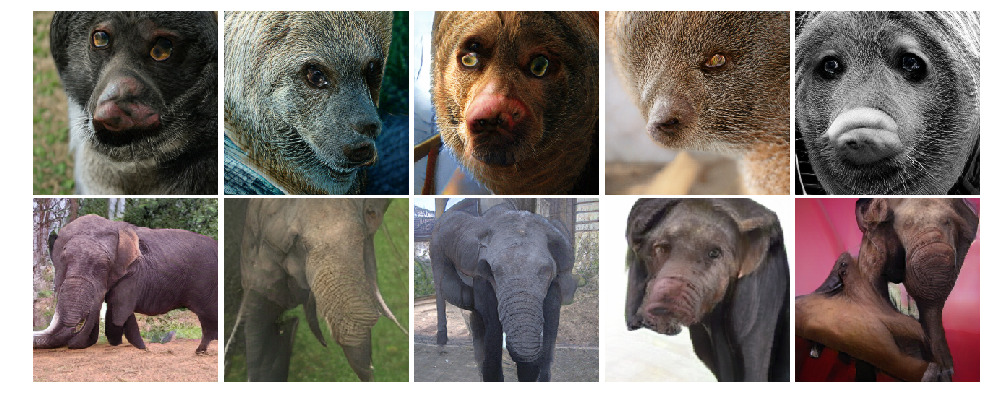} & 
\includegraphics[scale=0.15, align=c]{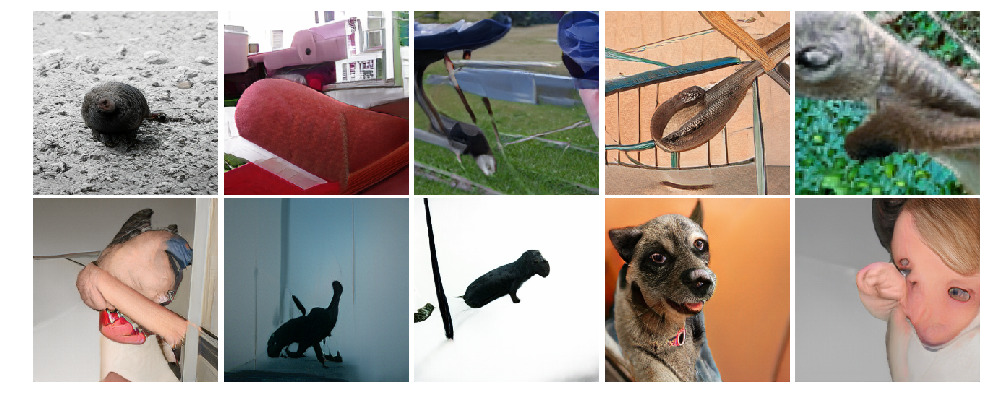} \\
\midrule
\includegraphics[scale=0.15, align=c]{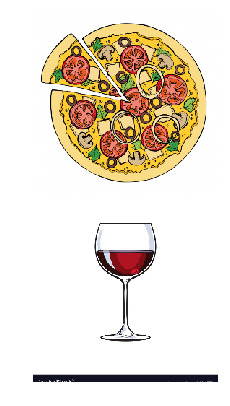} & 
\includegraphics[scale=0.15, align=c]{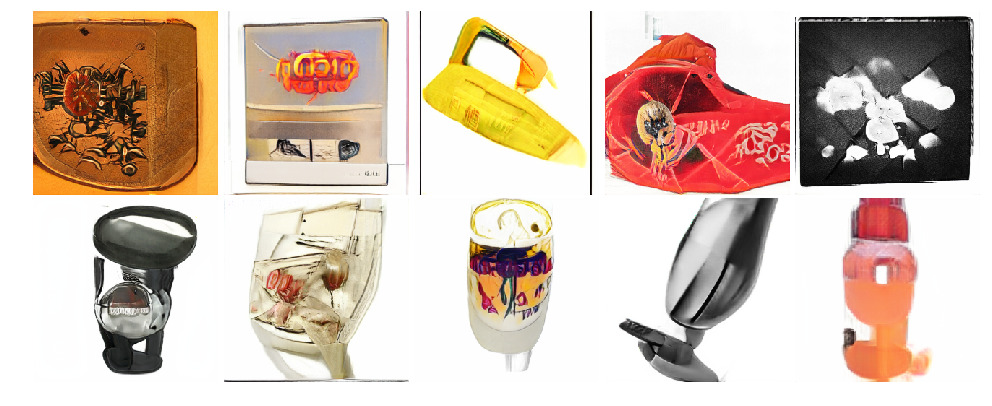} & 
\includegraphics[scale=0.15, align=c]{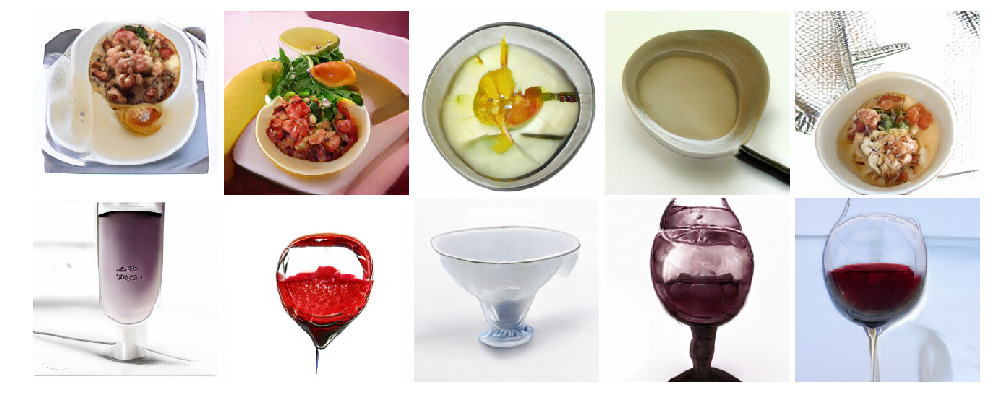} \\
\end{tabular}
\caption{\emph{Texture bias}: Additional examples for representation-conditional samples of two variants of \emph{ResNet-50}, one trained on standard \emph{ImageNet}, the other on a stylized version of \emph{ImageNet}. See also Tab.~\ref{fig:texturebias}.}
\label{supp:fig:texturebias}
\end{figure}
}

\newcommand{\suppcomparisontransposed}{
\begin{figure}[htb]
\centering
\begin{tabular}{ccccc}
 & \multicolumn{4}{c}{reconstructions from model representations} \\
  \cmidrule{2-5}
 & \multicolumn{2}{c}{example: snow leopard} & \multicolumn{2}{c}{example: wolf}\\ 
  \cmidrule{2-5}
layer &\textbf{our} & D\&B &\textbf{our} & D\&B \\
\midrule
input & 
\multicolumn{2}{c}{\includegraphics[scale=0.6, align=c]{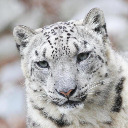}} &
\multicolumn{2}{c}{\includegraphics[scale=0.6, align=c]{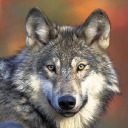}} \\
\midrule
conv5 & 
\includegraphics[scale=0.15, align=c]{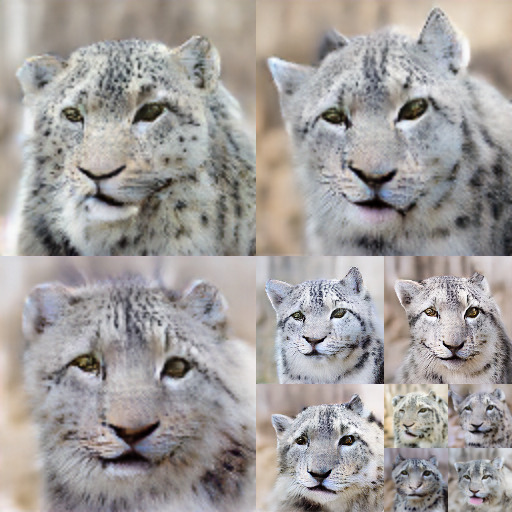} & 
\includegraphics[scale=0.6, align=c]{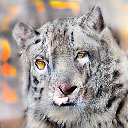} & 
\includegraphics[scale=0.15, align=c]{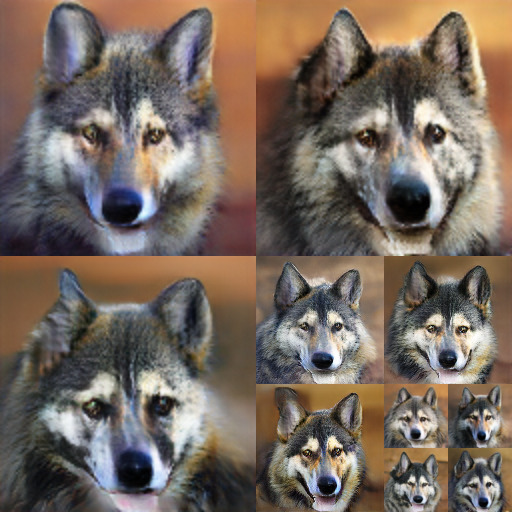} & 
\includegraphics[scale=0.6, align=c]{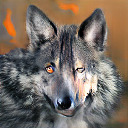} \\
\midrule
fc6 &
\includegraphics[scale=0.15, align=c]{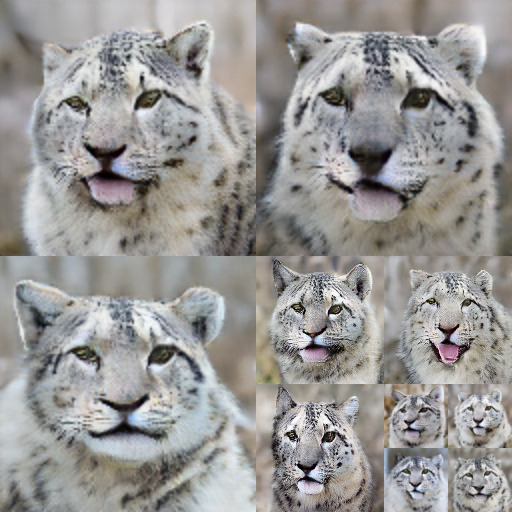} & 
\includegraphics[scale=0.6, align=c]{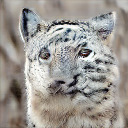} & 
\includegraphics[scale=0.15, align=c]{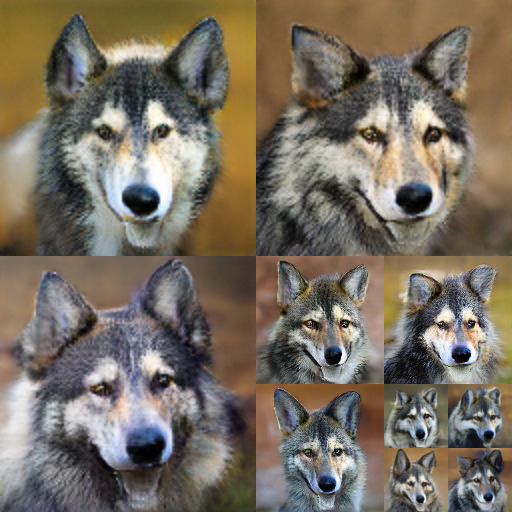} & 
\includegraphics[scale=0.6, align=c]{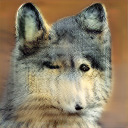} \\
\midrule
fc7 & 
\includegraphics[scale=0.15, align=c]{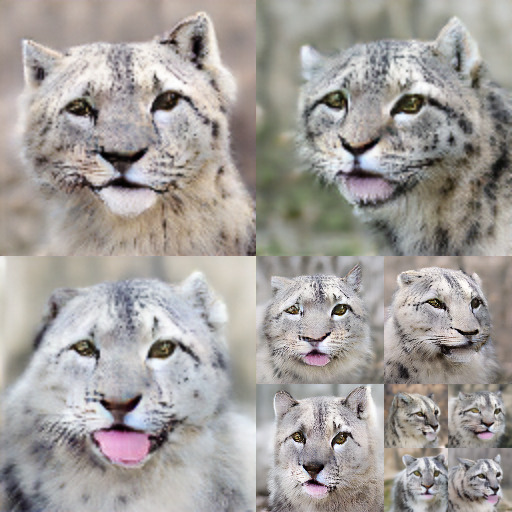} & 
\includegraphics[scale=0.6, align=c]{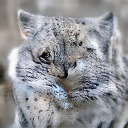} & 
\includegraphics[scale=0.15, align=c]{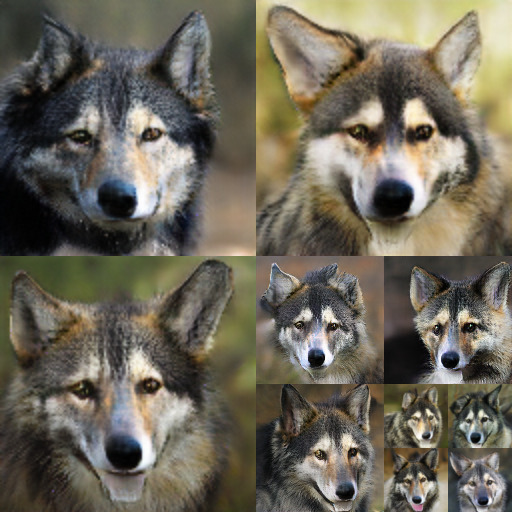} & 
\includegraphics[scale=0.6, align=c]{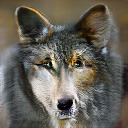} \\
\midrule
fc8 &
\includegraphics[scale=0.15, align=c]{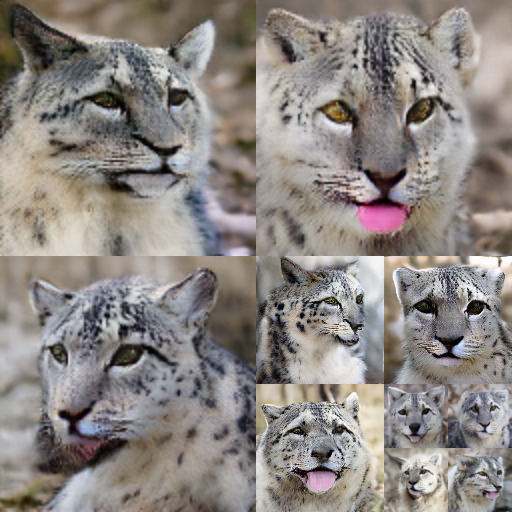} & 
\includegraphics[scale=0.6, align=c]{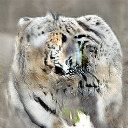} & 
\includegraphics[scale=0.15, align=c]{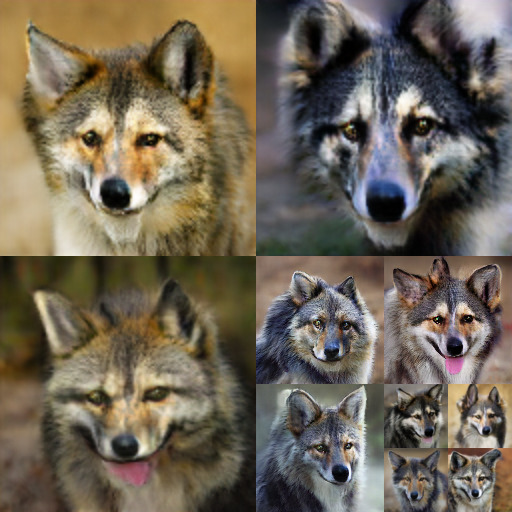} & 
\includegraphics[scale=0.6, align=c]{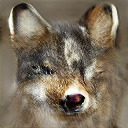} \\
\midrule
$\sigma(\text{logits})$ & 
\includegraphics[scale=0.15, align=c]{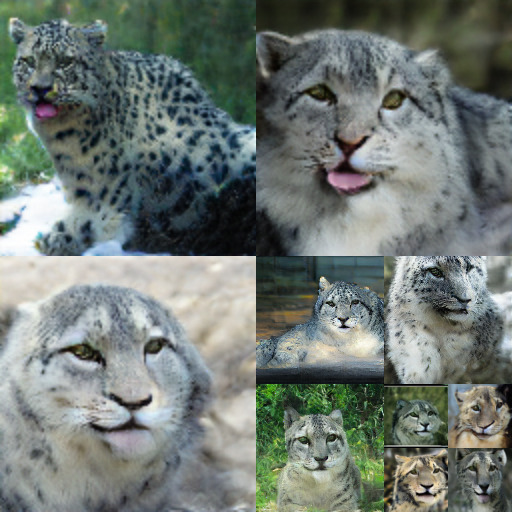} & 
\includegraphics[scale=0.6, align=c]{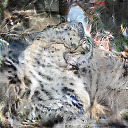} & 
\includegraphics[scale=0.15, align=c]{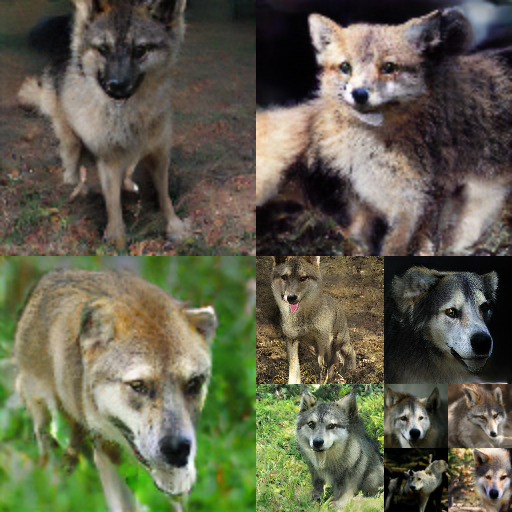} & 
\includegraphics[scale=0.6, align=c]{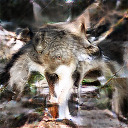} \\
\end{tabular}
\caption{Additional examples for layerwise reconstructions from model
  representations $\modelrep = \modelphi(\x)$ with our method and
  \cite{dosovitskiy2016generating} (D\&B). We show 10 samples per layer
  representation obtained with our generative approach. Here, $\sigma$ denotes
  the softmax function, \ie reconstructions are obtained from class
  probabilities provided by the model. We provide a comparison of equally sized
  images in Fig.~\ref{supp:fig:comparefairsnow} and
  Fig.~\ref{supp:fig:comparefairswolf}.}
\label{supp:fig:comparison}
\end{figure}
}

\newcommand{\suppcomparisontransposedfairsizesnow}{
\begin{figure}[htb]
\centering
\begin{tabular}{ccc}
 & \multicolumn{2}{c}{reconstructions from model representations} \\
  \cmidrule{2-3}
layer  &\textbf{our} & D\&B \\
\midrule
conv5 & 
\includegraphics[scale=0.3, align=c]{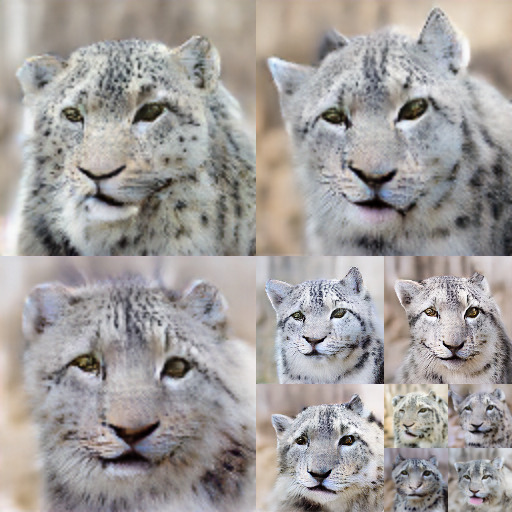} & 
\includegraphics[scale=0.6, align=c]{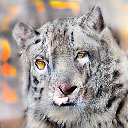} \\

\midrule
$\sigma(\text{logits})$ & 
\includegraphics[scale=0.3, align=c]{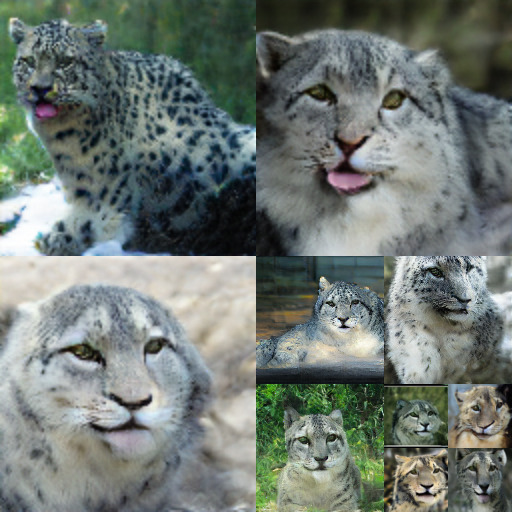} & 
\includegraphics[scale=0.6, align=c]{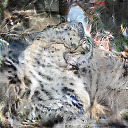} \\

\end{tabular}
\caption{Zooming into representation conditional samples for example \emph{snow
  leopard}. To verify that our samples are outperforming those of
  \cite{dosovitskiy2016generating} in visual quality, we repeat row 2 (conv5)
  and row 6 ($\sigma(\text{logits})$) of Fig.~\ref{supp:fig:comparison} with
  scaled images. Here, $\sigma$ denotes the softmax function.}
\label{supp:fig:comparefairsnow}
\end{figure}
}

\newcommand{\suppcomparisontransposedfairsizewolf}{
\begin{figure}[htb]
\centering
\begin{tabular}{ccc}
 & \multicolumn{2}{c}{reconstructions from model representations} \\
  \cmidrule{2-3}
layer  &\textbf{our} & D\&B \\

\midrule

conv5 & 
\includegraphics[scale=0.3, align=c]{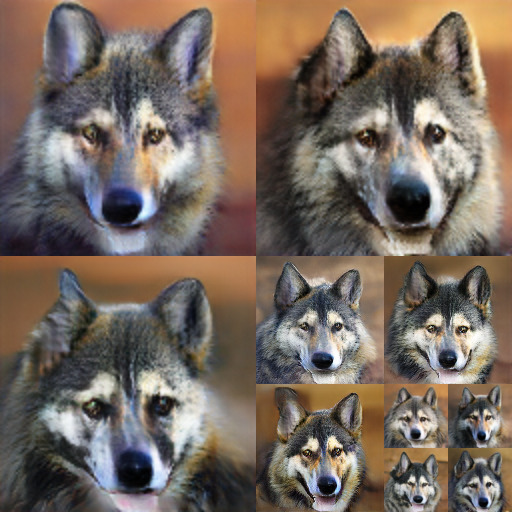} & 
\includegraphics[scale=0.6, align=c]{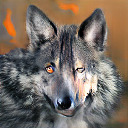} \\
\midrule
$\sigma(\text{logits})$ & 
\includegraphics[scale=0.3, align=c]{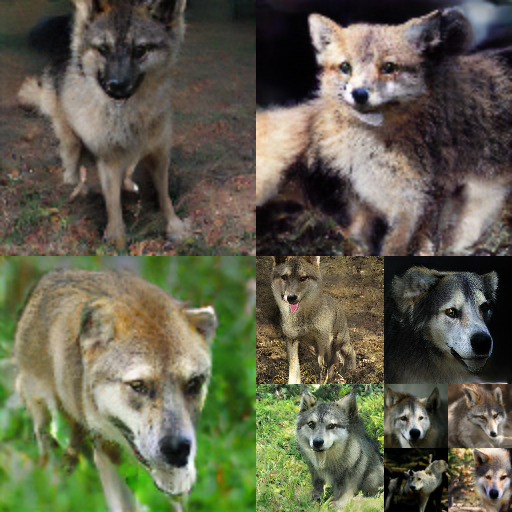} & 
\includegraphics[scale=0.6, align=c]{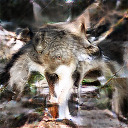} \\
\end{tabular}
\caption{Zooming into representation conditional samples for example
  \emph{wolf}. To verify that our samples are outperforming those of
  \cite{dosovitskiy2016generating} in visual quality, we repeat row 2 (conv5)
  and row 6 ($\sigma(\text{logits})$) of Fig.~\ref{supp:fig:comparison} with
  scaled images. Here, $\sigma$ denotes the softmax function.}
\label{supp:fig:comparefairswolf}
\end{figure}
}

\newcommand{\suppminivae}{
\begin{table}[htb]
 \centering
 \caption{\label{tab:minivae} Architecture of the VAE used for dequantization
  when training solely on synthetic BigGAN data. Here, a slope parameter of $\alpha = 0.01$ is used in Leaky ReLU.}
  \begin{tabular}[t]{c}
	\toprule
	Embedding $h\in \mathbb{R}^{128}$ \\
    \midrule
    (FC, LReLU) $\rightarrow \mathbb{R}^{4096}$\\
	\midrule
	$2\times$ (FC, LReLU) $\rightarrow \mathbb{R}^{4096}$\\
	\midrule
	$\mu, \sigma^2$: for each: \\
	\midrule
	(FC, LReLU) $\rightarrow \mathbb{R}^{128}$\\
    \midrule
    (FC, LReLU) $\rightarrow \mathbb{R}^{4096}$\\
	\midrule
	$2\times$ (FC, LReLU) $\rightarrow \mathbb{R}^{4096}$\\
	\midrule
	(FC, LReLU) $\rightarrow \mathbb{R}^{128}$\\
	\midrule
	$h \in \mathbb{R}^{128} \sim \normaldistr(\mu, \diag(\sigma^2))$ \\
    \midrule
    (FC, LReLU) $\rightarrow \mathbb{R}^{4096}$\\
    \midrule
    $3 \times $ (FC, LReLU) $\rightarrow \mathbb{R}^{4096}$\\
    \midrule
    (FC, LReLU) $\rightarrow \mathbb{R}^{128}$\\
    \bottomrule
   \end{tabular}  
\end{table}
}

\newcommand{\supphyperparams}{
\begin{table}[htb]
 \centering
 \caption{\label{tab:flowhyper} Hyperparameters of INNs for each experiment.
  $n_{flow}$ denotes the number of invertible blocks within in the model, see
  Fig.~\ref{fig:flowblock}. $h_w$ and $h_d$ refer to the width and depth of the
  fully connected subnetworks $s_i$ and $t_i$.}
  \begin{tabular}[t!]{cccccc}
	\toprule
	Experiment & INN & input dim. & $n_{flow}$ & $h_w$ & $h_d$ \\
	\midrule
	Comparison Sec.\ref{sec:compare} & $\condinn$ & 128 & 20 & 1024 & 2\\
	\midrule
	Understanding Models: FaceNet Sec.~\ref{subsubsec:noninvertiblenet} & $\condinn$ & 128 & 20 & 512 & 2 \\
	\midrule
	Understanding Models: FaceNet Sec.~\ref{subsec:factorevo} & $\semanticinn$ & 128 & 12 & 512 & 2 \\
	\midrule
	Data Effects: Adversarial Attack Sec.~\ref{sec:understandinputs} & $\condinn$ & 128 & 20 & 1024 & 2 \\
	\midrule
	Data Effects: Texture Bias Sec.~\ref{sec:understandinputs} & $\condinn$ & 268 & 20 & 1024 & 2 \\
	\midrule
	Data Effects: Domain Shift Sec.~\ref{suppsec:domainshift} & $\condinn$ & 128 & 20 & 1024 & 2 \\
	\midrule
	Modifications: FaceNet \& CelebA Sec.~\ref{sec:modrep} & $\semanticinn$ & 128 & 12 & 512 & 2 \\
  \bottomrule
  \end{tabular}  
\end{table}
}

\begin{center}
  \textbf{
  \Large Making Sense of CNNs: Interpreting Deep Representations \& Their Invariances with INNs} \\
  \Large 
-- \\
 \textbf{\large Supplementary Materials} \\
\hspace{1cm}
\end{center}

\section{Implementation Details}
\subsection{Autoencoder $\encoder, \decoder$}
\label{subsec:autoencoder}
\aearchitecturecmnist
\aearchitecturedisc
\aearchitecture
In Sec.~\ref{sec:recover}, we introduced an autoencoder to obtain a
representation $\aerep$ of $\x$, which includes the invariances abstracted away
by a given model representation $\modelrep$. This autoencoder consists of an
encoder $E(\x)$, and a decoder $D(\aerep)$.

Because the INNs $\condinn$ and $\semanticinn$ transform the distribution of
$\aerep$, we must ensure a strictly positive density for $\aerep$ to avoid
degenerate solutions. This is readily achieved with a stochastic encoder, \ie
we predict mean $\encoder(\x)_\mu$ and diagonal $\encoder(\x)_{\sigma^2}$ of a
Gaussian distribution, and obtain the desired representation as $\aerep \sim
\normaldistr(\aerep\vert \encoder(\x)_\mu, \diag(\encoder(\x)_{\sigma^2}))$.

Following \cite{dai2019diagnosing}, we train this autoencoder as a Variational
Autoencoder using the reparameterization trick \cite{VAE,VAE2} to match the
encoded distribution to a standard normal distribution, and jointly
learn the output variance $\gamma$ under an image metric $\Vert \x - \xrec
\Vert$ to avoid blurry reconstructions. The resulting loss function is thus
\begin{align}
  \loss(\encoder, \decoder, \gamma)
  = \mathbb{E}_{\begin{subarray}{l}x\sim p(x)\\ \epsilon \sim \normaldistr(\epsilon
  \vert 0, \id)\end{subarray}}
  &\left[ \frac{1}{\gamma}
  \Vert x - \decoder(\encoder(x)_\mu +
  \sqrt{\diag(\encoder(x)_{\sigma^2})}\;\epsilon) \Vert + \log \gamma \nonumber\right.\\
  &\left. +\frac{1}{2} \sum_{i=1}^{\aerepdim} \left\{
    (\encoder(\x)_\mu)_i^2 + (\encoder(\x)_{\sigma^2})_i - 1
    -\log (\encoder(\x)_{\sigma^2})_i \right\}
  \right] \nonumber
\end{align}

For experiments on \textsl{ColorMNIST}, we use the squared $L^2$ norm for the image
metric, and the encoder and decoder architectures are summarized in
Tab.~\ref{supp:aearchitecture:cmnist}.

For the experiments on \textsl{CelebA}, \textsl{AnimalFaces} and
\textsl{Animals}, we use an improved image metric as in
\cite{dosovitskiy2016generating}, which includes a perceptual loss and a
discriminator loss. The perceptual loss consists of feature distances obtained
from different layers of a fixed, pretrained network. We used a VGG-16 network
pretrained on ImageNet and weighted distances of different layers as in
\cite{zhang2018perceptual}. The discriminator is trained along with the
autoencoder to distinguish reconstructed images from real images using a binary
classification loss, and the autoencoder maximizes the log-probability that
reconstructed images are classified as real images. The architectures of VGG-16
and the discriminator are summarized in Tab.~\ref{tab:aemetricarchitecture}.
For $\encoder$ we use an architecture based on ResNet-101 and for $\decoder$ we
use an architecture based on BigGAN, where we include a small fully connected
network to replace the class conditioning used in BigGAN by a conditioning on
$\aerep$. See Tab.~\ref{tab:aearchitecture} for a summary of this autoencoder
architecture.

\subsection{Details on the INN for Revealing Semantics of Deep Representations}
\label{subsec:suppsemanticinn}

Previous works have successfully applied INNs for density estimation \cite{dinh2016density},
inverse problems \cite{ardizzone2018analyzing}, and on top of autoencoder
representations \cite{esser2020disentangling,xiao2019generative}. This section
provides details on how we embed the approach of \cite{esser2020disentangling}
to reveal the semantic concepts of autoencoder representations $\aerep$, \cf
Sec.~\ref{sec:semantic}.

Since we will never have examples for all relevant semantic concepts, we include a residual concept that captures the remaining variability of $\aerep$, which is not explained by the given semantic
concepts.

Following \cite{esser2020disentangling}, we learn a bijective transformation $\semanticinn(\aerep)$, which
translates the non-interpretable representation $\aerep$ invertibly into a
factorized representation
$(\semantic_i(\aerep))_{i=0}^\Nfactors=\semanticinn(\aerep)$, where each factor
$\semantic_i \in \RR^{\semanticdim{i}}$ represents one of the given semantic
concepts for $i = 1,\dots,\Nfactors$, and $\semantic_0 \in
\RR^{\semanticdim{0}}$ is the residual concept.

The INN $\semanticinn$ establishes a one-to-one correspondence between an
encoding and different semantic concepts and, conversely, enables semantic
modifications to correctly alter the original encoding (see next section).
Being an INN, $\semanticinn(\aerep)$ and  $\aerep$ need to have the same
dimensionality and we set $\semanticdim{0} = \aerepdim - \sum_{i=1}^\Nfactors
\semanticdim{i}$.  We denote the indices of concept $i$ with respect to
$\semanticinn(\aerep)$ as $\factorindices_i \subset \{1, \dots, \aerepdim\}$
such that we can write $\semantic_i =
(\semanticinn(\aerep)_k)_{k\in\factorindices_i}$.

\subsubsection*{Deriving a Loss Function for Training the Semantic INN}
Let $\semantic_i$ be the factor representing some semantic concept, \eg gender,
that the contents of two images $\x^a,  \x^b$ share. Then the projection of
their encodings $\aerep^a, \aerep^b$ onto this semantic concept must be
similar \cite{esser2020disentangling,kulkarni2015deep},
\begin{equation}
  \label{eq:semantic:invar}
  \semantic_i(\aerep^a) \simeq \semantic_i(\aerep^b) \quad \text{where }
  \aerep^a = \encoder(\x^a), \aerep^b = \encoder(\x^b) .
\end{equation}
Moreover, to interpret $\aerep$ we are interested in the separate contribution
of different semantic concepts $\semantic_i$ that explain $\aerep$. Hence, we
seek a mapping $\semanticinn(\bullet)$ that strives to disentangle different
concepts,
\begin{equation}
  \label{eq:semantic:indep}
  \semantic_i(\aerep) \perp \semantic_j(\aerep) \quad \forall i \neq j, \x \quad
  \text{where } \aerep = E(x) .
\end{equation}
The objectives in Eq.~\eqref{eq:semantic:invar}, \eqref{eq:semantic:indep}  imply a
correlation in $\semantic_i$ for pairs $\aerep^a$ and $\aerep^b$ and no
correlation between concepts $\semantic_i, \semantic_j$ for $i \neq j$. This
calls for a Gaussian distribution with a covariance matrix that reflects these
requirements.

Let $\semantic^a=(\semantic^a_i) = (\semantic_i(\encoder(\x^a)))$ and
$\semantic^b$ likewise, where $\x^a, \x^b$ are samples from a training
distribution $p(\x^a, \x^b)$ for the $i$-th semantic concept. The distribution
of pairs $\semantic^a$ and $\semantic^b$ factorizes into a conditional and a
marginal,
\begin{equation}
  \label{eq:pab}
  p(\semantic^a, \semantic^b) = p(\semantic^b \vert
  \semantic^a) p(\semantic^a)
\end{equation}
Objective Eq.~\eqref{eq:semantic:indep} implies a diagonal covariance for the
marginal distribution $p(\semantic^a)$, \ie a standard normal distribution, and
Eq.~\eqref{eq:semantic:invar} entails a correlation between $\semantic^a_i$ and
$\semantic^b_i$. Therefore, the correlation matrix is $\correlationmatrix =
\correlation \diag((\delta_{\factorindices_i}(k))_{k=1}^{\aerepdim})$.
By symmetry, $p(\semantic^b) = p(\semantic^a)$, which gives
\begin{equation}
  \label{eq:pbgivena}
  p(\semantic^b \vert \semantic^a) = \mathcal{N}(\semantic^b \vert
  \correlationmatrix \semantic^a,
  \id - (\correlationmatrix)^2) .
\end{equation}
Inserting Eq.~\eqref{eq:pbgivena} and a standard normal distribution for
$p(\semantic^a)$ into Eq.~\eqref{eq:pab} yields the negative log-likelihood for a
pair $\semantic^a, \semantic^b$. The detailed formulation can be found in the
supplementary material.

Given pairs $x^a, x^b$ as training data, another change of variables from
$\aerep^a=\encoder(x^a)$ to $\semantic^a=\semanticinn(\aerep^a)$ gives the
training loss function for $\semanticinn$ as the negative log-likelihood of
$\aerep^a, \aerep^b$,
  \begin{align}
    \label{eq:semanticinnloss}
    \loss(\semanticinn)
    = \mathbb{E}_{x^a, x^b} 
      &\left[-\log p(\semanticinn(\encoder(x^a)), \semanticinn(\encoder(x^b))) \right. \nonumber \\
      & \left. -\log \vert \det \nabla \semanticinn(\encoder(x^a)) \vert
      -\log \vert \det \nabla \semanticinn(\encoder(x^b)) \vert \right]
  \end{align}
For simplicity we have derived the loss for a single semantic concept
$\semantic_i$. Simply summing over the losses of different semantic concepts
yields their joint loss function and allows us to learn a joint translator
$\semanticinn $ for all of them.

\subsubsection*{Log-likelihood of Pairs}
The loss for $\semanticinn$ in Eq.~\eqref{eq:semanticinnloss} contains the
log-likelihood of pairs $\semantic^a, \semantic^b$.
Inserting Eq.~\eqref{eq:pbgivena} and a standard normal distribution for
$p(\semantic^a)$ into Eq.~\eqref{eq:pab} yields
\begin{align}
  \label{eq:logpab}
  -\log p(\semantic^a, \semantic^b)
  = \frac{1}{2} \left(
  \sum_{k\in \factorindices_i}
    \frac{(\semantic^b_k - \correlation \semantic^a_k)^2}{1-\correlation^2}
  + \sum_{k\in \factorindices_i^c}
    (\semantic^b_k)^2
  + \sum_{k=1}^{\aerepdim}
    (\semantic^a_k)^2
    \right) + C
\end{align}
where $C=C(\correlation, \aerepdim)$ is a constant that can be ignored for the
optimization process. $\rho \in (0,1)$ determines the relative importance
of loss terms corresponding to the similarity requirement in Eq.~\eqref{eq:semantic:invar} and the
independence requirement in Eq.~\eqref{eq:semantic:indep}. We use a fixed value
of $\rho=0.9$ for all experiments.

\flowblock
\flowarchs
\subsubsection*{Architecture of the Semantic INN}
In our implementation, $\semanticinn$ is built by stacking invertible blocks,
see Fig.~\ref{fig:flowblock}, which consist of three invertible layers:
coupling blocks \cite{dinh2016density}, actnorm layers \cite{kingma2018glow}
and shuffling layers.  The final output is split into the factors
$(\semantic_i)$, see Fig.~\ref{fig:flowarchs}.

Coupling blocks split
their input $x = (x_1 , x_2)$ along the channel dimension and use fully
connected neural networks $s_i$ and $t_i$ to perform the following computation:
\begin{align}
\tilde{x}_1 &= x_1\cdot s_1(x_2) + t_1(x_2) \\
\tilde{x}_2 &= x_2\cdot s_2(\tilde{x}_1) + t_2(\tilde{x}_1)
\label{eq:affinecoupling}
\end{align}
Actnorm layers consist of learnable shift and scale parameters for each
channel, which are initialized to ensure activations with zero mean and unit
variance on the first training batch.  Shuffling layers use a fixed, randomly
initialized permutation to shuffle the channels of its input, which provides a
better mixing of channels for subsequent coupling layers. 

\subsection{Conditional INN for Recovering Invariances of Deep Representations}
\subsubsection*{Architecture of the Conditional INN:}
We build the conditional invertible neural network $\condinn$ by expanding the semantic model $\semanticinn$ as follows:
Given a model representation $\modelrep$, which is used as the conditioning of the INN, we first calculate its embedding
\begin{equation}
h = H(\modelrep)
\end{equation}
which is subsequently fed into the affine coupling block:
\begin{align}
\tilde{x}_1 &= x_1\cdot s_1(x_2, h) + t_1(x_2, h) \\
\tilde{x}_2 &= x_2\cdot s_2(\tilde{x}_1, h) + t_2(\tilde{x}_1, h)
\label{eq:condaffinecoupling}
\end{align}
where $s_i$ and $t_i$ are modified from Eq.~\eqref{eq:affinecoupling} such that they are capable of processing a concatenated input $(x_i, h)$. The embedding module $H$ is usually a shallow convolutional neural network, used to down-/upsample a given model representation $\modelrep$ to a size that the networks $s_i$ and $t_i$ are able to process. 
This means that $\condinn$, analogous to $\semanticinn$, consists of stacked
invertible blocks, where each block is composed of coupling blocks, actnorm
layers and shuffling layers, \cf Sec.~\ref{subsec:suppsemanticinn} and Fig.~\ref{fig:flowblock}. The complete architectures of both $\condinn$ and $\semanticinn$ are depicted in Fig.~\ref{fig:flowarchs}.
Additionally, Fig.~\ref{fig:traintest} provides a graphical distinction of the
training and testing process of $\condinn$. During training, the autoencoder
$\decoder \compose \encoder$ provides a representation of the data that
contains both the invariances and the representation of some model \wrt the
input $\x$. After training of $\condinn$, the encoder may be discarded and
visual decodings and/or semantic interpretations of a model representation
$\modelrep$ can be obtained by sampling and transforming $\modelinv$ as
described in Eq.~\eqref{eq:sample}.
\trainvstest

\FloatBarrier
\section{Evaluation Details}
\supphyperparams
An overview of INN hyperparameters for all experiments is provided in
Tab.~\ref{tab:flowhyper}.
\subsection{Architectures of Interpreted Models}
\interpretedarchitectures
\interpretedarchitecturestwo
Throughout our experiments, we interpret four different models:
\emph{SqueezeNet}, \emph{AlexNet}, \emph{ResNet} and \emph{FaceNet}. Summaries
of each of model's architecture are provided in Tab.~\ref{tab:interpretmodels}
and Tab.~\ref{tab:interpretmodelstwo}. Implementations and pretrained weights of these models are taken from:
\begin{itemize}
\item \textsl{SqueezeNet (1.1)} {\scriptsize \url{https://pytorch.org/docs/stable/_modules/torchvision/models/squeezenet}}
\item \textsl{ResNet:} {\scriptsize \url{https://pytorch.org/docs/stable/_modules/torchvision/models/resnet.html}}
\item \textsl{AlexNet:} {\scriptsize\url{https://pytorch.org/docs/stable/_modules/torchvision/models/alexnet.html}}
\item \textsl{FaceNet:} {\scriptsize \url{https://github.com/timesler/facenet-pytorch}}
\end{itemize} 
\subsection{Explained Variance}
\label{supp:explainedvar}
To quantify the amount of invariances and semantic concepts, we use the
fraction of the total variance explained by invariances
(Fig.~\ref{fig:attackanimalslayers}) and the fraction of the variance of a
semantic concept explained by the model representation
(Fig.~\ref{fig:factorevolution}).

Using the INN $\condinn$, we can consider $\aerep = \condinn^{-1}(\modelinv
\vert \modelrep)$ as a function of $\modelinv$ and $\modelrep$. The total
variance of $\aerep$ is then obtained by sampling $\modelinv$, via its prior
which is a standard normal distribution, and $\modelrep$, via $\modelrep =
\modelphi(\x)$ with $\x \sim p_{\text{valid}}(\x)$ sampled from a validation set.
We compare this total variance to the average variance obtained when sampling
$\modelinv$ for a given $\modelrep$ to obtain the fraction of the total
variance explained by invariances:
\begin{equation}
  \label{eq:explainedbyinv}
\mathbb{E}_{\x' \sim p_{\text{valid}}(\x')} 
  \left[
  \frac{
    \mathrm{Var}_{\modelinv \sim
    \normaldistr(\modelinv \vert 0, \id)} \; \condinn^{-1}(\modelinv \vert
    \modelphi(\x'))
  }{
    \mathrm{Var}_{
      \begin{subarray}{l}
        \x \sim p_{\text{valid}}(\x)\\
        \modelinv \sim \normaldistr(\modelinv \vert 0, \id)
      \end{subarray}
      } \; \condinn^{-1}(\modelinv \vert
      \modelphi(\x))
    }
    \right]
\end{equation}

In combination with the INN $\semanticinn$, which transform $\aerep$ to
semantically meaningful factors, we can analyze the semantic content of a model
representation $\modelrep$. To analyze how much of a semantic concept
represented by factor $\semantic_i$ is captured by $\modelrep$, we use
$\semanticinn$ to transform $\aerep$ into $\semantic_i$ and measure its
variance. To measure how much the semantic concept is explained by $\modelrep$,
we simply swap the roles of $\modelrep$ and $\modelinv$ in
Eq.~\eqref{eq:explainedbyinv}, to obtain
\begin{equation}
  \label{eq:explainedbyrep}
  \mathbb{E}_{
    \modelinv' \sim \normaldistr(\modelinv' \vert 0, \id)
  }
  \left[
  \frac{
     \mathrm{Var}_{
      \x \sim p_{\text{valid}}(\x)
    } \; \semanticinn(\condinn^{-1}(\modelinv' \vert
    \modelphi(\x)))_i
  }{
    \mathrm{Var}_{
      \begin{subarray}{l}
        \modelinv \sim \normaldistr(\modelinv \vert 0, \id)\\
        \x \sim p_{\text{valid}}(\x)
      \end{subarray}
    } \; \semanticinn(\condinn^{-1}(\modelinv \vert
      \modelphi(\x)))_i
    }
    \right]
\end{equation}

Fig.~\ref{fig:attackanimalslayers} reports Eq.~\eqref{eq:explainedbyinv} and
its standard error when
evaluated via 10k samples, and
Fig.~\ref{fig:factorevolution} reports Eq.~\eqref{eq:explainedbyrep} and its
standard error when
evaluated via 10k samples.

\subsection{Comparison to Existing Visualization Methods}
\label{supp:comparison}
\suppcomparisontransposed
\suppcomparisontransposedfairsizewolf
\suppcomparisontransposedfairsizesnow
In Sec.~\ref{sec:compare}, we compare to existing layer inversion methods that aim to reconstruct an input $\x$ from its representation $\modelrep = \modelphi(\x)$. Both our method and D\&B's ~\cite{dosovitskiy2016generating} method were trained on the \emph{Animals} dataset, which consists of a mixture of all carnivorous mammal animal classes from \emph{ImageNet} and all animals from the \emph{Animals with Attributes 2} \cite{xian2018zero} dataset. Hyperparameters of our autoencoder model can be found in Tab.~\ref{tab:aearchitecture}. The decoder in \cite{dosovitskiy2016generating} was re-implemented based on our decoder shown in Tab.~\ref{tab:aedecoder}, where we set the latent dimension to 4096 to avoid introduction of an artificial bottleneck and allow for a fair comparison. Both methods were trained by minimizing the image metric described in Sec.\ref{subsec:autoencoder} and Tab.~\ref{tab:aemetricarchitecture}, where no Kullback-Leibler divergence term was used for D\&B's method. Images from \cite{mahendran2016visualizing} are taken from their publication.
Additional visual comparisons can be found in Fig.~\ref{supp:fig:comparison},~\ref{supp:fig:comparefairswolf},~\ref{supp:fig:comparefairsnow}.
\FloatBarrier

\subsection{Relevance of Factors}
\suppcmnistvis
In Sec.~\ref{sec:understandingmodels}, we trained \textsl{SqueezeNet} for digit
classification on \textsl{ColorMNIST}, which consists of \textsl{MNIST} images
with randomly choosen fore- and background colors. In addition, we trained the
autoencoder of Tab.~\ref{supp:aearchitecture:cmnist} on \textsl{ColorMNIST} and
the INN $\semanticinn$ to obtain the following factors
\begin{itemize}
  \item $\semantic_1$ representing the digit class defined by pairs of images
    showing the same digit in different styles and colors,
  \item $\semantic_2$ representing the foreground color defined by pairs of
    images showing the same foreground color on different digits and
    backgrounds,
  \item $\semantic_3$ representing the background color defined by pairs of
    images showing the same background color for differently colored digits.
\end{itemize}

Finally, we trained the INN $\condinn$ for 20 different checkpoints of
\textsl{SqueezeNet} obtained between training steps zero and 40k, to obtain the
stochastic mapping from $\modelrep$, the penultimate \textsl{Fire} layer of
\textsl{SqueezeNet}, to the semantic factors $(\semantic_i)$.
Fig.~\ref{fig:factorevolution} plots Eq.~\eqref{eq:explainedbyrep}
against the training step, with shaded areas representing the standard error
obtained with 10k samples.

At step zero, \ie for a randomly initialized \textsl{SqueezeNet}, we observe
that the representation $\modelrep$ mostly contains the background color and,
to a lesser degree, the foreground color. This observation is consistent with
the fact that color information is directly encoded in the pixel representation
of the image and that there are more background pixels than foreground pixels.
In contrast, information about the digit class is not directly encoded in pixel
values and requires learning.
As the network starts to learn between steps 10k and 15k, we indeed observe a
drastic change in the semantic content of $\modelrep$, which becomes invariant
to color information and sensitive to digit class information. Note that the
network could also learn to retain color information while seperating digit
classes in the last classification layer, but our results demonstrate that the
network learns to abstract away task-irrevant information before that.

We show additional $\modelrep$ conditional samples, both before and after
learning, in Fig.~\ref{supp:cmnistvis}.

\newpage
\subsection{Modifying Representations}
\label{supp:modrep}
\suppcelebamod
\suppcelebamodtwo
\paragraph{Training Details:} In Sec.~\ref{sec:modrep} we trained the autoencoder of
Tab.~\ref{tab:aearchitecture} on \textsl{CelebA} at resolution $128\times 128$.
Using the attribute labels provided for this dataset, we
trained an INN $\semanticinn$ for the semantic factors
\begin{itemize}
  \item $\semantic_1$ representing hair color, defined by pairs %
    with the same %
    \textsl{Black\_Hair} attribute.
  \item $\semantic_2$ representing glasses, defined by pairs %
    with the same %
    \textsl{Eyeglasses} attribute.
  \item $\semantic_3$ representing gender, defined by pairs %
    with the same %
    \textsl{Male} attribute.
  \item $\semantic_4$ representing beard, defined by pairs %
    with the same %
    \textsl{No\_Beard} attribute.
  \item $\semantic_5$ representing age, defined by pairs %
    with the same %
    \textsl{Young} attribute.
  \item $\semantic_6$ representing smiling, defined by pairs %
    with the same %
    \textsl{Smiling} attribute.
\end{itemize}

\subsubsection*{Additional Results and Comparisons} We provide a larger version of
Fig.~\ref{fig:facemod} with more examples in Fig.~\ref{supp:celebamod} and
Fig.~\ref{supp:celebamodtwo}. While our approach aims to provide semantic
understanding of representations learned by models, the invertibility of
$\semanticinn$ together with the decoder $\decoder$ enables semantic image
editing. To evaluate our approach on this task, we compare it to
\textsl{StarGAN}\footnote{We used the author's official implementation
available at \url{https://github.com/yunjey/stargan}} \cite{choi2018stargan}, a
specialized approach for attribute modifications of face images.
Our approach consistently outperforms \cite{choi2018stargan} across all semantic
attributes in terms of the quality of modified images, which is quantified by FID
scores \cite{heusel2017gans} in Fig.~\ref{supp:celebamodtwo}.
Moreover, we
observe some particular qualitative differences between our method and
\cite{choi2018stargan}: Changing factors with our approach produces
more coherent changes, \ie changes in gender cause changes in hair length (for
all examples in Fig.~\ref{supp:celebamod}),
changes to an older age cause thin, white hairs (\eg examples 1, 2, 6
in Fig.~\ref{supp:celebamod}),
and changes in the beard factor have no effect on female faces (\eg examples 2,
3, 5, 6 in Fig.~\ref{supp:celebamod}), suggesting that
our model has learned the correct causal structure (as present in the data) where beard is caused by gender and not the other way around. In contrast, \cite{choi2018stargan}
produces very localized, pixelwise changes without taking the global structure
into account. While such a behavior might be desired for some specialized
applications, it generally leads to unnatural results, \eg when changing
gender, beard and age in example 2 or gender and beard in example 3 of
Fig.~\ref{supp:celebamod}.
\FloatBarrier

\subsection{Effects of Data Shifts - Additional Results}
\label{suppsec:domainshift}
\domainshiftfacenet
\subsubsection*{Data Shift from Humans to Animals} As an extension of Sec.~\ref{sec:understandinputs}, we run an experiment on
FaceNet and condition the invariance recovering model $\condinn$ on five
different representations of the model (see Tab.~\ref{tab:interpretfacenet}) by
training $\condinn$ on \textsl{AnimalFaces} instead of \textsl{CelebA} and an
autoencoder which is trained on both  \textsl{AnimalFaces} and \textsl{CelebA},
\cf Tab.~\ref{tab:aearchitecture} for details. Furthermore, note that FaceNet
is not re-trained on the new data and fixed during training, \cf Fig.~\ref{fig:traintest}.

Fig.~\ref{fig:facenetdomainshift} depicts the visualized representations and corresponding learned invariances accross several layers of FaceNet. Evidently, even deep representations of the off-domain input image may be visualized, at least as deep as the penultimate layer (\textbf{AdaAvgPool}).
Another interesting result is that FaceNet seems to conserve class identity of the input to some degree: The appearance of samples generated by conditioning on model representations is similar, to some extend even for the last layer (\textbf{identity embedding}). This suggests that the model is able to generalize to a surprisingly large margin of data, given the input images show some kind of symmetry and perceptual similarity to human faces.

\subsubsection*{Verifying the Texture-Bias Hypothesis}
\bigganconcat
\suppminivae
\supptexturebias
In Section~\ref{sec:understandinputs} we trained the INN $\condinn$ conditioned on representations of ResNet-50 from the penultimate layer (\ie extracted before the final classification layer, \cf Tab.~\ref{tab:interpretresnet}) with the goal of validating the \emph{texture-bias hypothesis} from \cite{geirhos2018imagenet}. In their work, \cite{geirhos2018imagenet} showed that typical convolutional neural classification networks are biased towards texture when being trained on ImageNet. They proposed that this bias can be removed by training the CNNs on a \emph{stylized} version of ImageNet instead.

Following \cite{rombach2020network}, we gained access to the dataset and a powerful decoder by relying on a synthetic version of ImageNet, provided through the pre-trained generator of BigGAN \cite{brock2018large}.\footnote{We used a pretrained generator available at \url{https://github.com/LoreGoetschalckx/GANalyze}} Thus, with Eq.~\eqref{eq:cinnloss} in mind, we identify the concatenated vector $(\tilde{z}, \mathbf{W}c))$ as $\aerep$. Here, $\mathbb{R}^{140} \ni \tilde{z} \sim \normaldistr(0, \id)$ is sampled from a multivariate normal distribution and $c\in \{0,1\}^{K}$ is a one-hot vector representing one of the $K=1000$ ImageNet classes. 
$\mathbf{W}$ maps the one-hot class representation $c$ to the space of real numbers, \ie $\mathbf{W}c = h \in \mathbb{R}^{128}$. Note that $\mathbf{W}$ is part of the BigGAN generator $\decoder_{BIG}$ and is thus also pre-trained.
See Fig.~\ref{fig:concatbig} for a visual summaray of the application of our approach to BigGAN. To avoid overfitting $\condinn$ on a single dimension of $\aerep$, the vector $h$ is passed trough a small, fully connected variational autoencoder before being concatenated with $\tilde{z}$ as $\aerep = (\tilde{z}, h)$. The architecture of this VAE is depicted in Tab.~\ref{tab:minivae}. Utilizing this approach can be interpreted as a variant of deep dequantization. Equipped with a dequantized version of $\aerep = (\tilde{z}, h)$ and corresponding images $\x = \decoder_{BIG}(\aerep)$, we trained $\condinn$ as described in Sec.~\ref{sec:recover}.

Additional samples conditioned on representations of (i) a ResNet-50 trained on
standard ImageNet and (ii) a ResNet-50 trained on the stylized version
of ImageNet are provided in Fig.~\ref{supp:fig:texturebias}. 
These results further confirm the texture\&shape-bias of (i) and the reverse
behavior for (ii). Line 7 and 8 explicitly show that a texture-agnostic classifier can be used to create new content based on input sketches or cartoons.

Furthermore, note that both models perform reasonably well on the domain
of natural images, \cf line 1-2 of Tab.~\ref{supp:fig:texturebias}.
\extraattacksfig